\documentclass[10pt,twocolumn,letterpaper]{article}

\usepackage{3dv}
\usepackage{times}
\usepackage{epsfig}
\usepackage{graphicx}
\usepackage{amsmath}
\usepackage{amssymb}

\usepackage{booktabs}
\usepackage{subcaption}
\usepackage{xcolor}
\usepackage{multirow}
\usepackage{tabularx}
\usepackage{colortbl}
\usepackage{tikz}
\usepackage{pgfplots}
\usepackage[numbers,sort]{natbib}

\newcommand{\comment}[1]{}

\definecolor{DarkGreen}{rgb}{0.0, 0.5, 0.0}

\definecolor{BrightRoyalPurple}{rgb}{0.65, 0.05, 0.78}

\newcommand{\PNP}{P\emph{n}P}

\usepackage[pagebackref=true,breaklinks=true,letterpaper=true,colorlinks,bookmarks=false]{hyperref}

\threedvfinalcopy 

\def\threedvPaperID{3} 

\begin{document}

\title{Location Field Descriptors:\\Single Image 3D Model Retrieval in the Wild}

\author{Alexander Grabner$^1$\\
\and
Peter M. Roth$^1$\\
\and
Vincent Lepetit$^{2,1}$\\
\and
\small $^1$Institute of Computer Graphics and Vision, Graz University of Technology, Austria\\
\and
\small $^2$Laboratoire Bordelais de Recherche en Informatique, University of Bordeaux, France\\
\and
{\tt\small \{alexander.grabner,pmroth,lepetit\}@icg.tugraz.at}
}

\maketitle

\begin{abstract}
We present Location Field Descriptors, a novel approach for single image 3D model retrieval in the wild. In contrast to previous methods that directly map 3D models and RGB images to an embedding space, we establish a common low-level representation in the form of location fields from which we compute pose invariant 3D shape descriptors. Location fields encode correspondences between 2D pixels and 3D surface coordinates and, thus, explicitly capture 3D shape and 3D pose information without appearance variations which are irrelevant for the task. This early fusion of 3D models and RGB images results in three main advantages: First, the bottleneck location field prediction acts as a regularizer during training. Second, major parts of the system benefit from training on a virtually infinite amount of synthetic data. Finally, the predicted location fields are visually interpretable and unblackbox the system. We evaluate our proposed approach on three challenging real-world datasets (Pix3D, Comp, and Stanford) with different object categories and significantly outperform the state-of-the-art by up to 20\% absolute in multiple 3D retrieval metrics.
\end{abstract}

\section{Introduction}
\label{sec:intro}

\begin{figure}[t]
	\begin{center}
		\includegraphics[width=\linewidth]{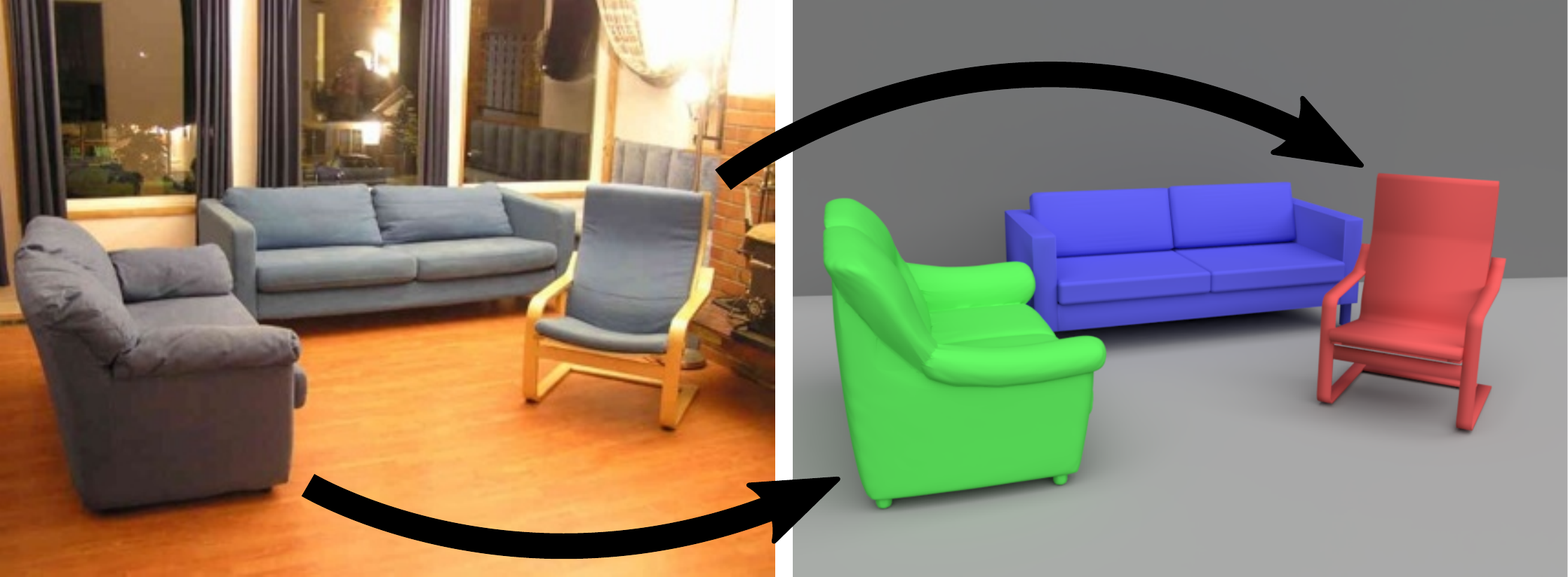}
	\end{center}
	\vspace{-0.25cm}
	\caption{Given a single RGB image, we retrieve a 3D model with accurate geometry for each object in the image from a previously seen or unseen 3D model database.}
	\label{fig:teaser}
\end{figure}

3D model retrieval from a single RGB image, as shown in Fig.~\ref{fig:teaser}, is a challenging but important task with applications in augmented reality, robotics, 3D printing, 3D scene understanding, and 3D scene modeling. Compared to reconstruction~\cite{Girdhar2016learning,Tulsiani2018factoring}, retrieval provides 3D models designed by humans which are rich in detail. Due to the growing number of large-scale 3D model databases, like ShapeNet~\cite{Shapenet2015} or 3D Warehouse\footnote{\url{https://3dwarehouse.sketchup.com}}, efficient image-based retrieval approaches have become a fundamental requirement.

Recent works address the retrieval task by directly mapping 3D models and RGB images to a common embedding space~\cite{Li2015joint,Tasse2016shape2vec}. However, previous approaches have a number of limitations in practice. First, the learned mapping is highly prone to overfitting, because training data in the form of RGB images with 3D model annotations is scarce~\cite{Sun2018pix3d,Wang2018fine}. Second, systems purely trained on synthetic data do not generalize to real data due to the domain gap between RGB images and RGB renderings~\cite{Massa2016deep,Girdhar2016learning}. Finally, the black box characteristic of these systems makes it hard to understand why the approaches fail in certain scenarios.

To overcome these limitations, we map 3D models and RGB images to a common low-level representation in the form of location fields from which we compute pose invariant 3D shape descriptors. Location fields~\cite{Taylor2012vitruvian,Wang2018fine} are image-like representations that encode a 3D surface coordinate for each object pixel (see Fig.~\ref{fig:representations}). In particular, we render location fields from 3D models and predict location fields from RGB images using a CNN. Then, instead of exhaustively comparing location fields from different viewpoints~\cite{Massa2016deep,Izadinia2017im2cad}, we compute pose invariant 3D shape descriptors in an embedding space optimized for retrieval from the location fields. Thus, we call our approach Location Field Descriptors.

Regarding 3D model retrieval, location fields have several advantages compared to other rendered representations: RGB renderings~\cite{Girdhar2016learning,Xiang2016objectnet3d} are subject to appearance variations which are irrelevant for the task caused by material, texture, and lighting. Texture-less gray-scale renderings~\cite{Massa2016deep,Lee2018cross} are still affected by the scene lighting. Silhouettes~\cite{Chen2003visual} are not affected by such appearance variations but discard valuable 3D shape information. Depth~\cite{Grabner2018a,Zhang2018learning} and normal renderings~\cite{Wu2018learning,Wang20183d} capture 3D geometry but lose the relation to the 3D pose in the object's canonical coordinate system. In contrast, location fields explicitly present 3D shape and 3D pose information, as they establish correspondences between 2D object pixels and 3D coordinates on the object surface. Considering 3D shape, the dense 3D coordinates provide a partial reconstruction of the object geometry. Considering 3D pose, the object rotation and translation can be geometrically recovered from the 2D-3D correspondences using a \PNP~algorithm~\cite{Brachmann2016uncertainty,Jafari2018ipose}.

The benefits of our approach are threefold: First, the intermediate location field prediction serves as a regularizing bottleneck which reduces the risk of overfitting in the case of limited training data compared to directly mapping to an embedding space. Second, major parts of the system benefit from training on a virtually infinite amount of synthetic data due to the early fusion of 3D models and RGB images. Third, the predicted location fields are visually interpretable and offer valuable insights in cases where the approach fails.

Finally, to demonstrate the benefits of our novel 3D model retrieval approach, we evaluate it on three challenging real-world datasets with different object categories: Pix3D~\cite{Sun2018pix3d} (\textit{bed}, \textit{chair}, \textit{sofa}, \textit{table}), Comp~\cite{Wang2018fine} (\textit{car}), and Stanford~\cite{Wang2018fine} (\textit{car}). We present quantitative as well as qualitative results and significantly outperform the state-of-the-art. To summarize, our main contributions are:
\begin{itemize}
	\item We present the first method that uses location fields for pose invariant 3D model retrieval. Our approach is accurate, scalable, and interpretable.
	\item We outperform the state-of-the-art by up to 20\% absolute in multiple 3D retrieval metrics given both previously seen and unseen 3D model databases.
\end{itemize}

\section{Related Work}
\label{sec:relatedwork}
In this section, we discuss previous works in the fields of 3D coordinate regression and 3D model retrieval from a single RGB image.

\subsection{Location Fields}
Regressing 3D coordinates from 2D observations is a well-studied problem in computer vision~\cite{Hartley2003multiple}. While traditional approaches generate 3D point clouds from multi-view RGB images~\cite{Stockman2001cv,Szeliski2010computer}, recent works predict unstructured 3D point clouds from a single RGB image using deep learning~\cite{Kl2019capnet,Mandikal20183dlmnet,Fan2017point}. Complementary to these works, PointNet~\cite{Qi2017pointnet} and successors~\cite{Qi2017pointnet++,Klokov2017escape} showed that even such unstructured 3D point clouds can be used to address various 3D vision tasks with deep learning.

In this work, however, we focus on predicting structured 3D point clouds in the form of location fields~\cite{Taylor2012vitruvian,Wang2018fine}. A location field encodes a 3D surface coordinate for each object pixel. Thus, it is important to know which pixels belong to an object and which pixels belong to the background or another object~\cite{Brachmann2014learning,Brachmann2016uncertainty}. Recent works showed that deep learning techniques for instance segmentation~\cite{He2017mask} significantly increase the accuracy on this task~\cite{Jafari2018ipose,Grabner2019a,Wang2019normalized}. However, until now location fields have only been used for 3D pose estimation, but not for 3D model retrieval or other tasks.

\subsection{3D Model Retrieval}
A large number of previous works perform retrieval given a query 3D model~\cite{Shrec2016a,Shrec2017a}. These methods either directly operate on 3D data, \eg, in the form of voxel grids~\cite{Wu20153d,Qi2016volumetric}, spherical maps~\cite{Esteves2018learning}, or point clouds~\cite{Qi2017pointnet}, or process multi-view renderings of the query 3D model~\cite{Chen2003visual,Su2015multi,Bai2016gift,Qi2016volumetric} to compute a shape descriptor. 

However, in this work, we focus on the much more challenging task of 3D model retrieval from a single RGB image~\cite{Kanezaki2018rotationnet}. One approach to address this task is to train a classifier which provides a 3D model for each fine-grained class on top of handcrafted~\cite{Aubry2014seeing} or learned~\cite{Mottaghi2015coarse} features extracted from an RGB image. This, in consequence, restricts the retrieval to 3D models seen during training.

The most popular strategy to overcome this limitation is to map 3D models and RGB images to a common embedding space in which retrieval is performed using distance-based matching~\cite{Kulis2013metric}. In this case, the mapping, the embedding space, and the distance measure can be designed in a variety of ways.

Numerous works match features extracted from an RGB image against features extracted from multi-view RGB renderings to predict both shape and viewpoint. In this context, \cite{Aubry2015understanding} uses a CNN trained for ImageNet classification~\cite{Russakovsky2015imagenet} to extract features. \cite{Massa2016deep} takes a similar approach, but additionally performs nonlinear feature adaption to overcome the domain gap between real and rendered RGB images.  \cite{Huang2018holistic} and \cite{Izadinia2017im2cad} use a CNN trained for object detection as a feature extractor. However, the CNNs used in these methods are not optimized for 3D model retrieval.

Thus, other approaches train mappings to predefined embedding spaces. \cite{Tasse2016shape2vec} trains CNNs to map 3D models, RGB images, depth maps, and sketches to an embedding space based on text for cross-modal retrieval. \cite{Hueting2018see} constructs a low-dimensional embedding space by performing PCA on 3D key points and maps 3D key points predicted using a CNN to that space for retrieval. \cite{Li2015joint} and \cite{Tatarchenko2019a} train a CNN to map RGB images to an embedding space computed from pairwise similarities between 3D models.

Instead of handcrafting an embedding space, an embedding space capturing 3D shape properties can be learned. \cite{Tulsiani2018factoring} reconstructs voxel grids from RGB images of objects using CNNs. The low-dimensional bottle-neck shape descriptor is also used for retrieval. \cite{Girdhar2016learning} combines a 3D voxel encoder and an RGB image encoder with a shared 3D voxel decoder to perform reconstruction from a joint embedding. 3D model retrieval is performed by matching embeddings of voxel grids against those of RGB images.

Finally, recent approaches explicitly learn an embedding space which is optimized for 3D model retrieval. \cite{Xiang2016objectnet3d} uses a single CNN to map RGB images and RGB renderings to an embedding space which is optimized using a Euclidean distance-based lifted structure loss~\cite{Oh2016deep}. At test time, the distances between an embedding of an RGB image and embeddings of multi-view RGB renderings are averaged to compensate for the unknown object pose. \cite{Lee2018cross} uses two CNNs to map RGB images and gray-scale renderings to an embedding space and optimizes a Euclidean distance-based Triplet loss~\cite{Weinberger2009distance}. Additionally, cross-view convolutions are employed to aggregate a sequence of multi-view renderings into a single descriptor to reduce the matching complexity. \cite{Grabner2018a} also trains two CNNs, but maps RGB images and depth maps to a common space. In contrast to other approaches, the 3D pose of the object in the RGB image is explicitly estimated and used in the 3D model retrieval.

Compared to these approaches, we also learn an embedding space which is optimized for 3D model retrieval, but first predict location fields from RGB images and then compute pose invariant 3D shape descriptors from predicted and rendered location fields in an end-to-end trainable way.
\section{Location Field Descriptors}
\label{sec:method}

Given a single RGB image and a 3D model database, we retrieve a 3D model for each object in the image, as shown in Fig.~\ref{fig:overview}. For this purpose, we first generate location fields from 3D models and RGB images. We then compute pose invariant 3D shape descriptors from the locations fields. Finally, we match the descriptors to find the best 3D model.

\subsection{Location Field Generation}

\begin{figure}
	\setlength{\tabcolsep}{1pt}
	\setlength{\fboxsep}{-2pt}
	\setlength{\fboxrule}{2pt}
	\newcommand{\colImgN}[1]{{\includegraphics[width=0.24\linewidth]{#1}}}
	\centering
	\begin{tabular}{cccc}
		\colImgN{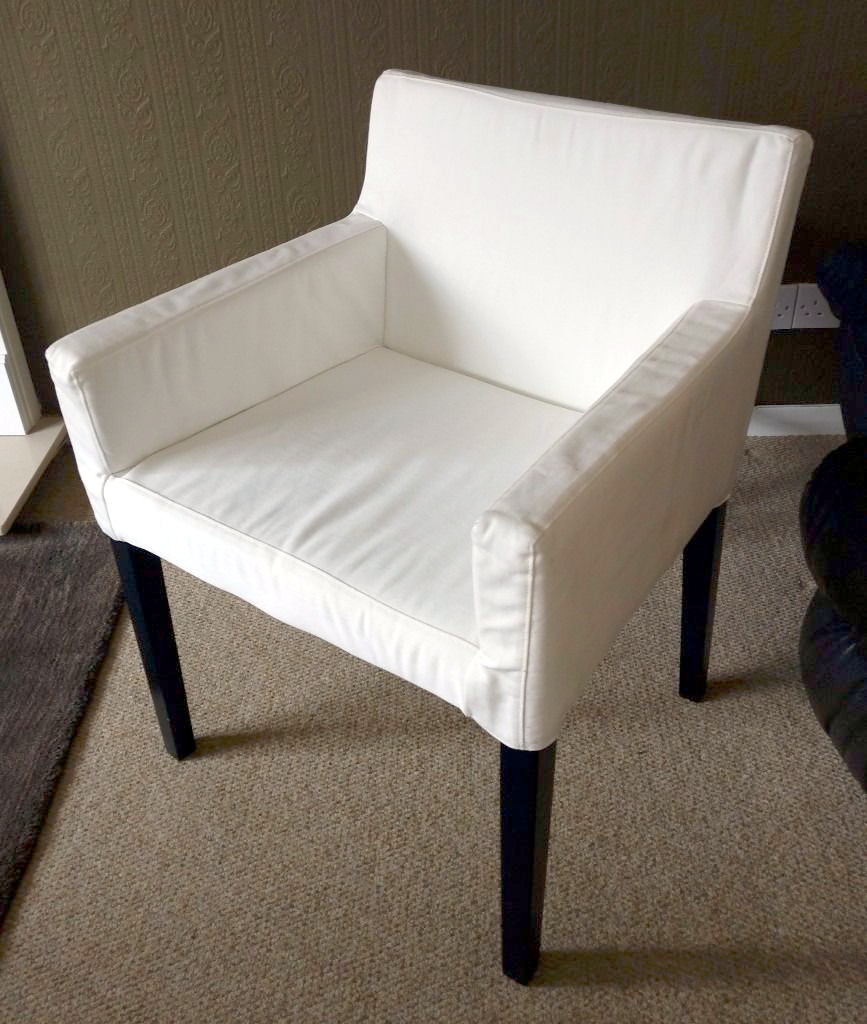}&   \colImgN{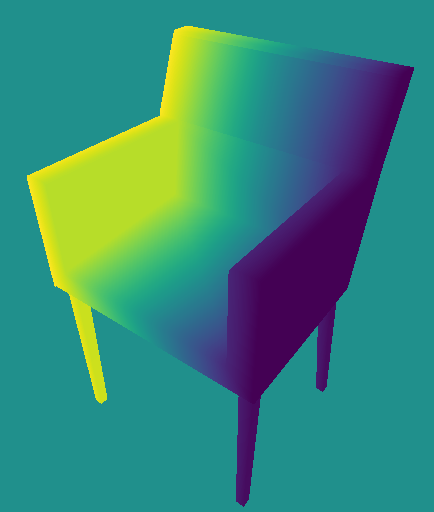}&   \colImgN{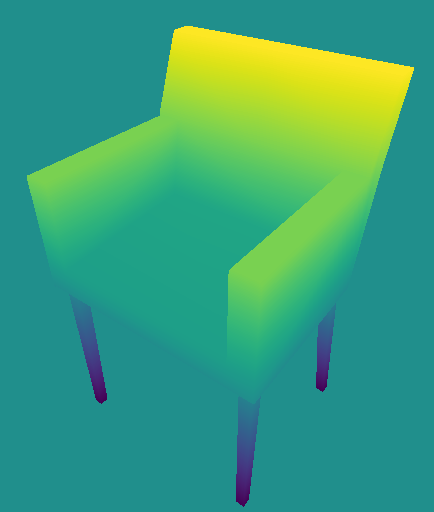}&   \colImgN{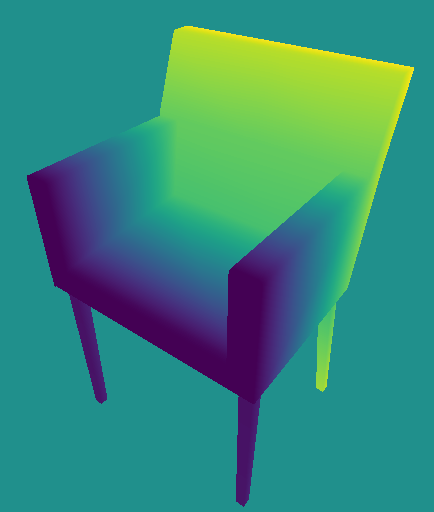}\\[-1.5pt]
		\footnotesize Image&\footnotesize LF (X)&\footnotesize LF (Y)&\footnotesize LF (Z)\\[-3pt]
	\end{tabular}
	\caption{An image of an object and its location field. Location fields encode a 3D surface coordinate in the canonical object coordinate system for each object pixel. We show the three channels which correspond to the X, Y, and Z values of the 3D coordinates in separate images.}
	\label{fig:representations}
\end{figure}

The first step in our approach is to map 3D models and RGB images to a common low-level representation in the form of location fields. As illustrated in Fig.~\ref{fig:representations}, a location field~\cite{Taylor2012vitruvian,Wang2018fine} is an image-like representation that encodes a 3D surface coordinate for each object pixel. Compared to its reference RGB image, a location field has the same size and spatial resolution, but the three channels encode XYZ 3D coordinates in the canonical object coordinate system instead of RGB colors. Locations fields explicitly present 3D shape and 3D pose information, because they encode dense correspondences between 2D pixel locations and 3D surface coordinates. From these 2D-3D correspondences, the 3D pose can be geometrically recovered using a \PNP~algorithm~\cite{Brachmann2016uncertainty,Jafari2018ipose}. Additionally, location fields can also be interpreted as structured partial 3D point clouds. 

Location fields can be directly rendered from 3D models. For this purpose, we rasterize 3D meshes using OpenGL and implement a custom fragment shader which linearly interpolates per-vertex 3D coordinates along the triangles of a 3D mesh. Because the interpolated values describe 3D coordinates in the canonical object coordinate system, the relation to the inherent object orientation is preserved.

\begin{figure*}
	\begin{center}
		\includegraphics[width=\linewidth]{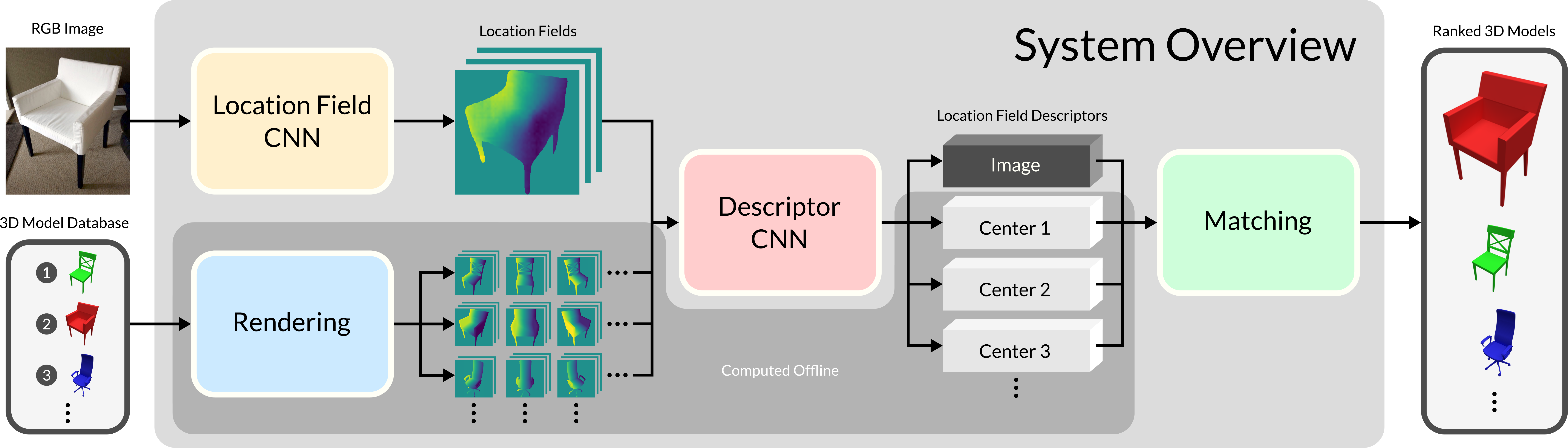}
	\end{center}
	\vspace{-0.25cm}
	\caption{Overview of our approach. Given a single RGB image and a 3D model database, we use CNNs to predict a location field and a pose invariant location field descriptor for each object in the image. For each 3D model in the database, we learn a single center descriptor from multi-view location fields offline during training. Finally, we match location field descriptors predicted from the image against offline computed center descriptors to retrieve a ranked list of the 3D models in the database.}
	\label{fig:overview}
\end{figure*}

In order to generate location fields from RGB images, we need to detect objects in 2D and predict a location field of each object. For this purpose, we introduce a Location Field CNN (see Fig.~\ref{fig:overview}) which extends the generalized Faster/Mask R-CNN framework~\cite{Ren2015faster,He2017mask}. This generic multi-task framework includes a 2D object detection pipeline to perform per-image and per-object computations. In this way, we address multiple different tasks using a single end-to-end trainable network.

In the context of the generalized Faster/Mask R-CNN framework, each output branch provides a task-specific subnetwork with different structure and functionality. We introduce a dedicated output branch for estimating location fields alongside the existing object detection branches~\cite{Wang2018fine}. Similar to the mask branch~\cite{He2017mask}, the location field branch performs region-based per-object computations: For each detected object, an associated spatial region of interest in the feature maps is aligned to a fixed size feature representation with a low spatial but high channel resolution using linear interpolation, \eg, $14\times14\times256$. These aligned features serve as a shared input to the classification, mask and location field branches. Each branch is evaluated $N$ times per image, where $N$ is the number of detected objects. 

Our location field branch uses a fully convolutional subnetwork to predict a tensor of 3D points at a resolution of $56\times56\times3$ from the shared aligned features. We also modify the mask branch to predict 2D masks at the same spatial resolution and use the predicted masks to threshold the tensor of 3D points to get low-resolution location fields. We experimentally found this approach to generate significantly higher accuracy location fields compared to directly regressing low-resolution location fields, which tends to predict over-smoothed 3D coordinates around the object silhouette. During training, we optimize our predicted location fields using the Huber loss~\cite{Huber1964robust}.

The resulting low-resolution location fields can be upscaled and padded to obtain high-resolution location fields with the same spatial resolution as the input image. This is especially helpful in cases where our approach fails. A visual overlay of the input image and the predicted location fields is intuitively interpretable and offers valuable insights to why the system fails.

However, we compute pose invariant descriptors from the low-resolution location fields, because upscaling does not provide additional information but increases the computational workload. Moreover, the predicted low-resolution location fields are tightly localized crops, which reduces the complexity of the descriptor computation.

\subsection{3D Shape Descriptors}
Instead of exhaustively comparing each predicted location field to multiple rendered location fields from different viewpoints~\cite{Massa2016deep,Izadinia2017im2cad}, we map location fields to pose invariant 3D shape descriptors in an embedding space. We refer to descriptors in this space as Location Field Descriptors.

For this purpose, we introduce a Descriptor CNN (see Fig.~\ref{fig:overview}) which utilizes a dense connection pattern~\cite{Huang2017densely}. Similar to ResNets~\cite{He2016deep,He2016identity}, DenseNets~\cite{Huang2017densely} introduce skip-connections in the computational graph, but concatenate feature maps instead of adding them. The dense connection pattern encourages feature reuse throughout the network and leads to compact but expressive models. This architecture is well-suited for computing descriptors from location fields, because they already provide a high level of abstraction. Location fields are not affected by task irrelevant appearance variations caused by color, material, texture or lighting. The object is already segmented from the background and occlusions in the predicted location fields are resolved by the 2D mask used for thresholding the tensor of 3D points. In fact, even the raw 3D coordinates provide useful matching attributes, for example by aligning  query and test point clouds using ICP~\cite{Besl1992method}. Thus, extensive feature reuse within the Descriptor CNN is rational.

In order to learn an embedding space which is optimized for 3D model retrieval, we need to address two requirements. First, the embedding space has to be discriminative in terms of 3D models. Second, the computed descriptors have to be invariant to the 3D pose of the object in the location field. We jointly address both requirements by learning a representative center descriptor for each 3D model, as shown in Fig.~\ref{fig:overview}. For this purpose, we train the Descriptor CNN to map location fields of a 3D model from different viewpoints close to its corresponding center descriptor. At the same time, we make sure that all center descriptors are discriminatively distributed in the embedding space. Thus, during training, we penalize the distances between location field descriptors and center descriptors in a way that each location field descriptor and its corresponding center descriptor are pulled closer together, while all center descriptors are pulled further apart. This approach resembles a nonlinear discriminant analysis~\cite{Santa1998nonlinear}, in which the intra-class variance is minimized, while the inter-class variance is maximized to train more discriminative embeddings.

In particular, we build on the ideas of Center loss~\cite{Wen2016discriminative} and Triplet-Center loss~\cite{He2018triplet} to optimize our embedding space. The Center loss
\begin{equation}
L_{\text{C}} = \sum_{i=1}^N D(f_i,c_{y_i})
\end{equation}
minimizes the distance $D(f_i,c_{y_i})$ between a location field descriptor $f_i$ and its corresponding center descriptor $c_{y_i}$. In this case, $y_i$ is the index of the corresponding 3D model and $N$ denotes the number of samples. For the distance function $D(\cdot)$, we use the Huber distance~\cite{Huber1964robust}. In contrast, the Triplet-Center loss
\begin{equation}
L_{\text{TC}} = \sum_{i=1}^N \max\Big(0~, D(f_i,c_{y_i}) + m - \min_{j\neq y_i}D(f_i,c_{j})\Big)
\end{equation}
enforces the same distance $D(f_i,c_{y_i})$ to be smaller than the distance between a location field descriptor and its closest non-corresponding center descriptor $\min_{j\neq y_i}D(f_i,c_{j})$ by at least the margin $m$.

As a consequence, the Center loss only minimizes intra-class variance, while the Triplet-Center loss aims at both minimizing intra-class variance and maximizing inter-class variance. In many cases, however, the Triplet-Center loss fails to achieve these goals. Instead, it learns degenerated clusterings, because the optimization criterion does not guarantee the desired properties~\cite{Keller2018learning}. Thus, we employ a combination of Center loss and Triplet-Center loss in our Descriptor loss
\begin{equation}
L_{\text{D}} = L_{\text{softmax}} + \alpha L_{\text{C}} + \beta L_{\text{TC}}
\label{eq:descriptor_loss}
\end{equation}
to achieve both low intra-class variance and high inter-class variance~\cite{Wohlhart2015CVPR,Keller2018learning}. In practice, these losses are combined with a softmax loss $L_{\text{softmax}}$ to learn more discriminative embeddings than classification alone~\cite{Wen2016discriminative,He2018triplet}. The parameters $\alpha$ and $\beta$ control the impact of the different loss terms.

We want to emphasize that the center descriptors are not fixed, but learned during training. In fact, the center descriptors are trainable weights of the Descriptor CNN in our implementation. Also, even though we optimize a triplet criterion, we do not require location field triplets as training input. Only a single location field and its corresponding 3D model index $y_i$ are needed. The center descriptors required for the Triplet loss are sampled within the Descriptor CNN. Additionally, hard triplet mining~\cite{Schroff2015facenet} is less important, because we always sample the closest non-corresponding center descriptor and also employ Center and softmax losses.

We jointly train the Descriptor CNN on predicted and rendered location fields. This is a major advantage compared to previous approaches that directly map to an embedding space, because training data in the form of RGB images with 3D model annotations is limited. In contrast, we benefit from training on a virtually infinite amount of synthetic data. Additionally, the intermediate location field prediction serves as a regularizing bottleneck and reduces the risk of overfitting, because regressing location fields is more difficult than computing embeddings.

Since there is a domain gap between predicted and rendered location fields, we perform Feature Mapping~\cite{Rad2018feature}. We use a residual block~\cite{He2016deep,He2016identity} to map location field descriptors from the predicted to the rendered domain. Thus, the training input either consists of pairs of corresponding predicted and rendered location fields or single location fields, and a 3D model index $y_i$ in both cases. In the case of pairs, we compute an additional Feature Mapping loss between corresponding feature-mapped predicted and rendered location field descriptor using the Huber distance~\cite{Huber1964robust}.

To perform retrieval from a previously unseen 3D model database, we generate center descriptors without retraining. For each unseen 3D model, we render 100 location fields under different 3D poses, compute their embeddings using the Descriptor CNN and average them to obtain a new center descriptor. Alternatively, we can retrain the Descriptor CNN by incorporating the new 3D models as additional rendered location fields. In any case, the center descriptors are computed offline which results in fast inference.

During inference, we only need to process RGB images using our CNNs, since the center descriptors have already been computed offline. For each RGB image, we evaluate the Location Field CNN once and the Descriptor CNN $N$ times to compute location field descriptors, where $N$ is the number of detected objects. We then match each computed location field descriptors against all center descriptors and generate a ranked list of 3D models based on the Euclidean distance between the descriptors, as shown in Fig.~\ref{fig:overview}.

Finally, the entire system, \ie, the Location Field CNN and the Descriptor CNN, is end-to-end trainable. The system loss is a combination of our Location Field loss, our Descriptor loss, our Feature Mapping loss, and the Detection losses of the generalized Faster/Mask R-CNN framework.
\section{Experimental Results}
\label{sec:experiments}

To demonstrate the benefits of Location Field Descriptors, we evaluate our approach on three challenging real-world datasets with different object categories: Pix3D~\cite{Sun2018pix3d} (\textit{bed}, \textit{chair}, \textit{sofa}, \textit{table}), Comp~\cite{Wang2018fine} (\textit{car}), and Stanford~\cite{Wang2018fine} (\textit{car}). Details on the datasets, the implementation, and the evaluation are provided in the {\bf supplementary material}.

In particular, we provide quantitative results for 3D model retrieval from seen and unseen databases in comparison to the state-of-the-art in Sec.~\ref{sec:sota}, present qualitative results of our approach in Sec.~\ref{sec:qualitative-results}, and perform an ablation study in Sec.~\ref{sec:ablation}. For our quantitative evaluation, we use the following well-established metrics:

\definecolor{lightgreen}{RGB}{200,240,217}
\definecolor{lightred}{RGB}{240,200,200}
\begin{table*}
	\centering
	\setlength{\tabcolsep}{9pt}
	\begin{tabular}{lcc|c|cccc|cc}
		\toprule
		\multicolumn{4}{c}{}&\multicolumn{4}{c}{\bf seen 3D models}&\multicolumn{2}{c}{\bf unseen 3D models}\\
		\cmidrule(lr){5-8}\cmidrule(lr){9-10}
		Method&Dataset&\multicolumn{1}{c}{Category}
		&\multicolumn{1}{c}{$Acc_{D_{0.5}}$}
		&\multicolumn{1}{c}{$Acc_{\textit{Top-1}}$}
		&\multicolumn{1}{c}{$Acc_{\textit{Top-10}}$}
		&\multicolumn{1}{c}{$d_{\text{HAU}}$}
		&\multicolumn{1}{c}{$d_{\text{IOU}}$}
		&\multicolumn{1}{c}{$d_{\text{HAU}}$}
		&\multicolumn{1}{c}{$d_{\text{IOU}}$}\\
		\midrule
		\cite{Aubry2015understanding}&\multirow{3}{*}{Pix3D}&\multirow{3}{*}{bed}&\multirow{3}{*}{99.0\%}&19.4\%&46.6\%&0.0821&0.3397&0.0960&0.2487\\
		\cite{Grabner2018a}&&&&35.1\%&83.2\%&0.0385&0.5598&0.0577&0.3013\\
		Ours&&&&\bf64.4\%&\bf89.0\%&\bf0.0152&\bf0.8074&\bf0.0448&\bf0.3490\\
		\midrule
		\cite{Aubry2015understanding}&\multirow{3}{*}{Pix3D}&\multirow{3}{*}{chair}&\multirow{3}{*}{91.5\%}&17.3\%&49.1\%&0.0559&0.3027&0.0843&0.1334\\
		\cite{Grabner2018a}&&&&41.3\%&73.9\%&0.0305&0.5469&0.0502&0.1965\\
		Ours&&&&\bf58.1\%&\bf81.8\%&\bf0.0170&\bf0.7169&\bf0.0375&\bf0.2843\\
		\midrule
		\cite{Aubry2015understanding}&\multirow{3}{*}{Pix3D}&\multirow{3}{*}{sofa}&\multirow{3}{*}{96.9\%}&21.7\%&52.2\%&0.0503&0.3824&0.0590&0.3493\\
		\cite{Grabner2018a}&&&&44.1\%&89.8\%&0.0197&0.7762&0.0294&0.6178\\
		Ours&&&&\bf67.0\%&\bf94.4\%&\bf0.0075&\bf0.9028&\bf0.0178&\bf0.7472\\
		\midrule
		\cite{Aubry2015understanding}&\multirow{3}{*}{Pix3D}&\multirow{3}{*}{table}&\multirow{3}{*}{91.2\%}&12.0\%&34.2\%&0.1003&0.1715&0.1239&0.1047\\
		\cite{Grabner2018a}&&&&33.9\%&66.1\%&0.0607&0.4500&0.0753&0.1730\\
		Ours&&&&\bf53.3\%&\bf80.1\%&\bf0.0288&\bf0.6383&\bf0.0482&\bf0.2573\\
		\midrule
		\cite{Aubry2015understanding}&\multirow{3}{*}{Pix3D}&\multirow{3}{*}{$mean$}&\multirow{3}{*}{94.6\%}&17.6\%&45.5\%&0.0722&0.2991&0.0908&0.2090\\
		\cite{Grabner2018a}&&&&38.6\%&78.3\%&0.0374&0.5832&0.0531&0.3222\\
		Ours&&&&\cellcolor{lightgreen}\bf60.7\%&\cellcolor{lightgreen}\bf86.3\%&\cellcolor{lightgreen}\bf0.0171&\cellcolor{lightgreen}\bf0.7663&\cellcolor{lightgreen}\bf0.0370&\cellcolor{lightgreen}\bf0.4095\\
		\midrule
		\midrule
		\cite{Aubry2015understanding}&\multirow{3}{*}{Comp}&\multirow{3}{*}{car}&\multirow{3}{*}{99.9\%}&2.4\%&18.2\%&0.0207&0.7224&0.0271&0.6344\\
		\cite{Grabner2018a}&&&&10.2\%&36.9\%&0.0158&0.7805&0.0194&0.7230\\
		Ours&&&&\cellcolor{lightgreen}\bf20.5\%&\cellcolor{lightgreen}\bf58.0\%&\cellcolor{lightgreen}\bf0.0133&\cellcolor{lightgreen}\bf0.8142&\cellcolor{lightgreen}\bf0.0165&\cellcolor{lightgreen}\bf0.7707\\
		\midrule
		\midrule
		\cite{Aubry2015understanding}&\multirow{3}{*}{Stanford}&\multirow{3}{*}{car}&\multirow{3}{*}{99.6\%}&3.7\%&20.1\%&0.0198&0.7169&0.0242&0.6526\\
		\cite{Grabner2018a}&&&&11.3\%&42.2\%&0.0153&0.7721&0.0183&0.7201\\
		Ours&&&&\cellcolor{lightgreen}\bf29.5\%&\cellcolor{lightgreen}\bf69.4\%&\cellcolor{lightgreen}\bf0.0110&\cellcolor{lightgreen}\bf0.8352&\cellcolor{lightgreen}\bf0.0150&\cellcolor{lightgreen}\bf0.7744\\
		\bottomrule
	\end{tabular}
	\caption{Experimental results on the Pix3D, Comp, and Stanford datasets. We provide results for 3D model retrieval from both seen (in training dataset) and unseen (ShapeNet) 3D model databases given unseen test images. We significantly outperform the state-of-the-art in all metrics and datasets. A detailed discussion of the reported numbers is presented in Sec.~\ref{sec:sota}.}
	\label{table:pix3d}
\end{table*}

\vspace{0.15cm}\noindent\textbf{Detection.} 
We report the detection accuracy $Acc_{D_{0.5}}$ which gives the percentage of objects for which the intersection over union between the ground truth 2D bounding box and the predicted 2D bounding box is larger than 50\%~\cite{Xiang2014beyond}. This metric is an upper bound for other $Acc$ metrics since we do not make blind predictions.

\vspace{0.15cm}\noindent\textbf{Retrieval Accuracy.} We evaluate the retrieval accuracies $Acc_{\textit{Top-1}}$ and $Acc_{\textit{Top-10}}$ which give the percentage of objects for which the ground truth 3D model equals the top ranked 3D model (\textit{Top-1})~\cite{Grabner2018a}, or is in the top ten ranked 3D models (\textit{Top-10})~\cite{Xiang2016objectnet3d}. These metrics can only be provided if the ground truth 3D model is in the retrieval database.

\vspace{0.15cm}\noindent\textbf{Hausdorff Distance.} We compute a modified Hausdorff distance~\cite{Aspert2002mesh,Bai2015neural}
\begin{equation}
d_{\text{H}}=\frac{1}{|\mathcal{X}|+|\mathcal{Y}|} \Big(\sum_{\boldsymbol{x}\in\mathcal{X}} \min_{\boldsymbol{y}\in\mathcal{Y}} D(\boldsymbol{x},\boldsymbol{y}) + \sum_{\boldsymbol{y}\in\mathcal{Y}} \min_{\boldsymbol{x}\in\mathcal{X}} D(\boldsymbol{y},\boldsymbol{x}) \Big)
\end{equation}
between the ground truth 3D model $\mathcal{X}$ and the retrieved 3D model $\mathcal{Y}$. For each vertex $\boldsymbol{x}\in\mathcal{X}$ and $\boldsymbol{y}\in\mathcal{Y}$ , we calculate the Euclidean distance $D(\cdot)$ to the closest vertex from the other 3D model and compute the mean over both sets. Before computing $d_{\text{H}}$, we regularly resample each 3D model. We report the mean modified Hausdorff distance for all detected objects ($d_{\text{HAU}}$). Since all 3D models are consistently aligned, the score is in the interval $[0,\sqrt{2}]$ (lower is better).

\vspace{0.15cm}\noindent\textbf{3D Intersection Over Union.} We compute the 3D intersection over union between a voxelization of the ground truth 3D model and a voxelization of the retrieved 3D model~\cite{Tatarchenko2019a}. For this purpose, we voxelize 3D models using \texttt{binvox}~\cite{nooruddin03} with a resolution of $128\times128\times128$. We report the mean 3D IOU for all detected objects ($d_{\text{IOU}}$). The score is in the interval $[0,1]$ (higher is better).

\begin{figure*}
	\setlength{\tabcolsep}{0pt}
	\newcommand{\colImgN}[1]{{\includegraphics[width=0.08333\linewidth]{#1}}}
	\newcommand{\colImgI}[1]{{\includegraphics[height=0.08333\linewidth]{#1}}}
	\centering
	\begin{tabular}{cc|cccccccccc}
		\colImgI{}&   \colImgN{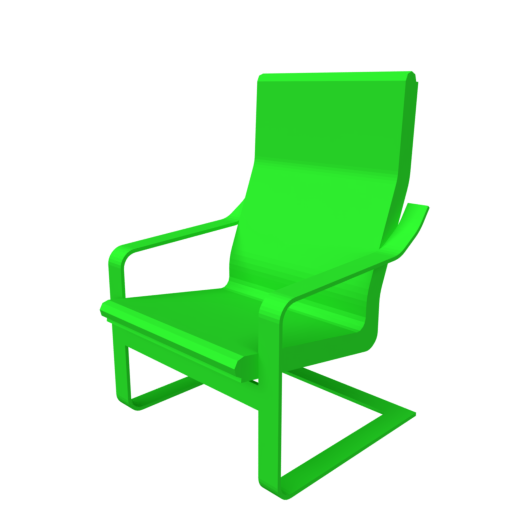}&
		\colImgN{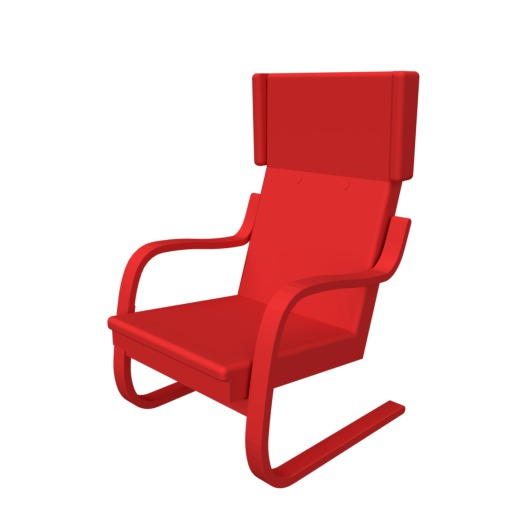}&
		\colImgN{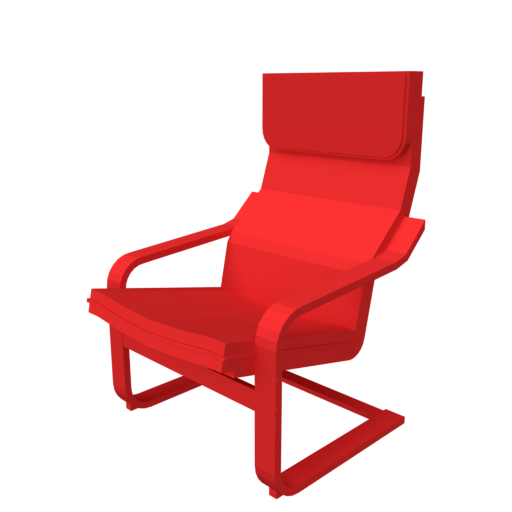}&
		\colImgN{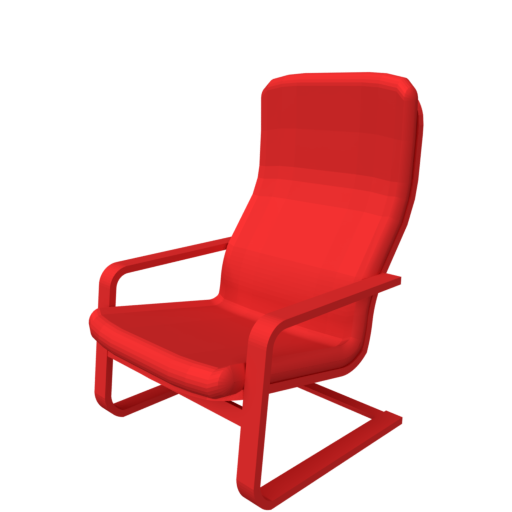}&
		\colImgN{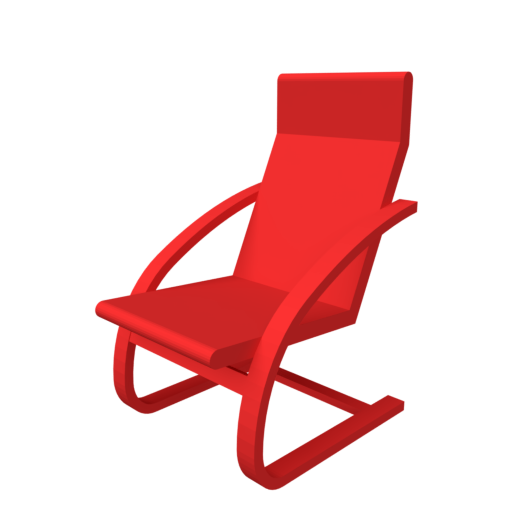}&
		\colImgN{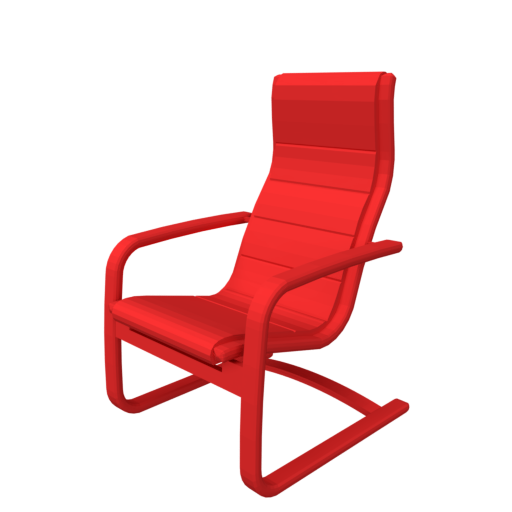}&
		\colImgN{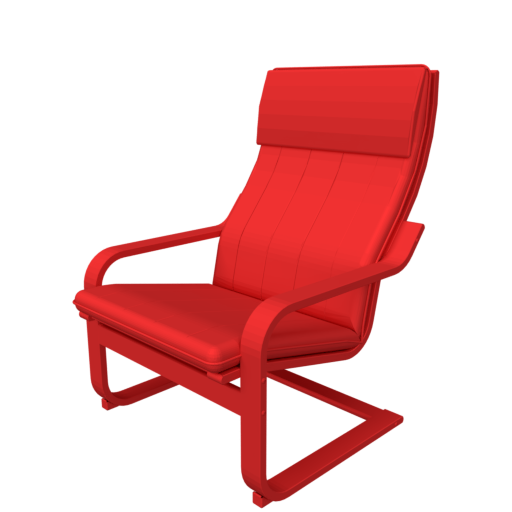}&
		\colImgN{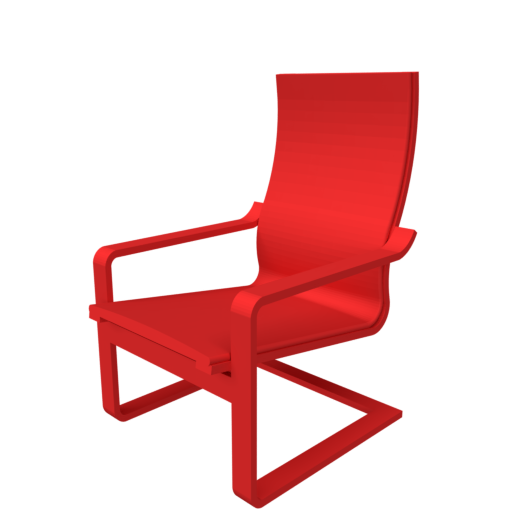}&
		\colImgN{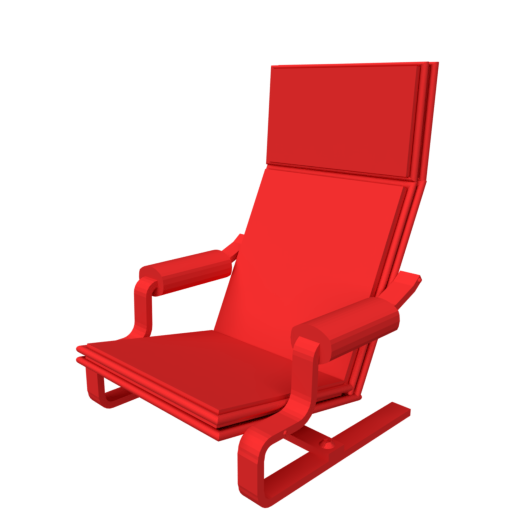}&
		\colImgN{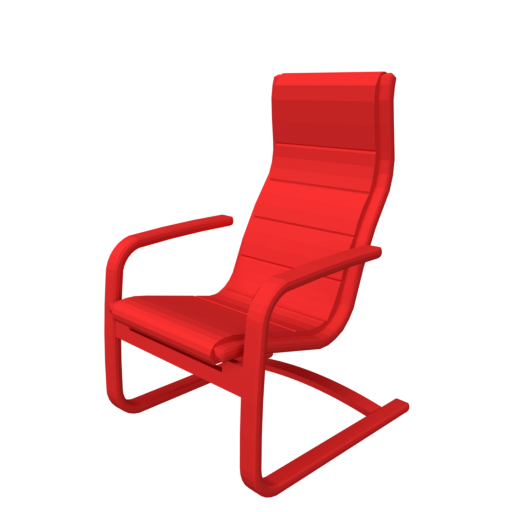}&
		\colImgN{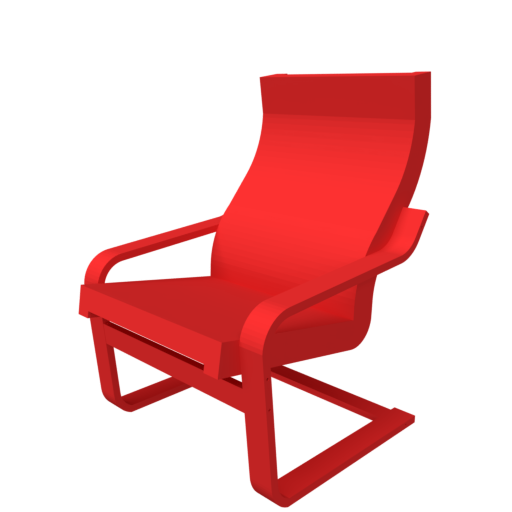}\\[-7pt]
		\colImgN{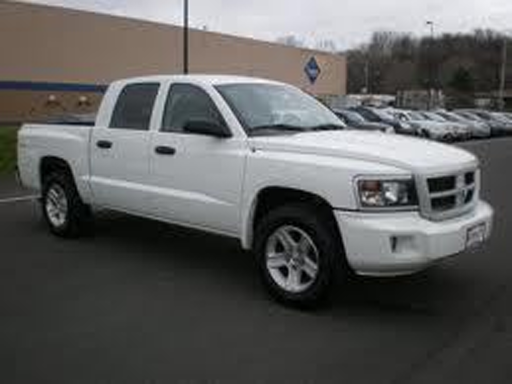}&   \colImgN{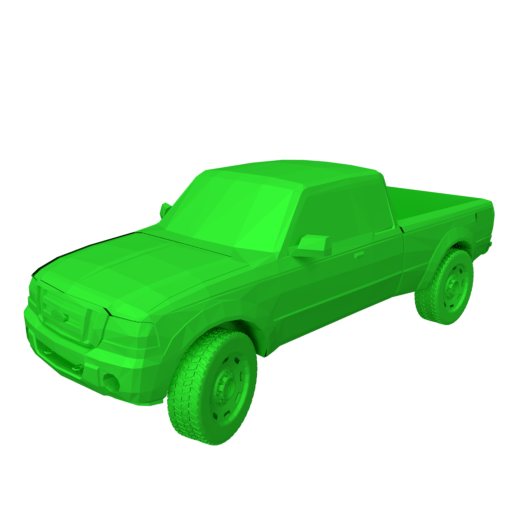}&
		\colImgN{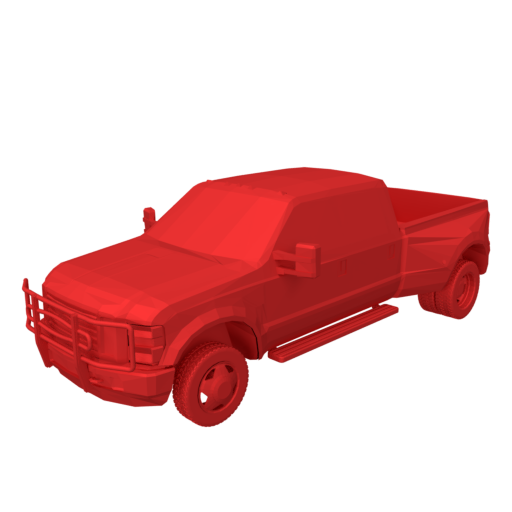}&
		\colImgN{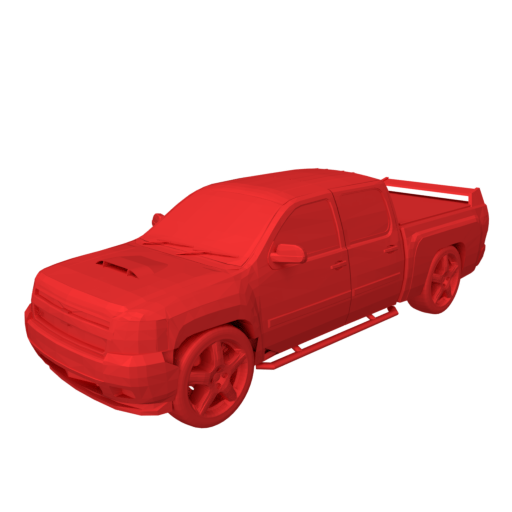}&
		\colImgN{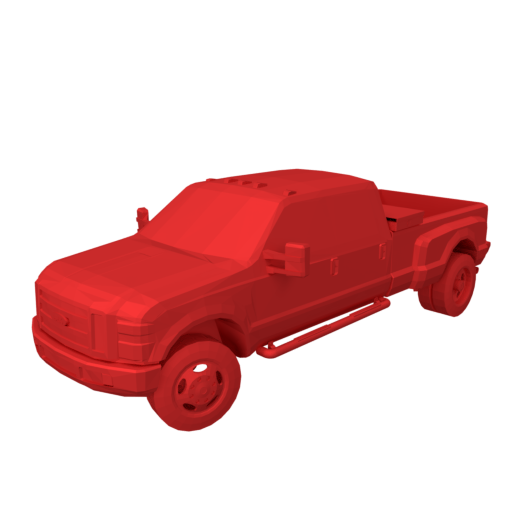}&
		\colImgN{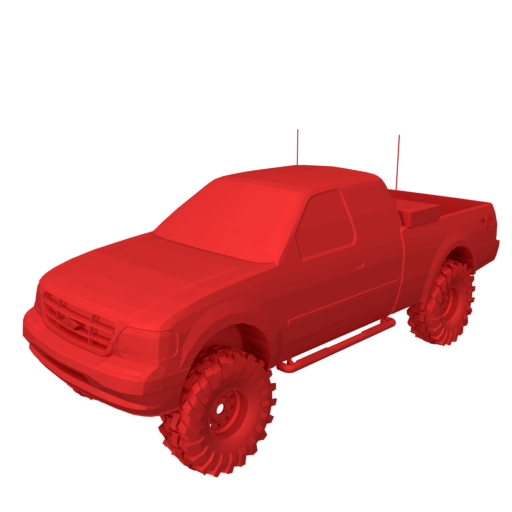}&
		\colImgN{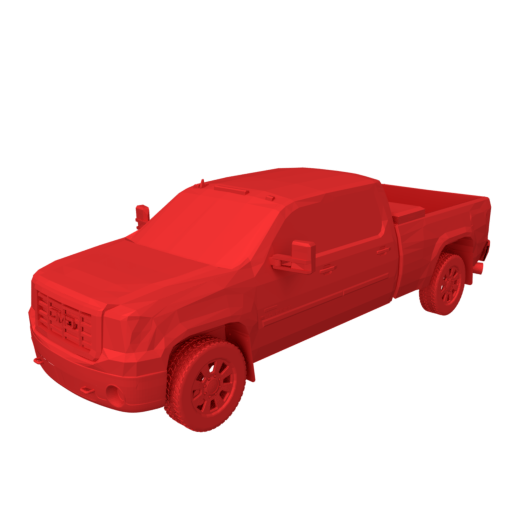}&
		\colImgN{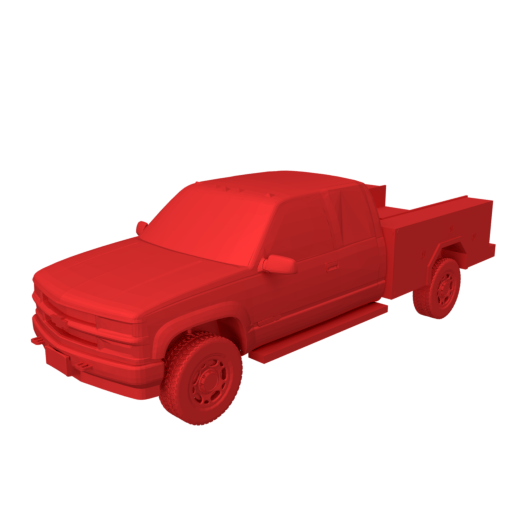}&
		\colImgN{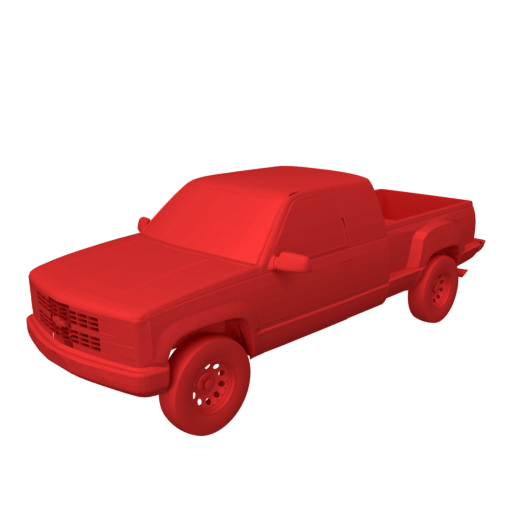}&
		\colImgN{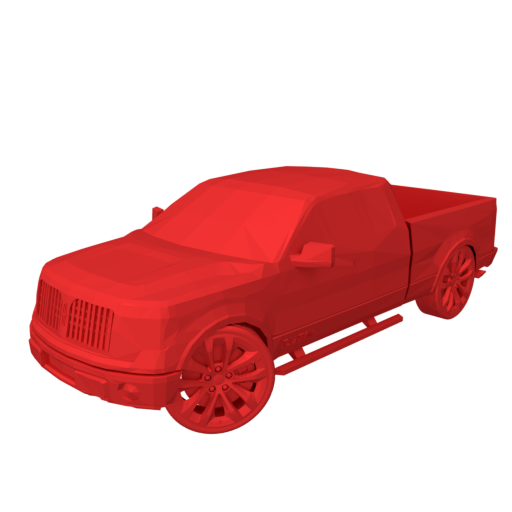}&
		\colImgN{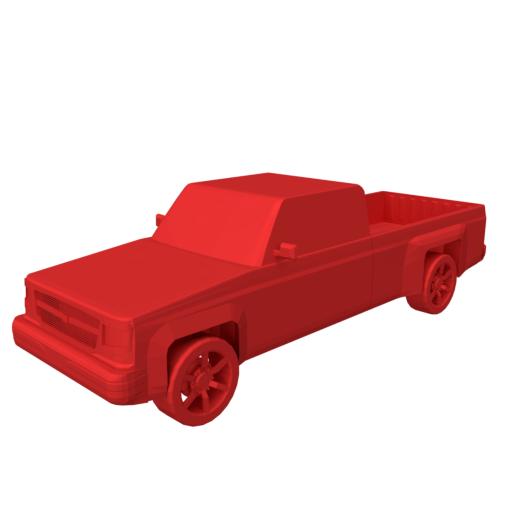}&
		\colImgN{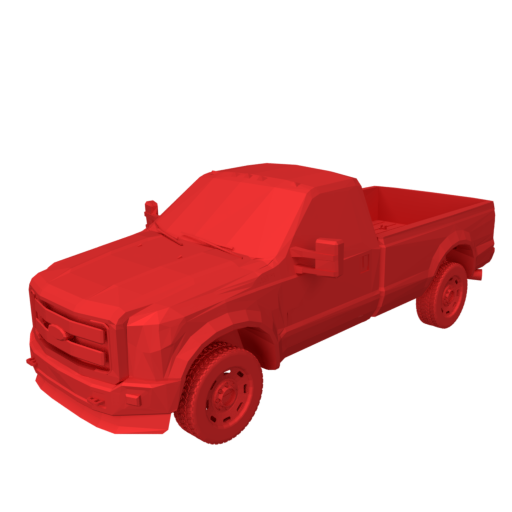}\\[-1.5pt]
		\footnotesize Image&\footnotesize GT&\footnotesize 1&\footnotesize 2&\footnotesize 3&\footnotesize 4&\footnotesize 5&\footnotesize 6&\footnotesize 7&\footnotesize 8&\footnotesize 9&\footnotesize 10\\[-3pt]
	\end{tabular}
	\caption{Qualitative results for 3D model retrieval from ShapeNet. From left to right, we show the input image, the ground truth 3D model and the top ten ranked 3D models. The overall 3D shape of the retrieved models is consistent and accurate.}
	\label{fig:top10}
\end{figure*}

\begin{figure}
	\setlength{\tabcolsep}{1pt}
	\setlength{\fboxsep}{-2pt}
	\setlength{\fboxrule}{2pt}
	\newcommand{\colImgN}[1]{{\includegraphics[width=0.24\linewidth]{#1}}}
	\centering
	\begin{tabular}{cccc}
		\colImgN{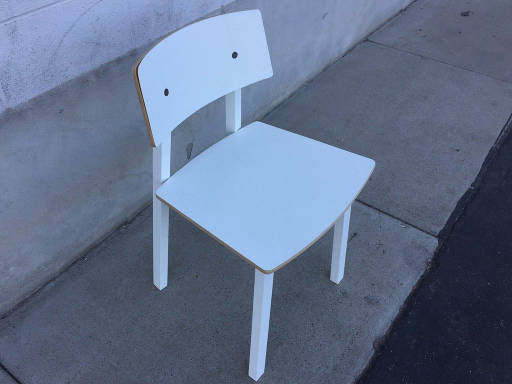}&   \colImgN{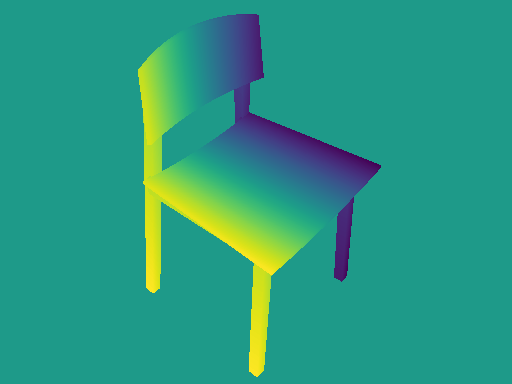}&   \colImgN{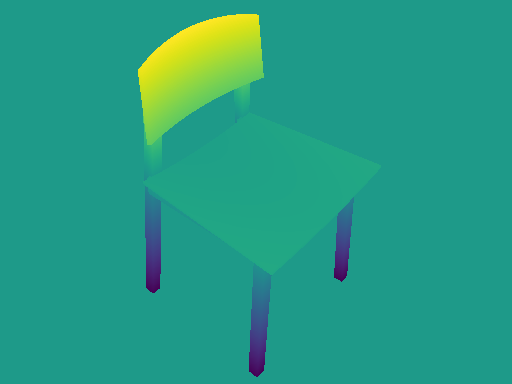}&   \colImgN{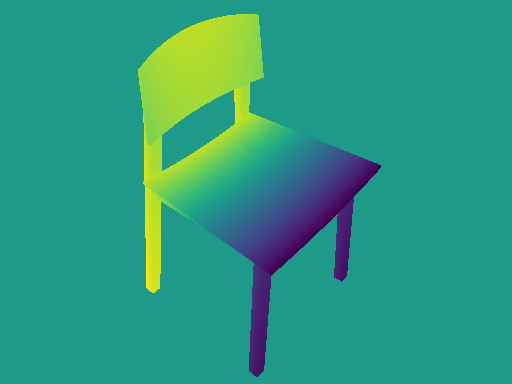}\\[-1.5pt]
		&\colImgN{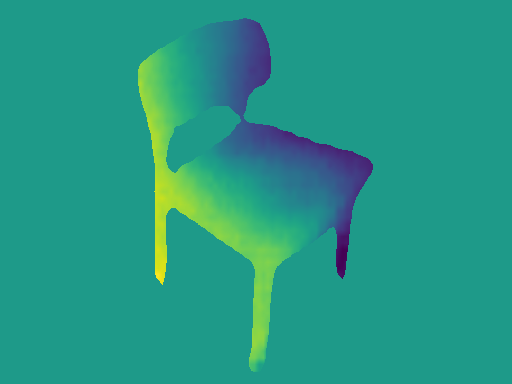}&   \colImgN{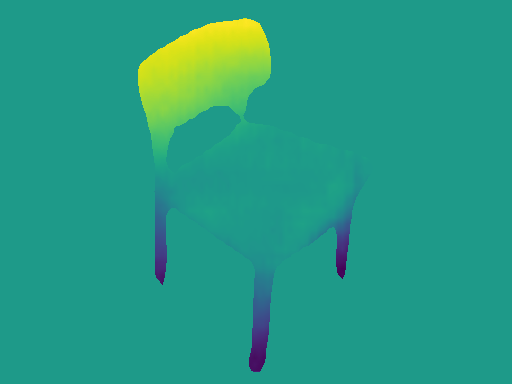}&   \colImgN{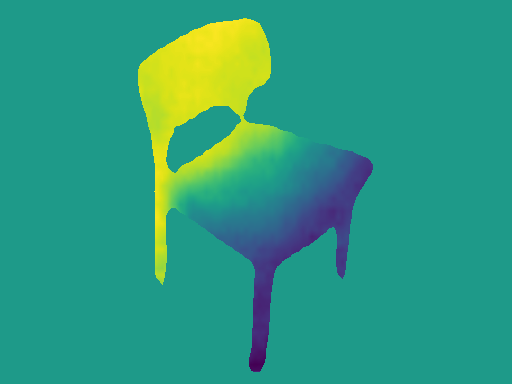}\\[-1.5pt]
		
		\colImgN{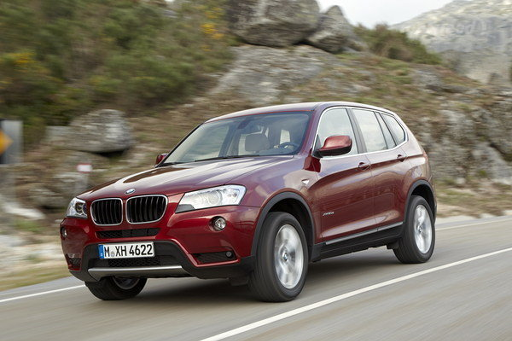}&   \colImgN{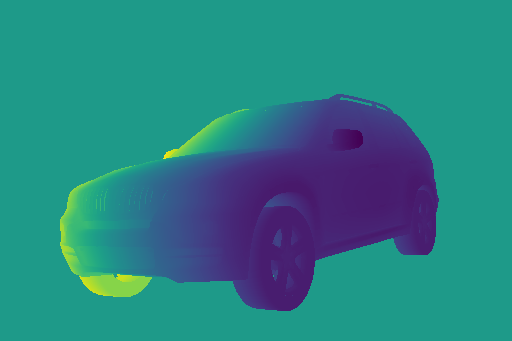}&   \colImgN{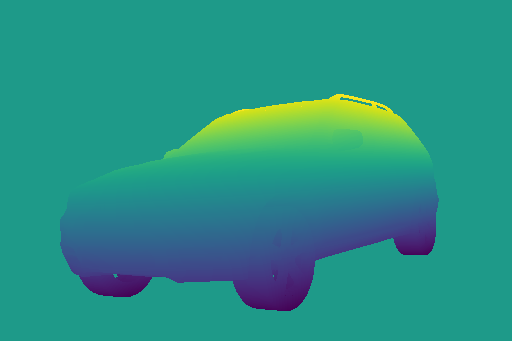}&   \colImgN{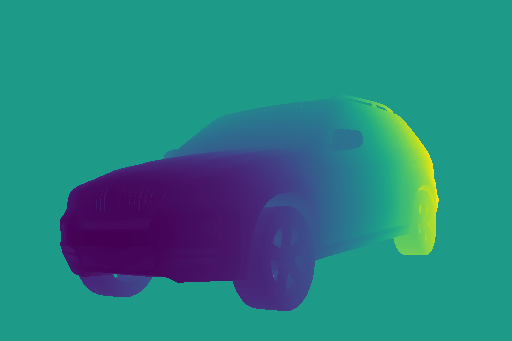}\\[-1.5pt]
		&\colImgN{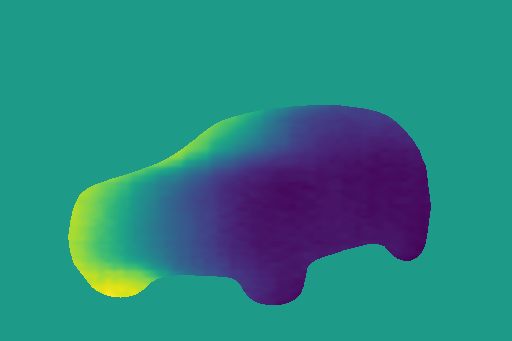}&   \colImgN{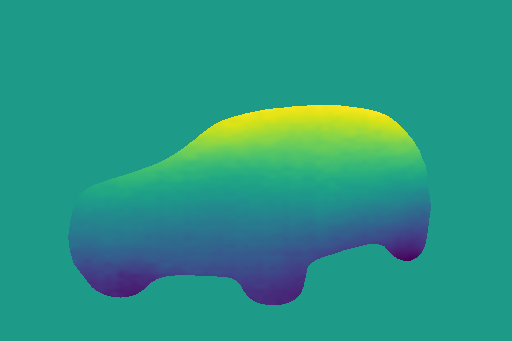}&   \colImgN{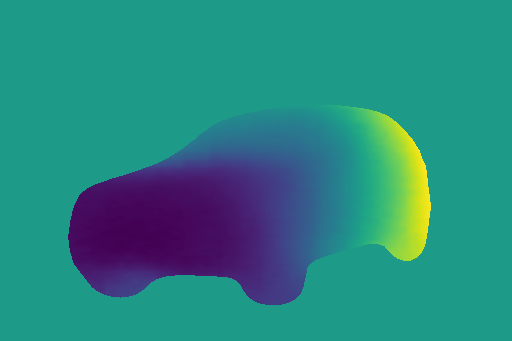}\\[-1.5pt]
		
		\footnotesize Image&\footnotesize LF (X)&\footnotesize LF (Y)&\footnotesize LF (Z)\\[-3pt]
	\end{tabular}
	\caption{Qualitative examples of our predicted location fields. For each example image, the top row shows the ground truth and the bottom row shows our prediction. The overall 3D shape is recovered well, but fine-grained details like the side mirrors of the car are missed.}
	\label{fig:pred_lfs}
\end{figure}

\subsection{Comparison to the State-of-the-Art}
\label{sec:sota}

We are the first to present results for 3D model retrieval on Pix3D, Comp, and Stanford. For this purpose, we compare our approach to a baseline method~\cite{Aubry2015understanding} and a state-of-the-art method~\cite{Grabner2018a}. Since~\cite{Aubry2015understanding} and~\cite{Grabner2018a} assume that objects are already detected in 2D, we use the detections given by our approach for a fair comparison. The results are summarized in Table~\ref{table:pix3d}, where we significantly outperform the state-of-the-art in all metrics and datasets.

First of all, we correctly detect 95\% of all objects in the images on average ($Acc_{D_{0.5}}$), since object detection is tightly integrated into our approach. In fact, our Location Field CNN is initialized with weights trained for instance segmentation~\cite{He2017mask} on COCO~\cite{Lin2014microsoft} and all evaluated categories are present in COCO.

Next, we evaluate two different retrieval setups: First, we use all 3D models from the respective dataset as a 3D model database for retrieval (\textit{seen 3D models}). In this case, we retrieve the correct 3D model ($Acc_{\textit{Top-1}}$) for more than 60\% of all test samples on average on Pix3D. This is a significant improvement of more than {\bf 20\% absolute} compared to the state-of-the-art. Also, the retrieval accuracy quickly raises if we consider the top ten ranked 3D models ($Acc_{\textit{Top-10}}$).

In contrast, the retrieval accuracy on Comp and Stanford is significantly lower for all evaluated methods. This is due to the significantly smaller variation in the overall shape of \textit{cars} compared to \textit{chairs}, for example. Thus, many 3D models of \textit{cars} have a similar appearance in multiple 3D poses and can only be discriminated by extremely fine-grained details like wheel rims or radiator grill structure. Such pixel-level information is usually discarded by CNNs.

However, by analyzing the mesh similarity between the ground truth 3D model and the top retrieved 3D model ($d_{\text{HAU}}$ and $d_{\text{IOU}}$), we observe consistent high performance across all datasets and categories. To put the reported numbers in perspective, we compute the mean of the modified Hausdorff distance ($0.1236$) and the 3D IOU ($0.0772$) for all pairs of 3D models in the training datasets. These numbers represent the accuracy for picking a random 3D model. For both metrics, the mesh similarity of our retrieved 3D model is around {\bf 10 times} better compared to picking a random 3D model. Additionally, we significantly outperform the state-of-the-art by up to {\bf 50\% relative} considering $d_{\text{HAU}}$.

Second, we perform retrieval from previously unseen 3D models from ShapeNet~\cite{Shapenet2015} (\textit{unseen 3D models}). Since the correct 3D model is not in the database in this case, the achievable performance is limited. Thus, the reported numbers are slightly worse compared to retrieval from previously seen 3D models. Still, the performance is much better compared to picking a random 3D model. In fact, for some categories, \eg, Stanford \textit{cars}, our approach retrieves more accurate 3D models from an {\bf unseen} database than the state-of-the-art from a database {\bf seen} during training. 

\subsection{Qualitative Results}
\label{sec:qualitative-results}

The quantitative performance of our approach is also reflected in our qualitative results. First, Fig.~\ref{fig:pred_lfs} shows examples of our predicted location fields. We upscale and pad the predicted location fields to match the input image resolution. The overall 3D shape is recovered well in the location fields, but fine-grained details like the side mirrors of the car are missed.

Next, Fig.~\ref{fig:top10} shows qualitative results for 3D model retrieval from ShapeNet. Considering the top ten ranked 3D models, we observe that the retrieved models have a consistent and accurate overall 3D shape and geometry. 

\begin{figure}
	\setlength{\tabcolsep}{1pt}
	\setlength{\fboxsep}{-2pt}
	\setlength{\fboxrule}{2pt}
	\newcommand{\colImgN}[1]{{\includegraphics[width=0.24\linewidth]{#1}}}
	\centering
	\begin{tabular}{cccc}
		
		\colImgN{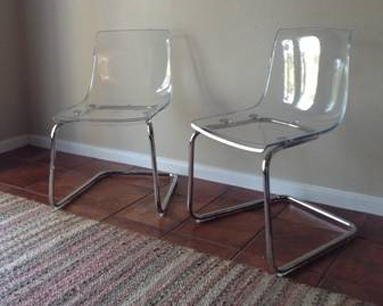}&   \colImgN{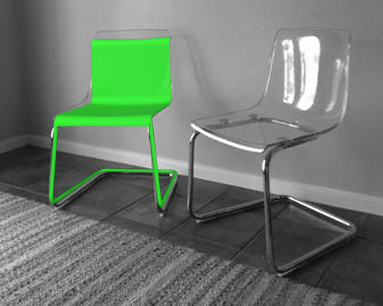}&
		\colImgN{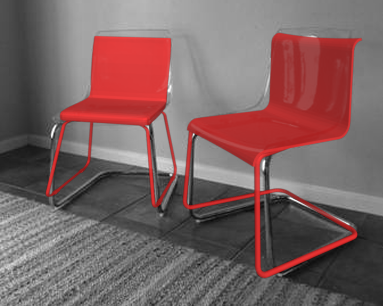}&  \colImgN{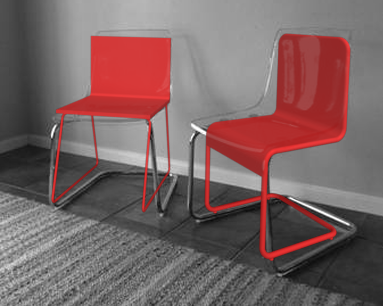}\\[-1.5pt]
		
		\colImgN{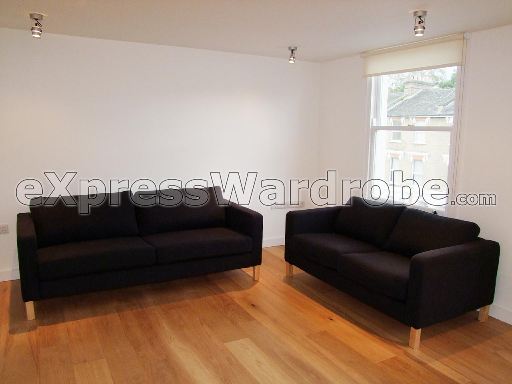}&   \colImgN{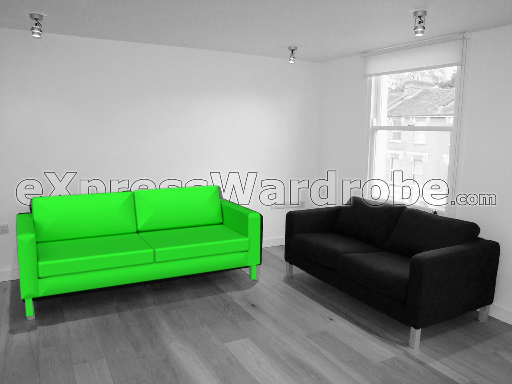}&
		\colImgN{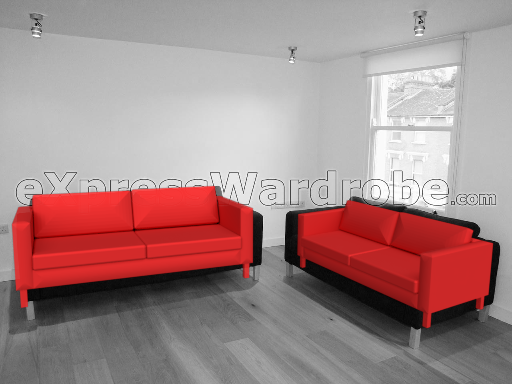}&  \colImgN{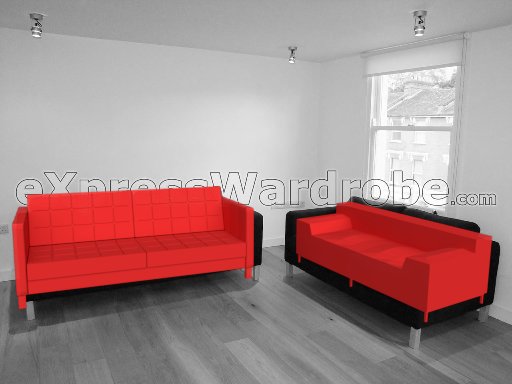}\\[-1.5pt]
		
		\colImgN{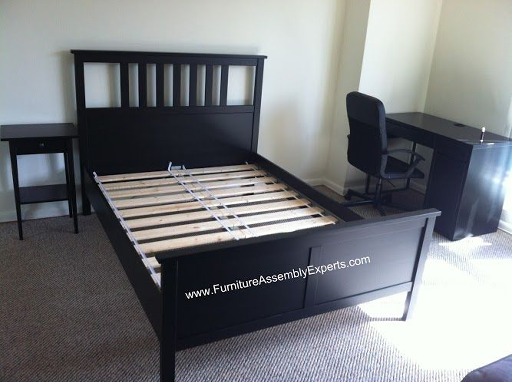}&   \colImgN{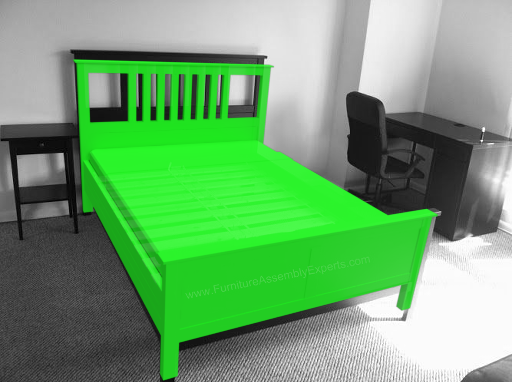}&
		\colImgN{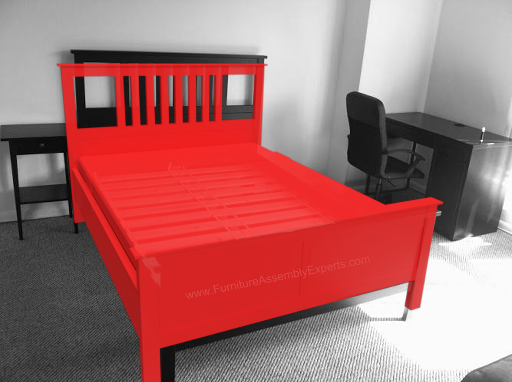}&  \colImgN{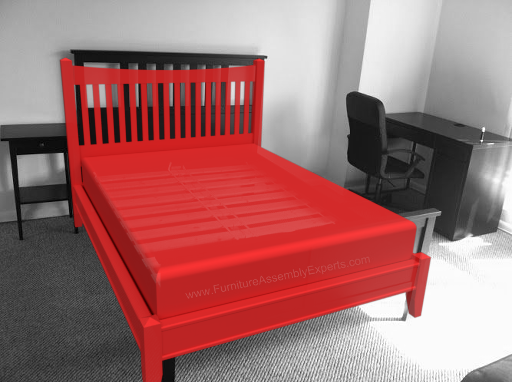}\\[-1.5pt]
		
		\colImgN{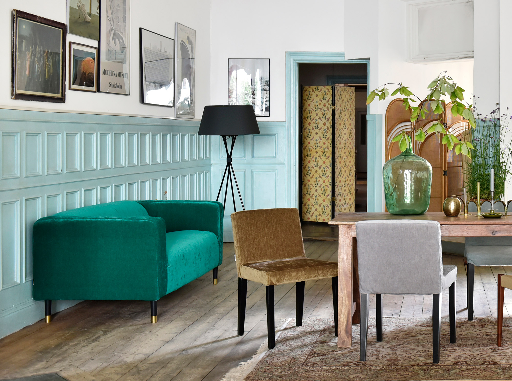}&   \colImgN{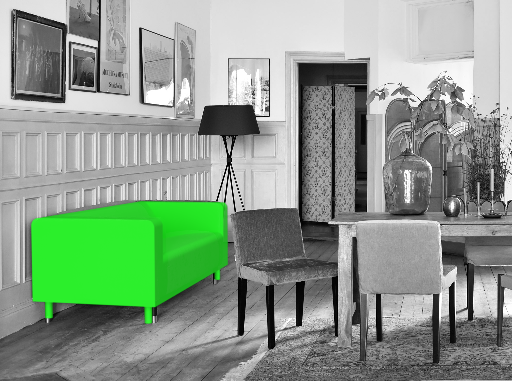}&
		\colImgN{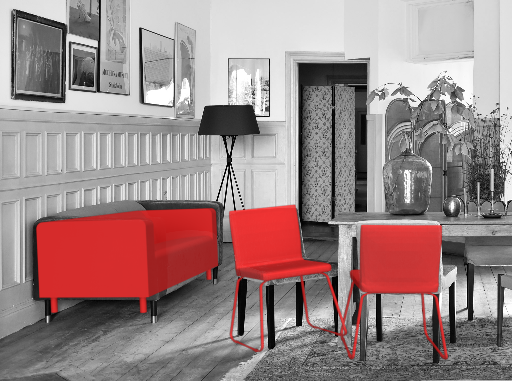}&  \colImgN{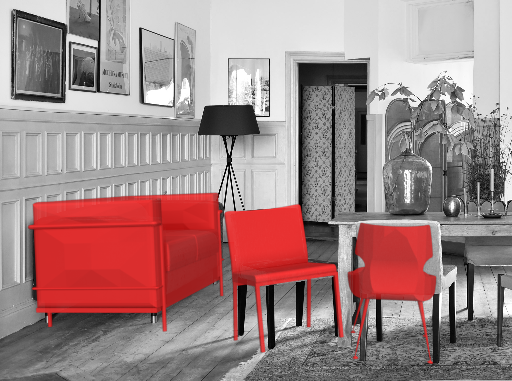}\\[-1.5pt]
		
		\colImgN{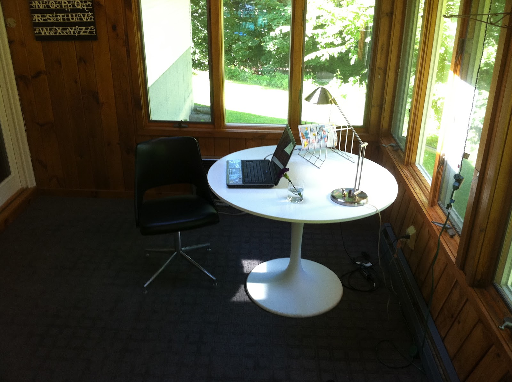}&   \colImgN{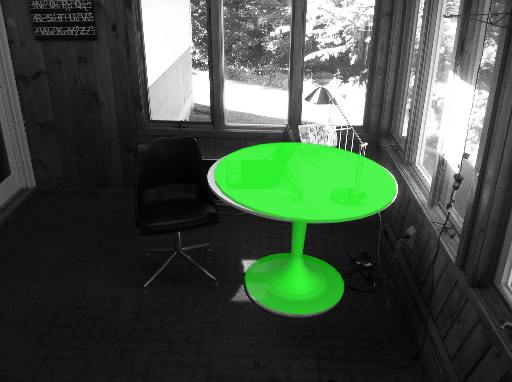}&
		\colImgN{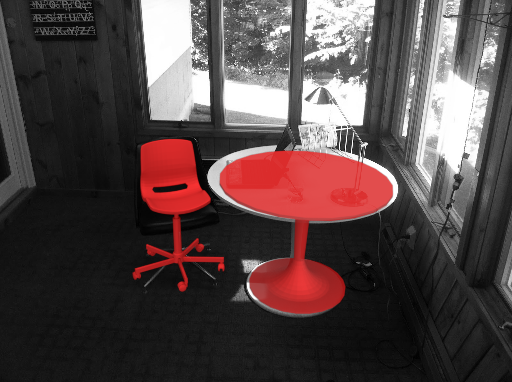}&  \colImgN{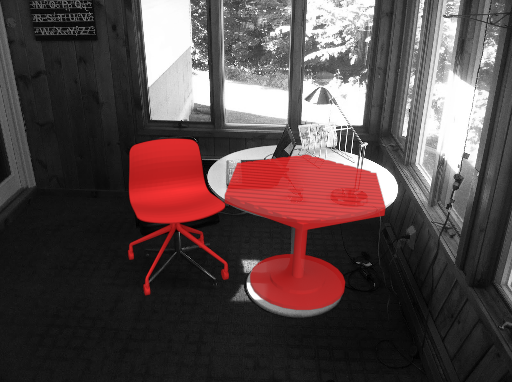}\\[-1.5pt]
		
		\colImgN{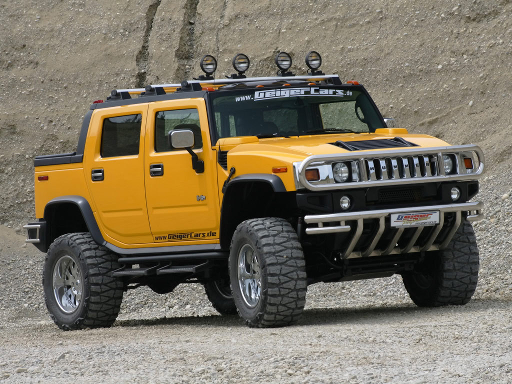}&   \colImgN{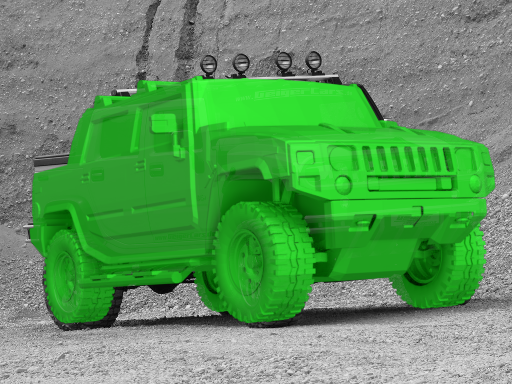}&
		\colImgN{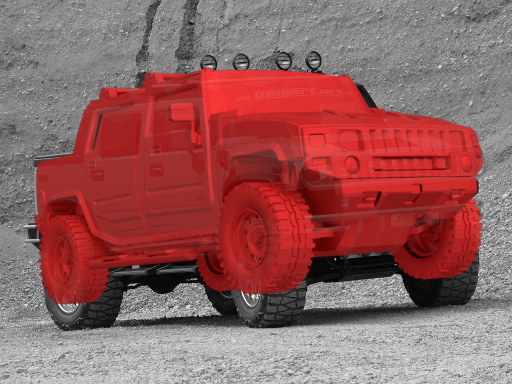}&  \colImgN{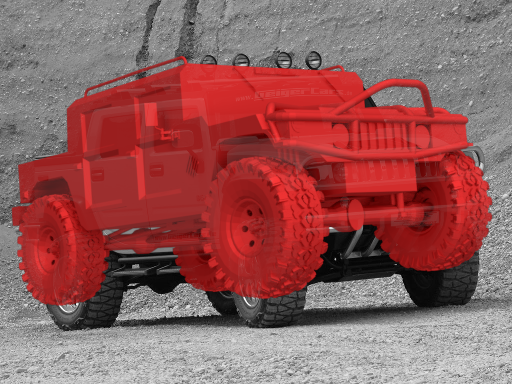}\\[-1.5pt]
		
		\colImgN{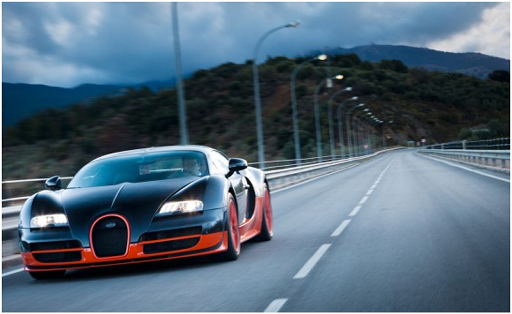}&   \colImgN{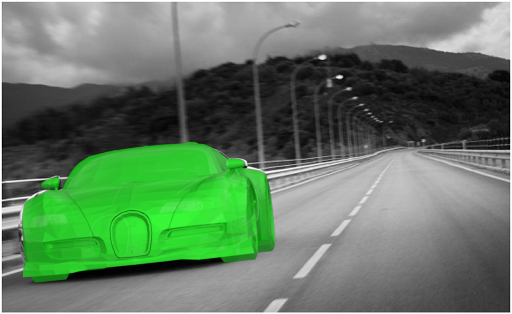}&
		\colImgN{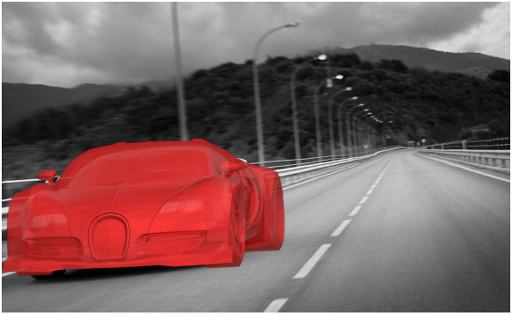}&  \colImgN{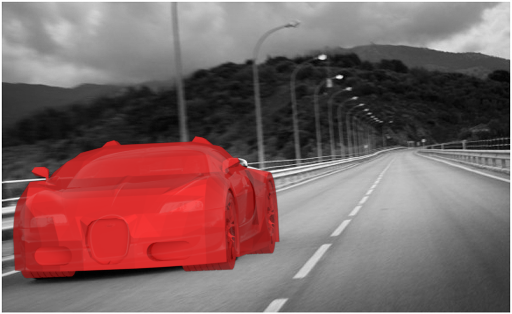}\\[-1.5pt]
		
		\footnotesize Image&\footnotesize GT&\footnotesize from seen&\footnotesize from unseen\\[-3pt]
	\end{tabular}
	\caption{Qualitative results for 3D pose estimation and 3D model retrieval from both seen and unseen databases. We project the retrieved 3D model onto the image using a 3D pose computed from the predicted location field by solving a \PNP~problem. For the ground truth 3D model, we use the ground truth 3D pose. In fact, location fields provide all relevant information to jointly address both tasks.}
	\label{fig:ret_pose}
	\vspace{-0.3269cm}
\end{figure}

Finally, Fig.~\ref{fig:ret_pose} presents examples for 3D model retrieval from both seen and unseen databases. In addition, we show that location fields provide all relevant information to also compute the 3D pose of objects. For this purpose, we sample 2D-3D correspondences from the location fields and solve a \PNP~ problem during inference. The projections onto the image show that both our retrieved 3D models and our computed 3D poses are highly accurate.  More qualitative results are shown in the {\bf supplementary material}.

\subsection{Ablation Study}
\label{sec:ablation}

To understand which aspects of our approach are crucial for performance, we conduct an ablation study. For this purpose, we perform experiments on Pix3D, which is the most challenging dataset, because it provides multiple categories and has the largest variation in object scale and pose. We report the mean performance across all categories in Table~\ref{table:ablations}.

If we train our approach without synthetic data, \ie, train our Descriptor CNN purely on predicted location fields, the performance decreases significantly. Since training data is limited, we do not see location fields from many different 3D poses during training in this scenario.

Next, if we predict location fields at half of our proposed resolution ($28\times28\times3$) the performance drops significantly. In this case, fine-grained structures, \eg, thin legs of a chair, cannot be recovered due to the limited spatial resolution.

Optimizing a pure softmax loss without our proposed combination of Center loss~\cite{Wen2016discriminative} and Triplet-Center loss~\cite{He2018triplet} results in a small performance decrease. This shows that our proposed Descriptor loss (see Eq.~\ref{eq:descriptor_loss}) indeed learns more discriminative embeddings than classification alone.

Training without Feature Mapping~\cite{Rad2018feature} only slightly decreases performance. This is in part due to the fact that we also address the domain gap by aggressively augmenting and degenerating rendered location fields during training of our Descriptor CNN to simulate predicted location fields. 

Finally, if we do not use our learned center descriptors but multi-view descriptors for matching, the performance almost remains the same. In this case, we match against 100 descriptors computed from location fields rendered under different 3D poses instead of a single center descriptor for each 3D model. This exhaustive comparison has a much higher computational complexity than our proposed approach. In fact, using center descriptors is not only significantly faster but also achieves better performance considering $d_{\text{HAU}}$. This experiment confirms that our approach indeed learns pose invariant 3D shape descriptors. 

\begin{table}
	\centering
	\begin{tabular}{lccc}
		\toprule
		Method
		&\multicolumn{1}{c}{$Acc_{\textit{Top-1}}$}
		&\multicolumn{1}{c}{$d_{\text{HAU}}$}
		&\multicolumn{1}{c}{$d_{\text{IOU}}$}\\
		\midrule
		Ours w/o synthetic data&55.0\%&0.0219&0.7156\\
		Ours half-res LFs&58.7\%&0.0204&0.7370\\
		Ours w/o (T)CL~\cite{Wen2016discriminative,He2018triplet}&59.9\%&0.0175&0.7621\\
		Ours w/o Mapping~\cite{Rad2018feature}&60.0\%&0.0174&0.7630\\
		Ours multi-view&\bf60.9\%&0.0173&\bf0.7686\\
		Ours&60.7\%&\bf0.0171&0.7663\\
		\bottomrule
	\end{tabular}
	\caption{Ablation study of our method on the Pix3D dataset. Exploiting synthetic data in the form of rendered location fields during training and employing location fields with sufficient resolution to capture thin structures are the most important aspects for increasing performance.}
	\label{table:ablations}
	\vspace{-0.2cm}
\end{table}

\section{Conclusion}

Learning a common embedding of 3D models and RGB images for single image 3D model retrieval is difficult due to limited training data and the domain gap between real and synthetic data. For this purpose, we map 3D models and RGB images to a common low-level representation in the form of location fields from which we compute pose invariant 3D shape descriptors. In this way, we bridge the domain gap and benefit from training on synthetic data. We evaluate our proposed approach on three challenging real-world datasets (Pix3D, Comp, and Stanford) and significantly outperform the state-of-the-art by up to 20\% absolute.

\vspace{-0.15cm}
\paragraph{Acknowledgement} This work was supported by the Christian Doppler Laboratory for Semantic 3D Computer Vision, funded in part by Qualcomm Inc. We gratefully acknowledge the support of NVIDIA Corporation with the donation of the Titan Xp GPU used for this research.

{\small
	\bibliographystyle{ieee}
	\bibliography{string,references}
}

\cleardoublepage


\twocolumn[{
   \newpage
   \null
   \vskip .375in
   \begin{center}
      {\Large \bf Location Field Descriptors:\\Single Image 3D Model Retrieval in the Wild\\Supplementary Material \par}
      \vspace*{24pt}
      {
      \large
      \lineskip .5em
      \begin{tabular}[t]{c}
      
      \end{tabular}
      \par
      }
      \vskip .5em
   \end{center}
   }]

In the following, we provide additional details and qualitative results of our novel 3D model retrieval approach called Location Field Descriptors.

\section{Datasets and Evaluation Setup}

We evaluate our proposed approach for 3D model retrieval in the wild on three challenging real-world datasets with different object categories: Pix3D~\cite{Sun2018pix3d} (\textit{bed}, \textit{chair}, \textit{sofa}, \textit{table}), Comp~\cite{Wang2018fine} (\textit{car}), and  Stanford~\cite{Wang2018fine} (\textit{car}). These datasets have only been released recently and, to the best of our knowledge, we are the first to report results for 3D model retrieval on all of them. In addition to retrieval from 3D models provided by these datasets, we also retrieve 3D models from ShapeNet~\cite{Shapenet2015}. Table~\ref{table:datasets} presents an overview of the object categories, the number of RGB images, and the number of 3D models in the evaluated datasets.

\begin{table*}
	\centering
	\setlength{\tabcolsep}{10pt}
	\begin{tabular}{lccccc}
		\toprule
		Dataset&Category&Train Images&Test Images&3D Models from Dataset& 3D Models from ShapeNet\\
		\midrule
		\multirow{4}{*}{Pix3D}&bed&203&191&19&254\\
		&chair&1506&1388&221&6778\\
		&sofa&552&540&20&3173\\
		&table&387&351&62&8443\\
		\midrule
		Comp&car&3798&1898&98&7497\\
		\midrule
		Stanford&car&8144&8041&134&7497\\
		\bottomrule
	\end{tabular}
	\caption{Overview of the evaluated datasets. The Pix3D, Comp, and Stanford datasets provide a large number of RGB images and a moderate number of 3D models with corresponding annotations. In contrast, ShapeNet does not provide RGB images but a large number of 3D models.}
	\label{table:datasets}
\end{table*}

The Pix3D dataset provides multiple categories, however, we only train and evaluate on categories which have more than 300 non-occluded and non-truncated samples (\textit{bed}, \textit{chair}, \textit{sofa}, \textit{table}). Further, we restrict the training and evaluation to samples marked as non-occluded and non-truncated, because we do not know which objects parts are occluded nor the extent of the occlusion, and many objects are heavily truncated. For each 3D model, we randomly choose 50\% of the corresponding images for training and the other 50\% for testing.

The Comp and Stanford datasets only provide one category (\textit{car}). Most images show one prominent car which is non-occluded and non-truncated. The two datasets already provide a train-test split. Thus, we use all available samples from Comp and Stanford for training and evaluation.

In contrast to these datasets which provide a large number of RGB images and a moderate number of 3D models with corresponding annotations, ShapeNet does not provide RGB images but a large number of 3D models. Due to its enormous size, ShapeNet does not only cover many different object categories but also presents a large variety in 3D model geometry. If 3D models are present in the respective dataset and in ShapeNet, we exclude them for retrieval from ShapeNet to evaluated retrieval from an entirely unseen database.

We consistently orient, scale and translate all 3D models. In particular, we rotate all 3D models to have common front facing, up, and starboard directions. Additionally, we scale and translate all 3D models to fit inside a unit cube centered at the coordinate origin $(0,0,0)$ while preserving the aspect-ratio of the 3D dimensions. This 3D model alignment is not only important for training our approach but also for the evaluated metrics. For example, computing the modified Hausdorff distance and the 3D IOU between two 3D models is only meaningful if they are consistently oriented, scaled and centered.

\section{Implementation and Training Details}

For our Location Field CNN, we use a Feature Pyramid Network~\cite{Lin2017feature} on top of a ResNet-101 backbone~\cite{He2016deep,He2016identity}. For our Descriptor CNN, we use a DenseNet-50 architecture~\cite{Huang2017densely} with 3 dense blocks and a growth rate of 24. For our implementation, we resize and pad RGB images to a spatial resolution of $512\times512\times3$ maintaining the aspect ratio. For the location fields, we employ a resolution of $58\times58\times3$. In this configuration, the Descriptor CNN maps low-resolution location fields to a 270-dimensional embedding space. 

We initialize the convolutional backbone and the detection branches of the Location Field CNN with weights trained for instance segmentation~\cite{He2017mask} on COCO~\cite{Lin2014microsoft}. The location field branch and the Descriptor CNN are trained from scratch. We train our networks for 300 epochs using a batch size of 32. The initial learning rate of $1e^{-3}$ is decreased by a factor of 5 after 150 and 250 epochs.

We employ different forms of data augmentation. For RGB images, we use mirroring, jittering of location, scale, and rotation, and independent pixel augmentations like additive noise. For rendered location fields, we additionally use different forms of blurring to simulate predicted location fields. During training of the Descriptor CNN, we further leverage synthetic data and train on predicted and rendered location fields using a ratio of $1:3$.

To balance the individual terms in the system loss
\begin{equation}
L = L_{\text{D}} + L_{\text{softmax}} + \alpha L_{\text{C}} + \beta L_{\text{TC}} + \gamma L_{\text{LF}} + \delta L_{\text{FM}},
\label{eq:descriptor_loss}
\end{equation}
we assign unit weights to classification losses and non-unit weights to regression losses. Thus, we combine the unmodified Detection losses ($L_{\text{D}}$) of the generalized Faster/Mask R-CNN framework and the Descriptor CNN softmax loss ($L_{\text{softmax}}$) with the weighted Center ($L_{\text{C}}$), Triplet-Center ($L_{\text{TC}}$), Location Field ($L_{\text{LF}}$) and Feature Mapping ($L_{\text{FM}}$) losses. We experimentally set $\alpha=0.01$, $\beta=0.1$, $\gamma=10$, $\delta=0.01$, and use a margin of $m=1$. For the Huber distance~\cite{Huber1964robust}, we set the threshold to $1$. 

\section{Failure Cases}
\begin{figure*}
	\setlength{\tabcolsep}{1pt}
	\setlength{\fboxsep}{-2pt}
	\setlength{\fboxrule}{2pt}
	\newcommand{\colImgN}[1]{{\includegraphics[width=0.12\linewidth]{#1}}}
	\centering
	\begin{tabular}{cccccc}
		\colImgN{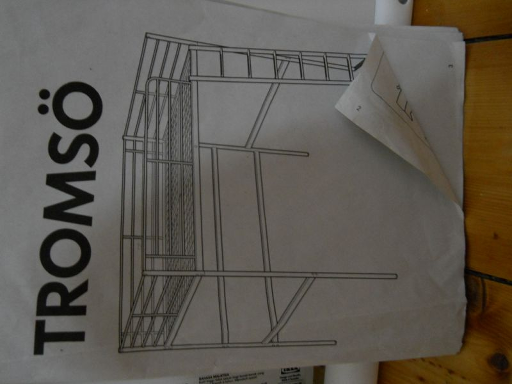}&   \colImgN{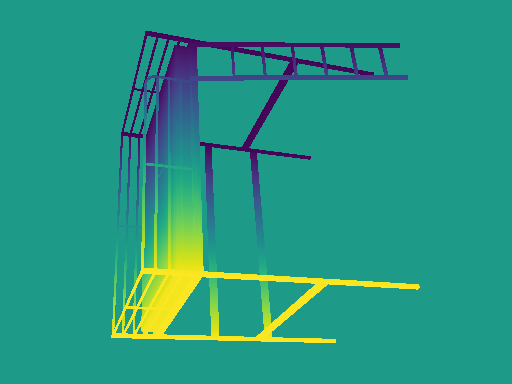}&   \colImgN{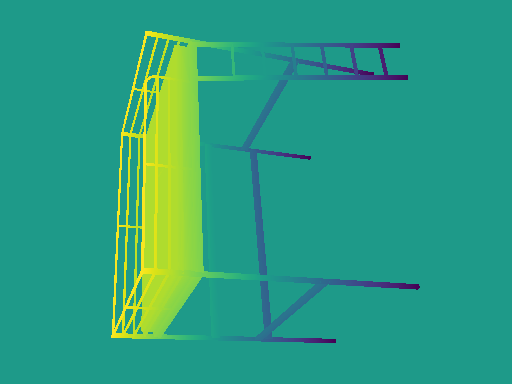}&   \colImgN{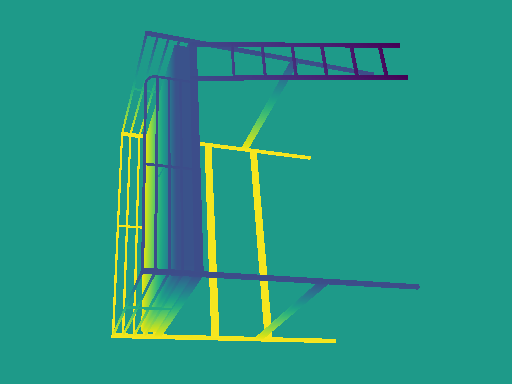}&
		\colImgN{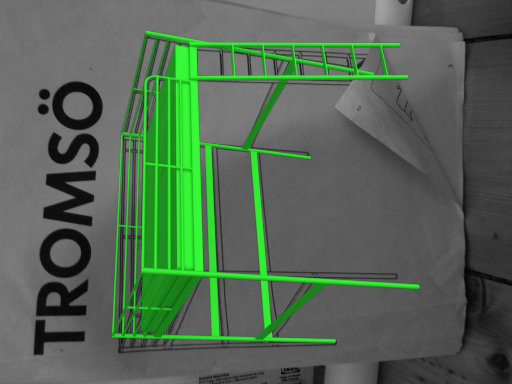}\\[-1.5pt]
		&\colImgN{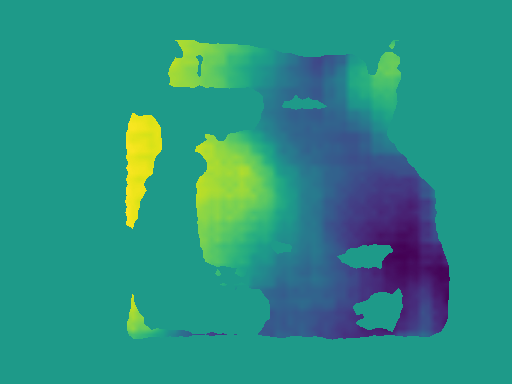}&   \colImgN{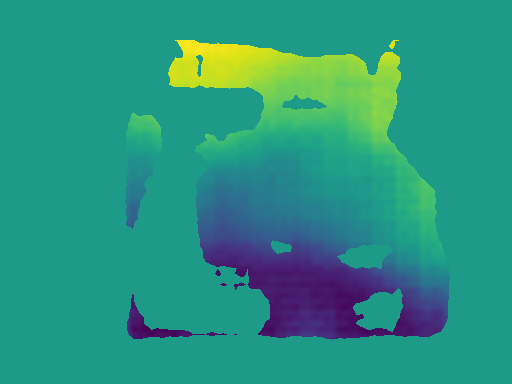}&   \colImgN{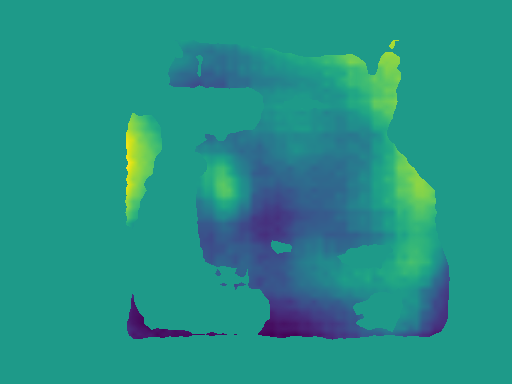}&
		\colImgN{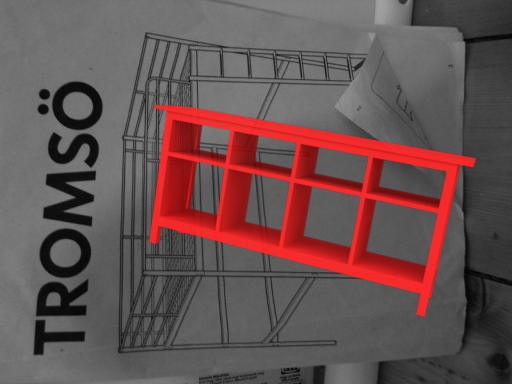}&
		\colImgN{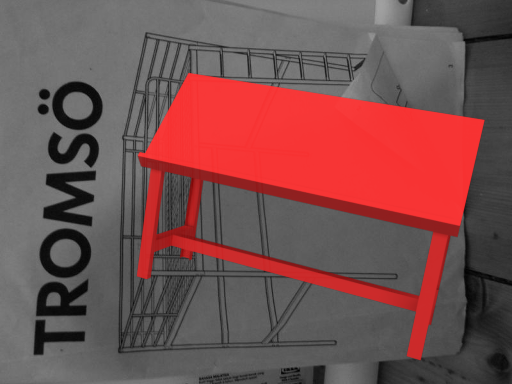}\\[-1.5pt]
		
		\colImgN{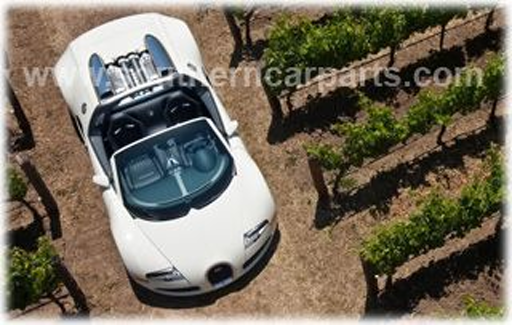}&   \colImgN{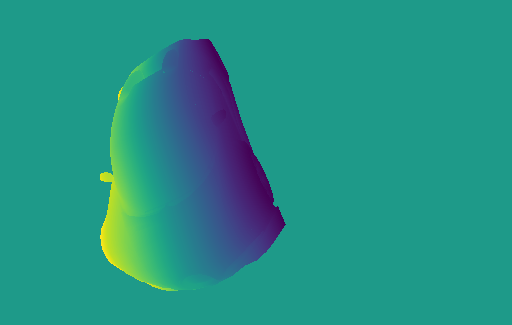}&   \colImgN{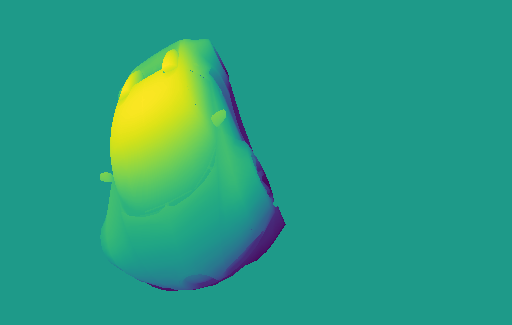}&   \colImgN{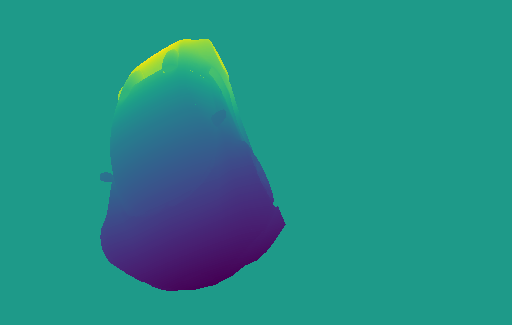}&
		\colImgN{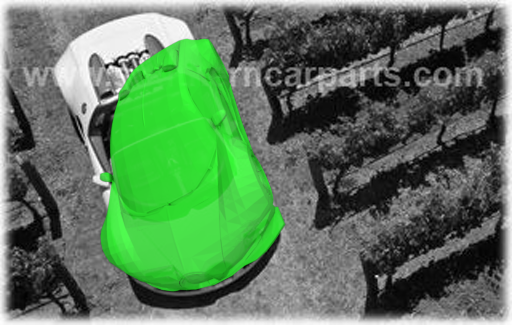}\\[-1.5pt]
		&\colImgN{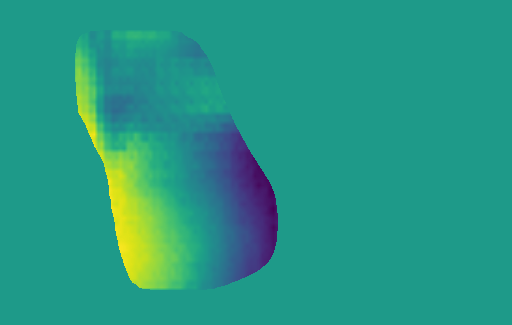}&   \colImgN{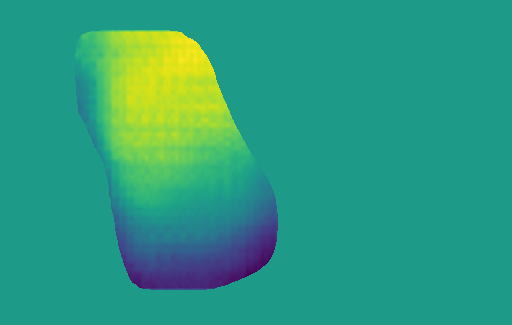}&   \colImgN{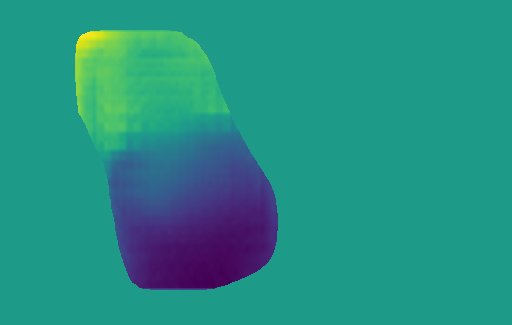}&
		\colImgN{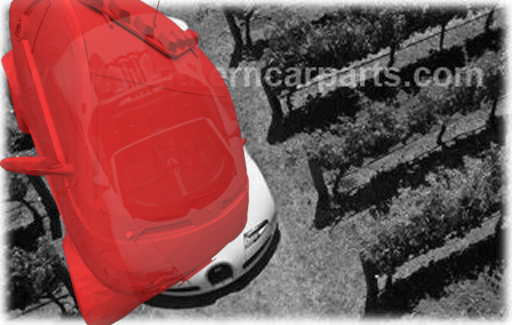}&
		\colImgN{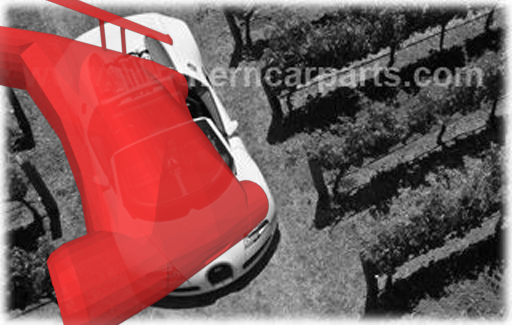}\\[-1.5pt]
		
		\colImgN{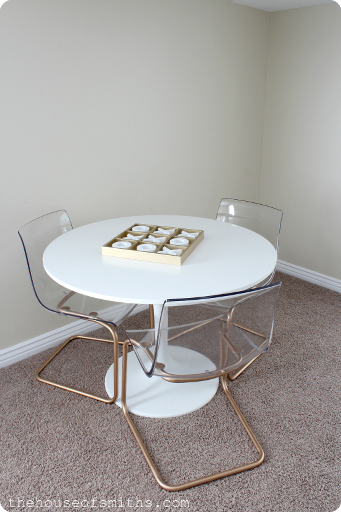}&   \colImgN{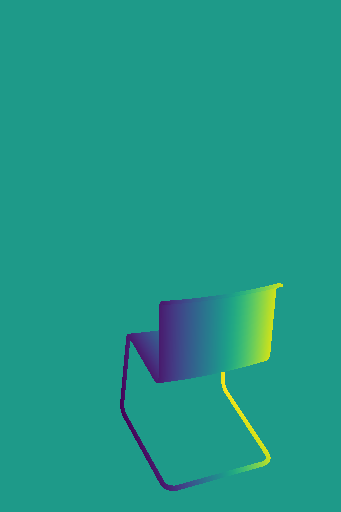}&   \colImgN{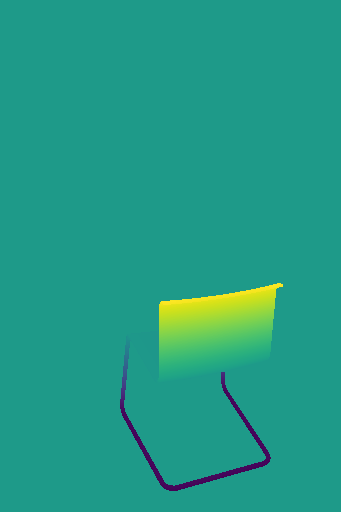}&   \colImgN{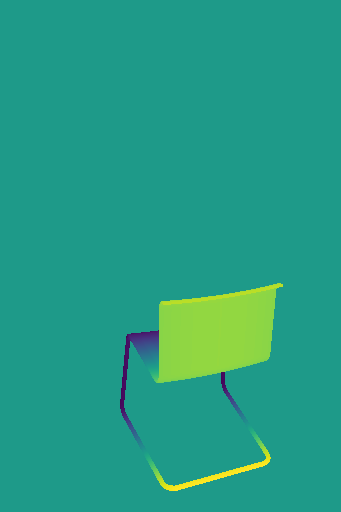}&
		\colImgN{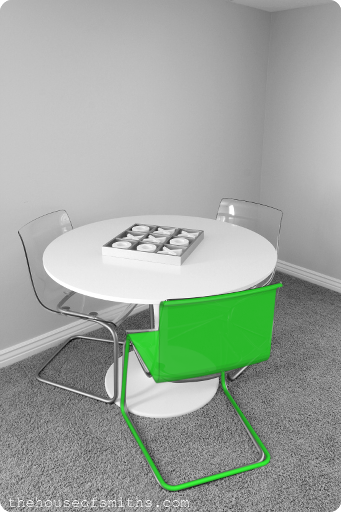}\\[-1.5pt]
		&\colImgN{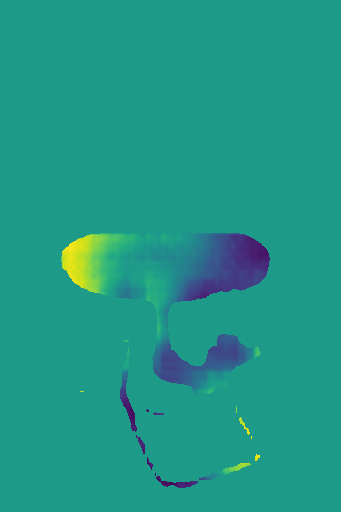}&   \colImgN{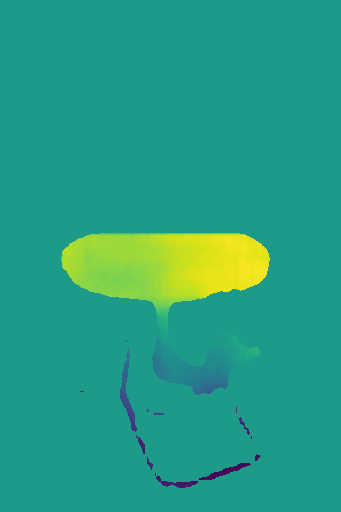}&   \colImgN{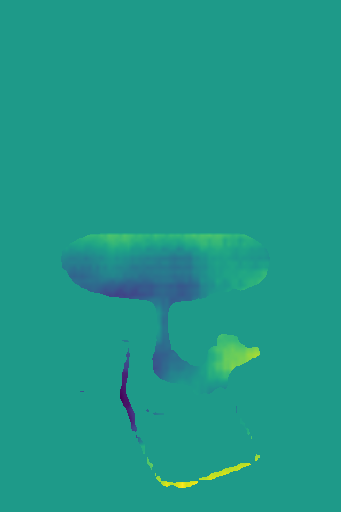}&
		\colImgN{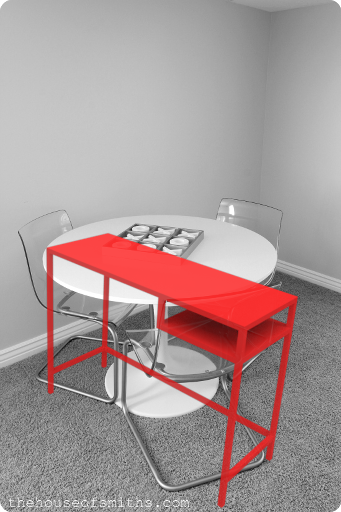}&
		\colImgN{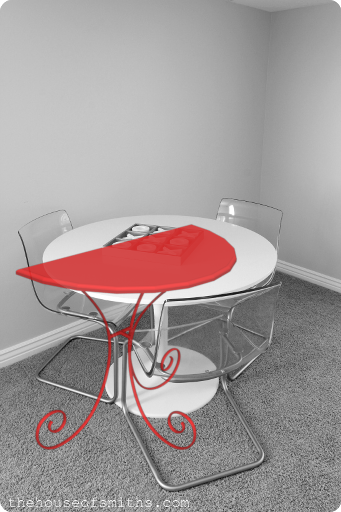}\\[-1.5pt]
		
		\colImgN{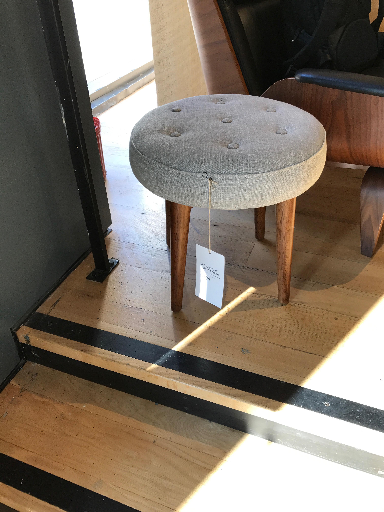}&   \colImgN{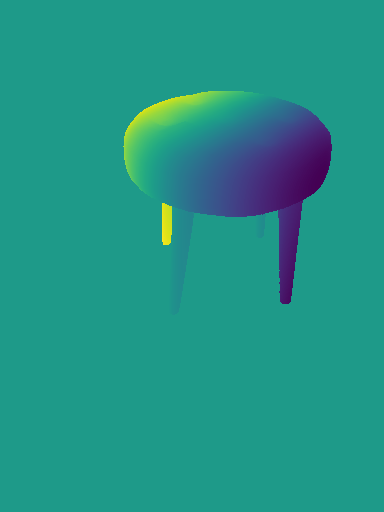}&   \colImgN{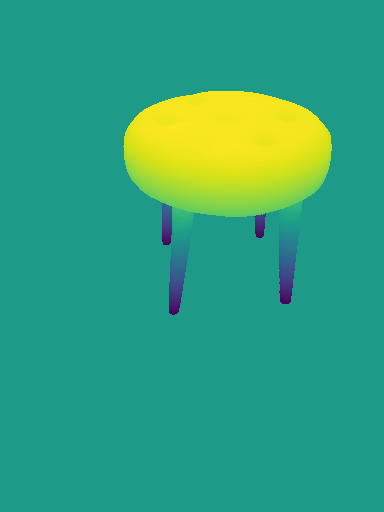}&   \colImgN{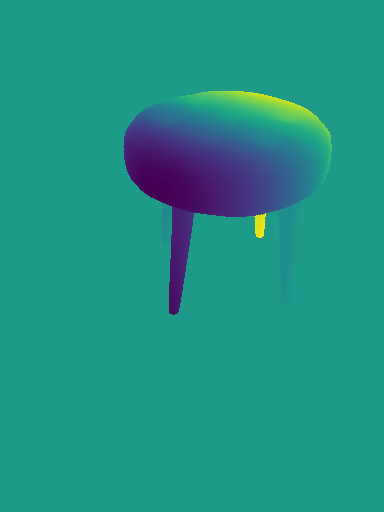}&
		\colImgN{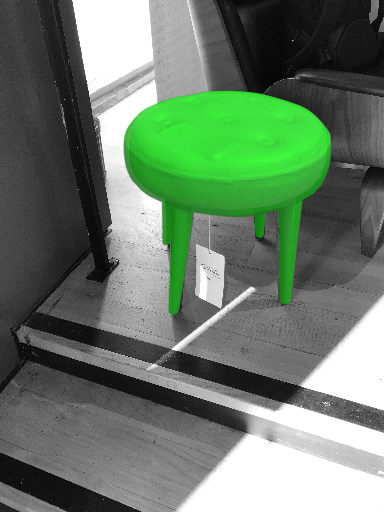}\\[-1.5pt]
		&\colImgN{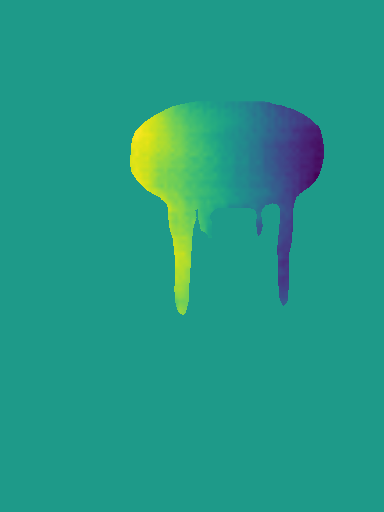}&   \colImgN{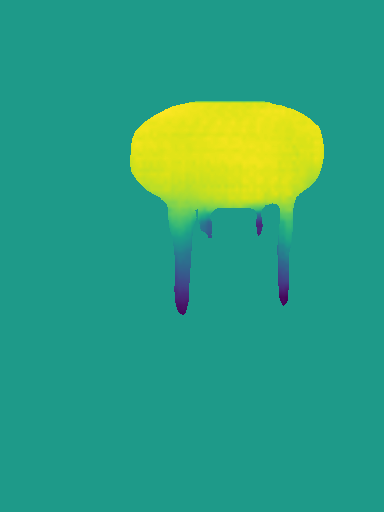}&   \colImgN{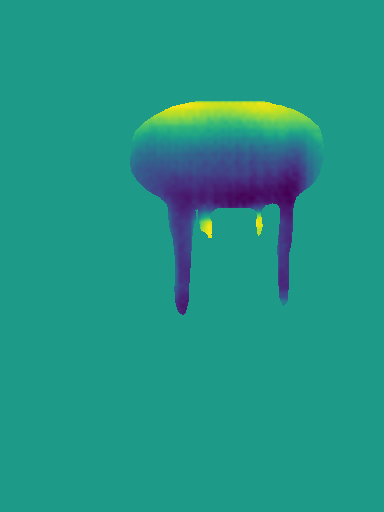}&
		\colImgN{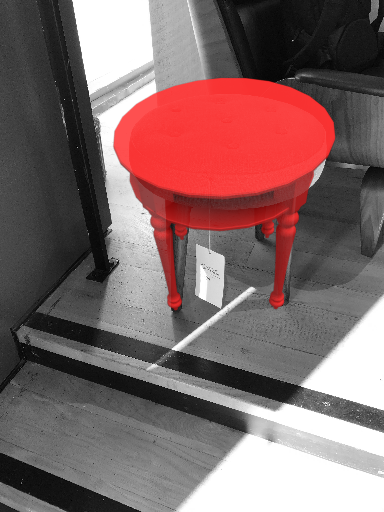}&
		\colImgN{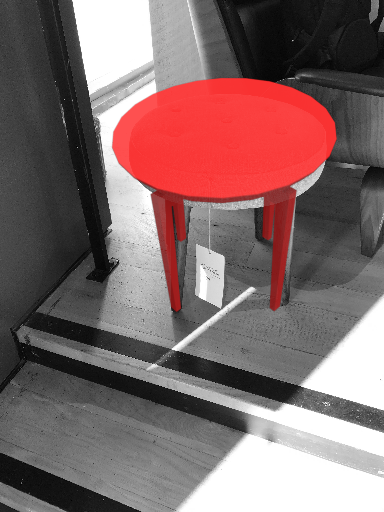}\\[-1.5pt]
		
		\footnotesize Image&\footnotesize LF (X)&\footnotesize LF (Y)&\footnotesize LF (Z)&\footnotesize from seen&\footnotesize from unseen\\[-3pt]
	\end{tabular}
	\caption{Failure cases of our approach. For each example image, the top row shows the ground truth and the bottom row shows our prediction. Most failure cases relate to incorrect location field predictions. If the 3D pose of the object in the image is far from the 3D poses seen during training (first and second example), we cannot predict an accurate location field. The location field prediction can also fail in complex occlusion scenarios if multiple objects are detected as a single object (third example). In other cases, we predict an accurate location field, but retrieve a 3D model from a different category due to ambiguous 3D geometries, \eg, \textit{table} instead of \textit{chair}.}
	\label{fig:fails}
\end{figure*}

Fig. \ref{fig:fails} shows failure cases of our approach. Most failure cases relate to incorrect location field predictions. For example, if the 3D pose of the object in the image is far from the 3D poses seen during training, or if multiple objects are detected as a single object in a complex occlusion scenario, we cannot predict an accurate location field. In other failure cases, we predict an accurate location field, but retrieve a 3D model from a different category due to ambiguous 3D geometries, \eg, \textit{table} instead of \textit{chair}. While the detection branch of the generalized Faster/Mask R-CNN predicts a category for each detected object in an RGB image, we do not use this information during retrieval, because there is a category ambiguity for many objects. For example, it is unclear if a couch with sleeping functionality is a \textit{sofa} or a \textit{bed} (see Fig.~\ref{fig:ret_pose1}, left column, last example).

\section{Additional Qualitative Results}
Finally, we present additional qualitative results which complement those presented in the main paper.

\begin{figure*}[t]
	\setlength{\tabcolsep}{1pt}
	\setlength{\fboxsep}{-2pt}
	\setlength{\fboxrule}{2pt}
	\newcommand{\colImgN}[1]{{\includegraphics[width=0.24\linewidth]{#1}}}
	\centering
	\begin{subfigure}[t]{\columnwidth}
	\begin{tabular}{ccccc}
		\colImgN{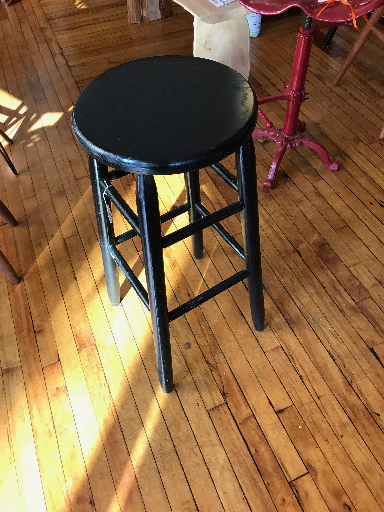}&   \colImgN{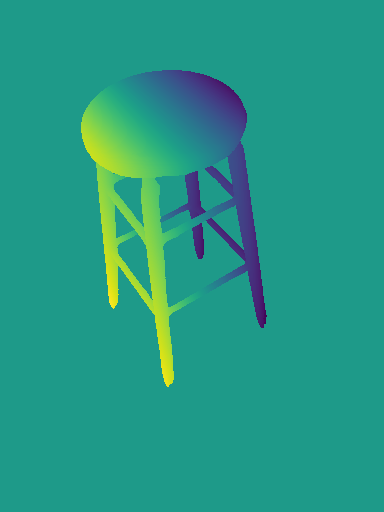}&   \colImgN{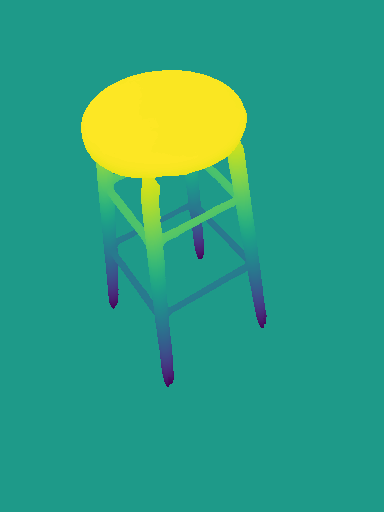}&   \colImgN{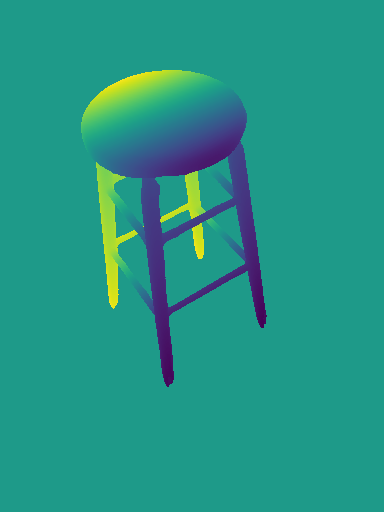}\\[-1.5pt]
		&\colImgN{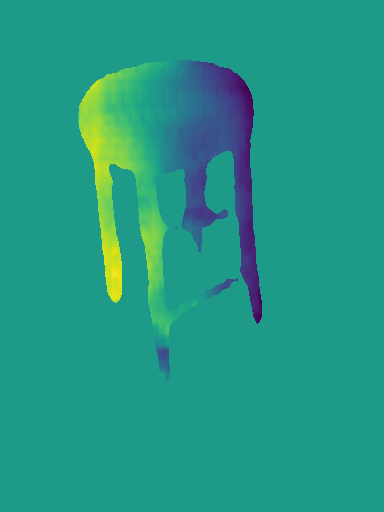}&   \colImgN{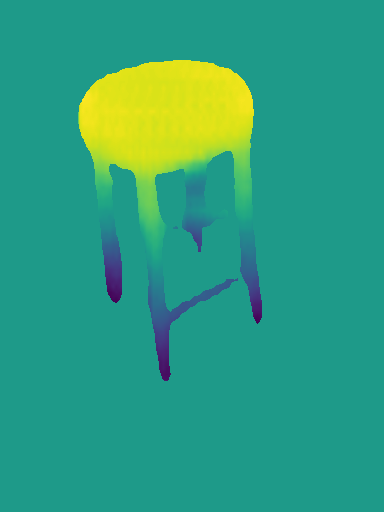}&   \colImgN{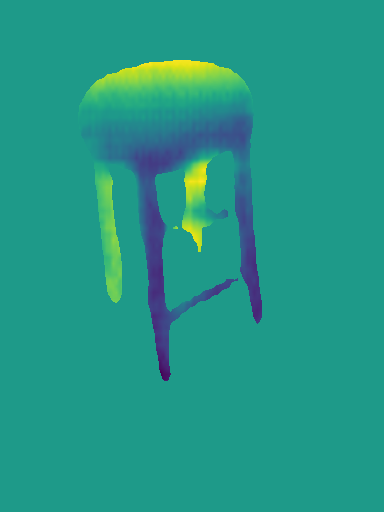}\\[-1.5pt]
		
		\colImgN{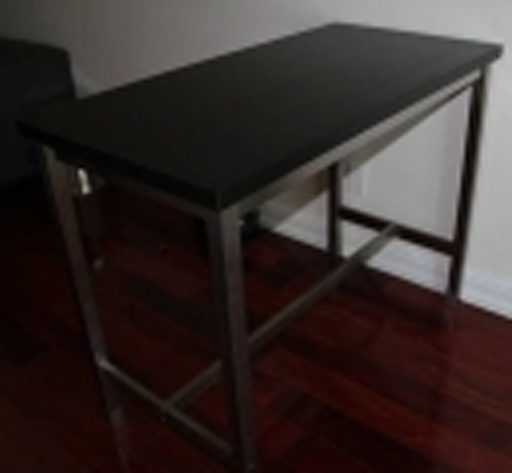}&   \colImgN{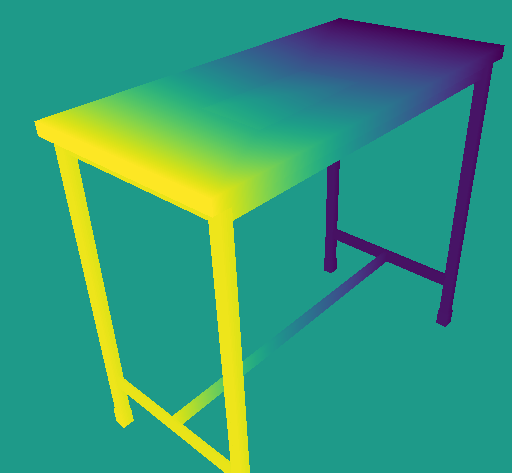}&   \colImgN{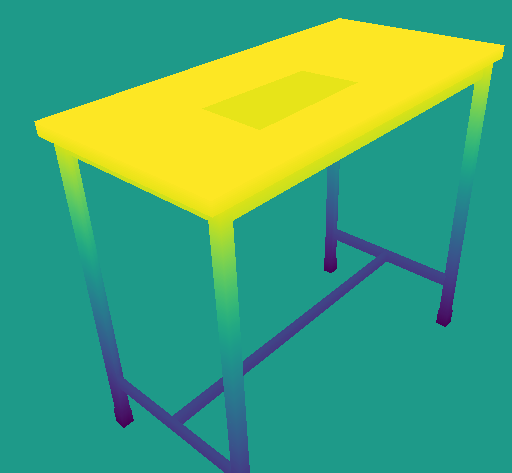}&   \colImgN{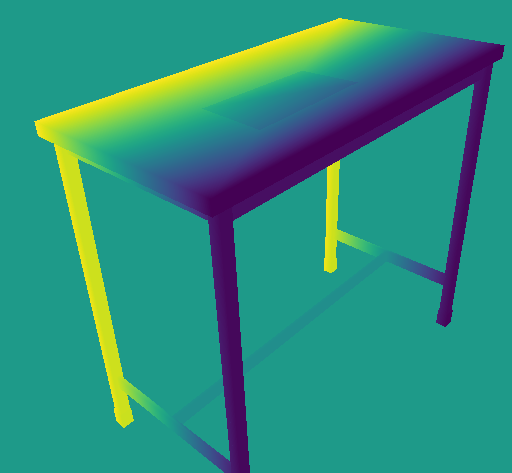}\\[-1.5pt]
		&\colImgN{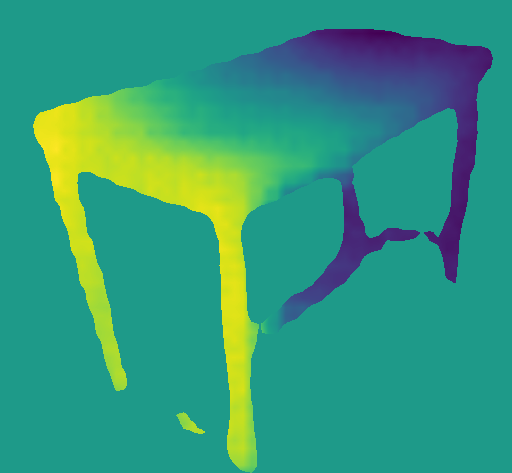}&   \colImgN{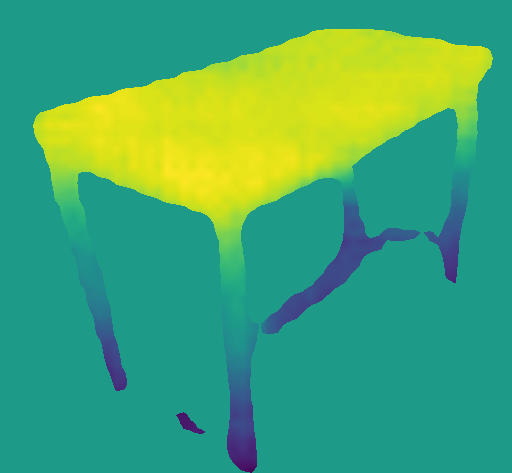}&   \colImgN{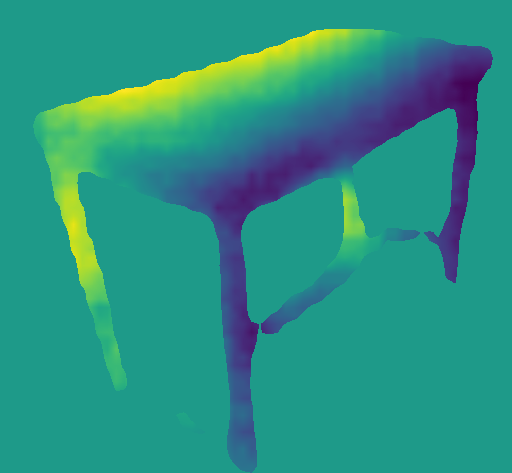}\\[-1.5pt]
		
		\colImgN{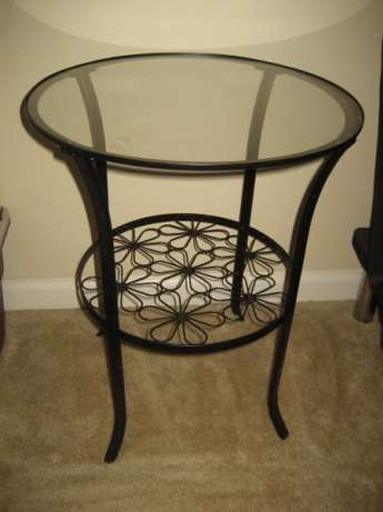}&   \colImgN{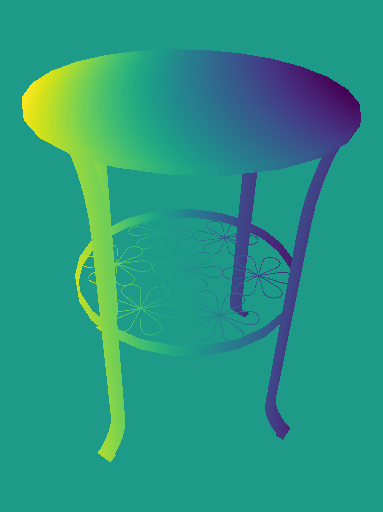}&   \colImgN{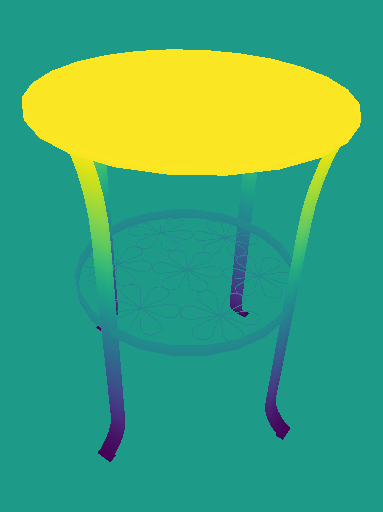}&   \colImgN{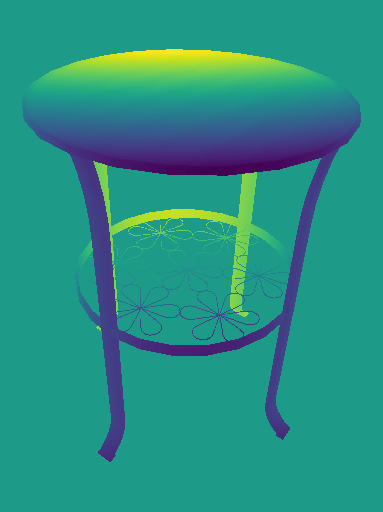}\\[-1.5pt]
		&\colImgN{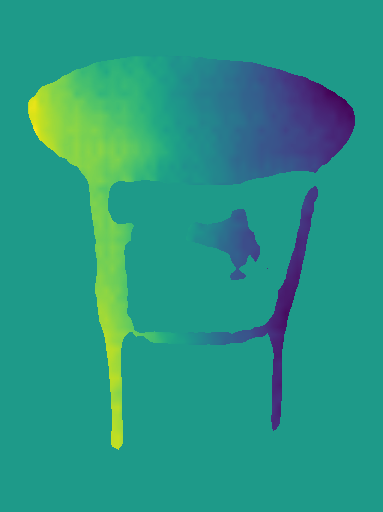}&   \colImgN{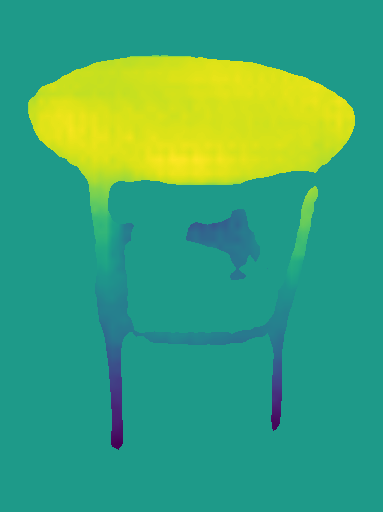}&   \colImgN{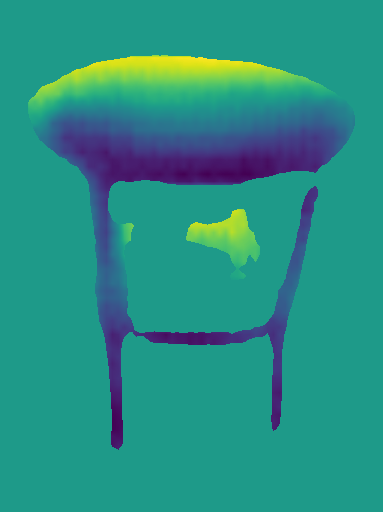}\\[-1.5pt]
		
		\colImgN{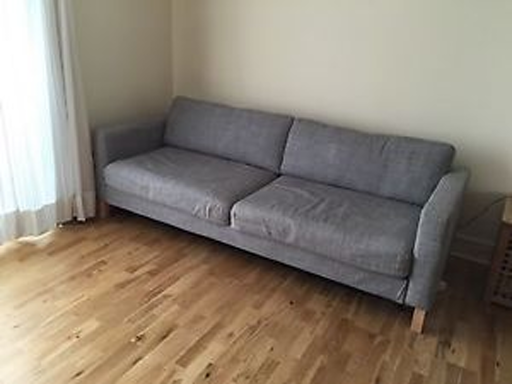}&   \colImgN{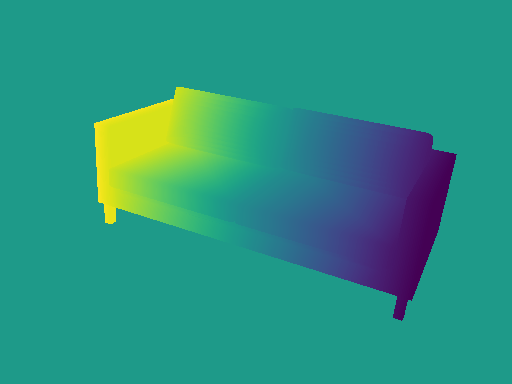}&   \colImgN{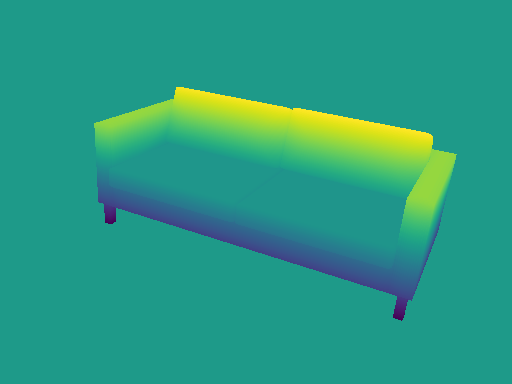}&   \colImgN{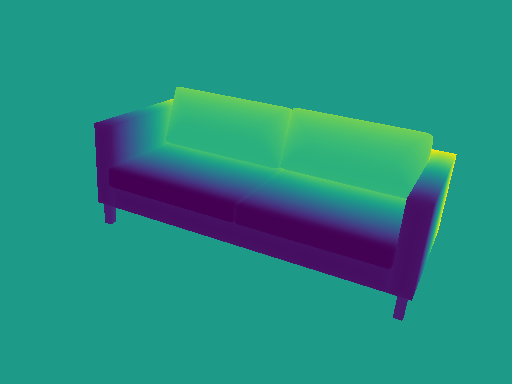}\\[-1.5pt]
		&\colImgN{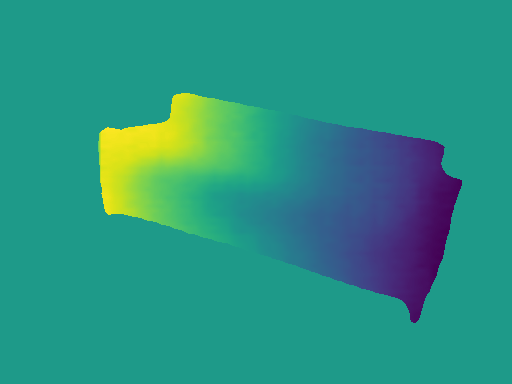}&   \colImgN{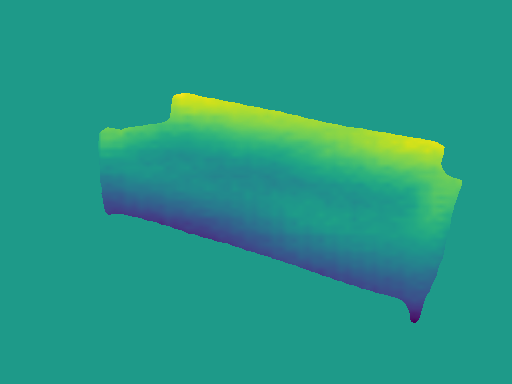}&   \colImgN{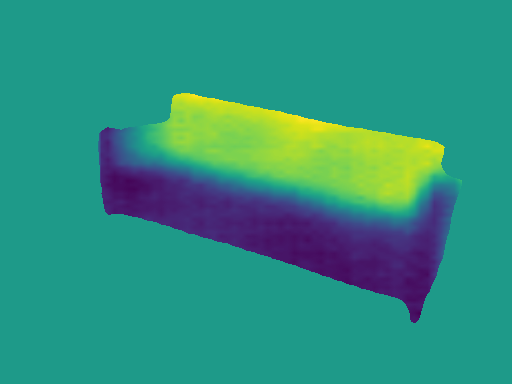}\\[-1.5pt]
		
		\colImgN{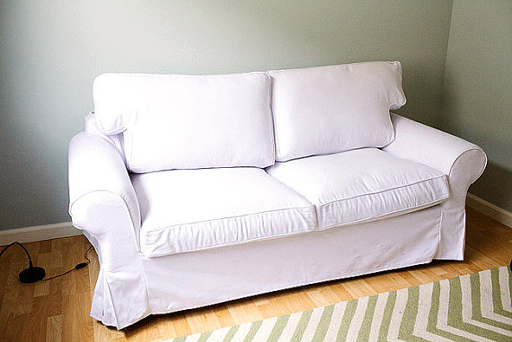}&   \colImgN{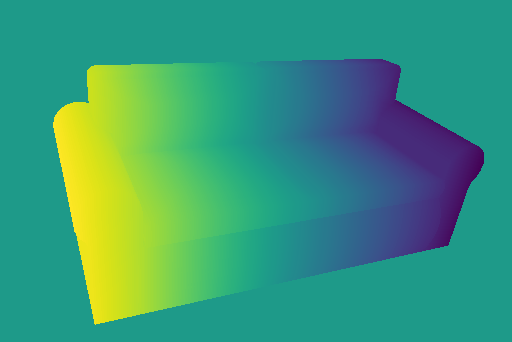}&   \colImgN{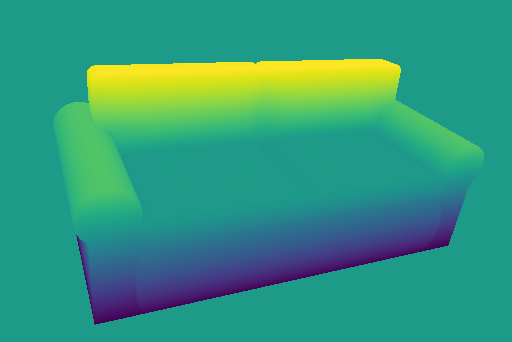}&   \colImgN{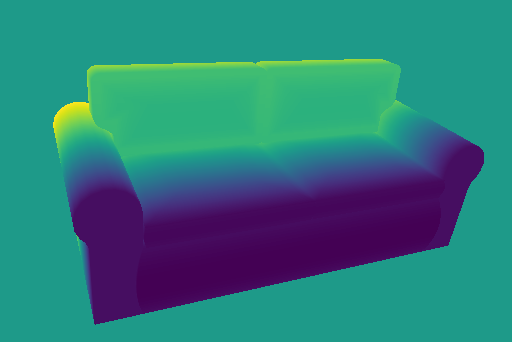}\\[-1.5pt]
		&\colImgN{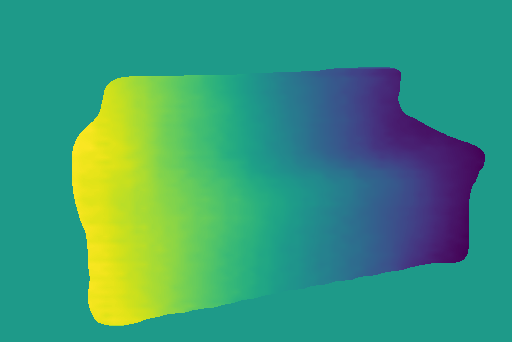}&   \colImgN{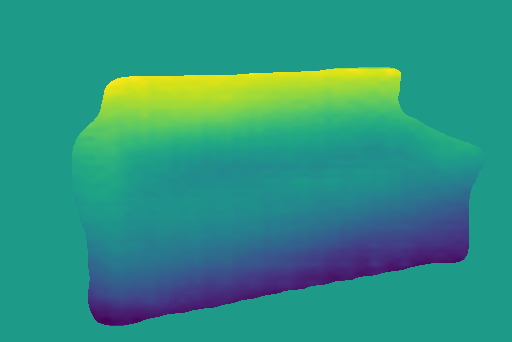}&   \colImgN{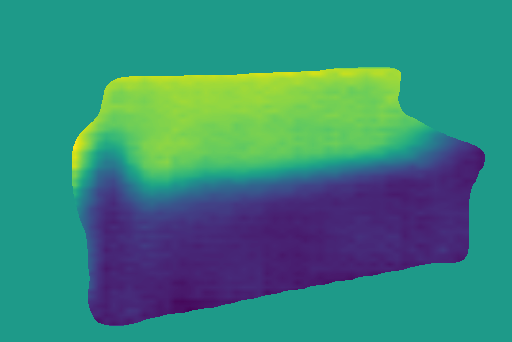}\\[-1.5pt]
		\footnotesize Image&\footnotesize LF (X)&\footnotesize LF (Y)&\footnotesize LF (Z)\\[-3pt]
	\end{tabular}
	\end{subfigure}\hfill\begin{subfigure}[t]{\columnwidth}
	\begin{tabular}{ccccc}
		\colImgN{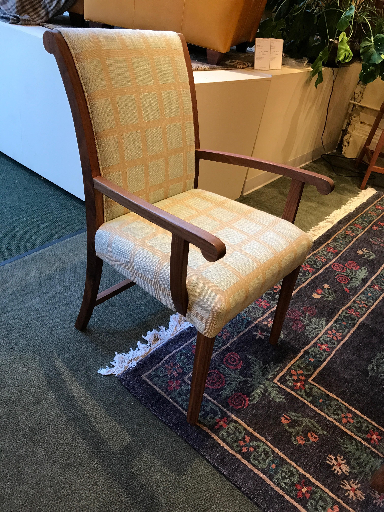}&   \colImgN{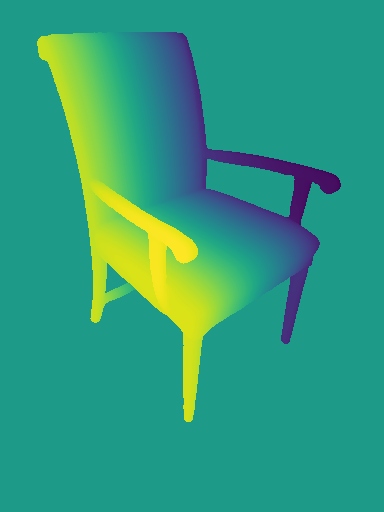}&   \colImgN{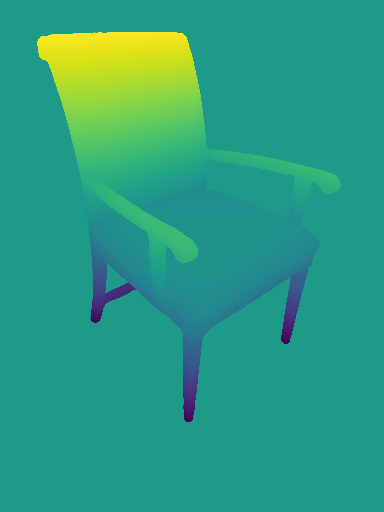}&   \colImgN{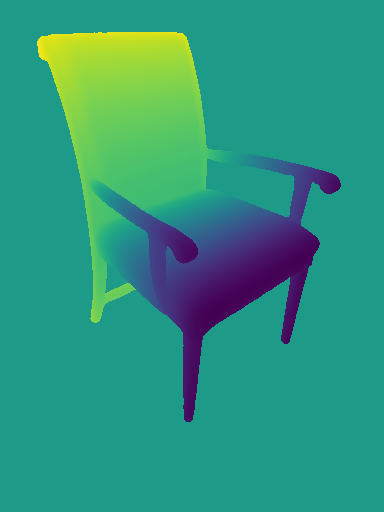}\\[-1.5pt]
		&\colImgN{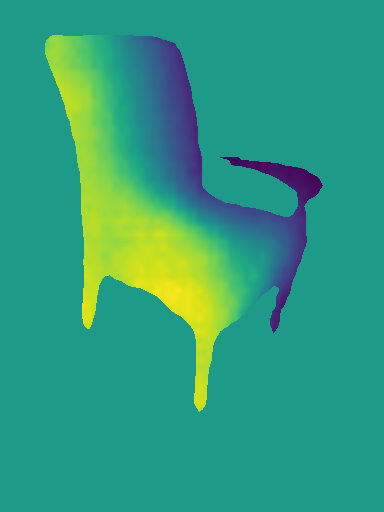}&   \colImgN{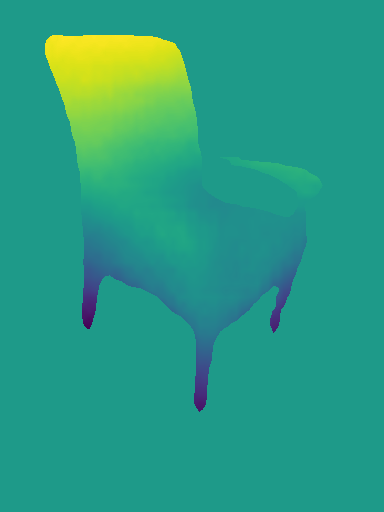}&   \colImgN{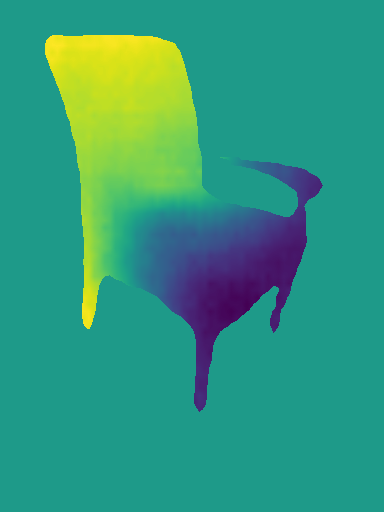}\\[-1.5pt]
		
		\colImgN{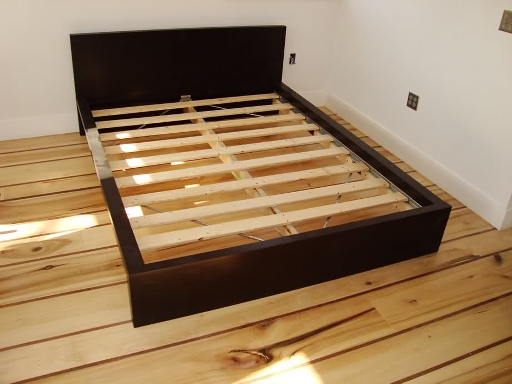}&   \colImgN{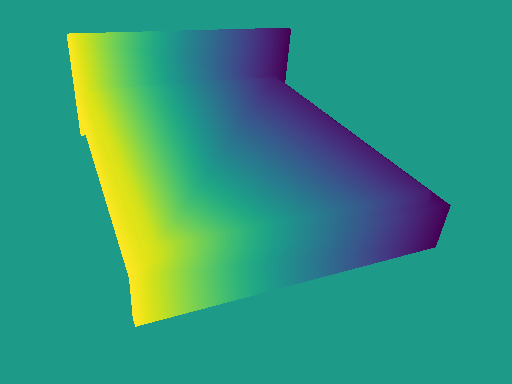}&   \colImgN{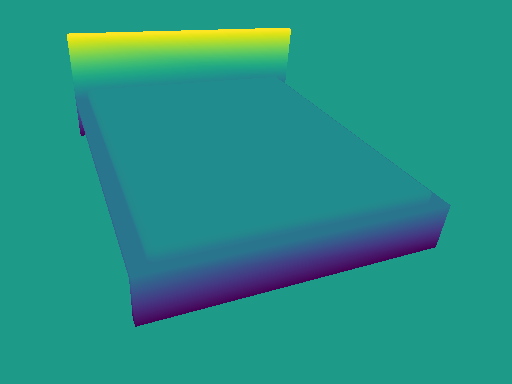}&   \colImgN{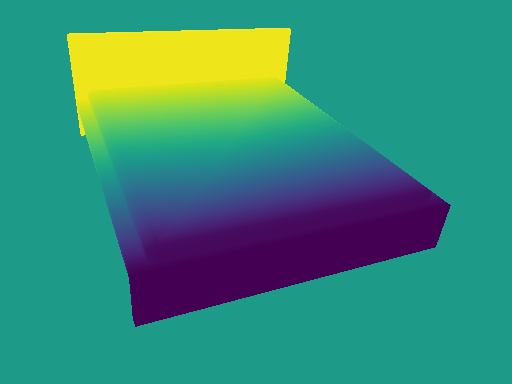}\\[-1.5pt]
		&\colImgN{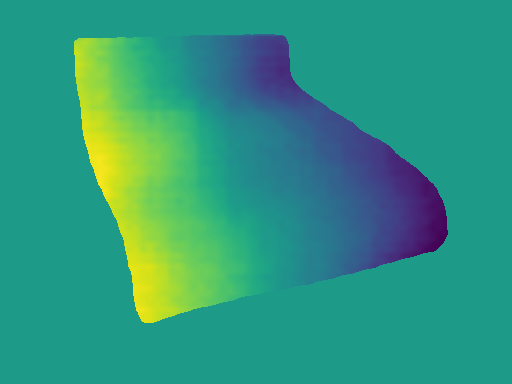}&   \colImgN{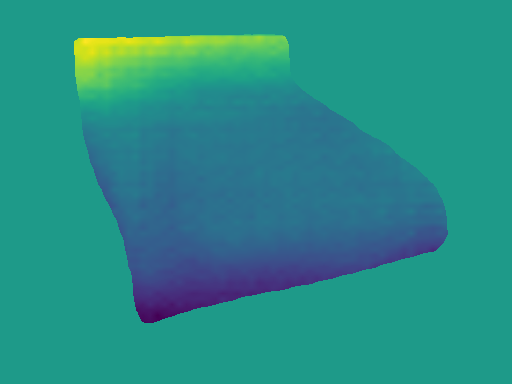}&   \colImgN{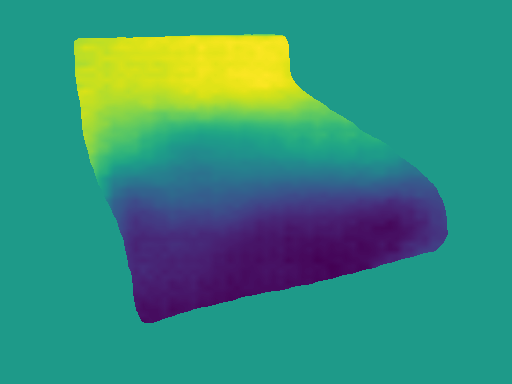}\\[-1.5pt]
		
		\colImgN{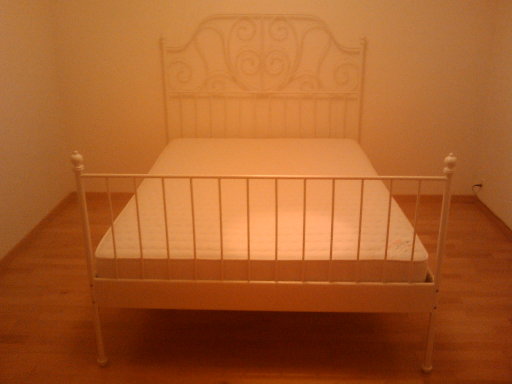}&   \colImgN{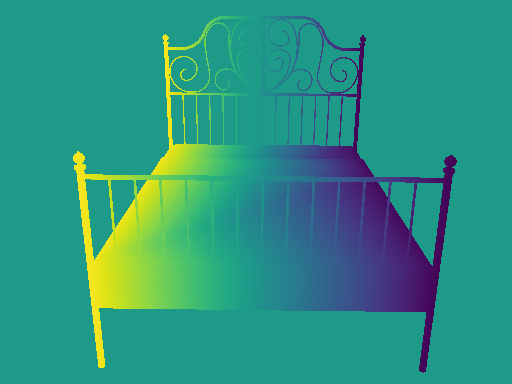}&   \colImgN{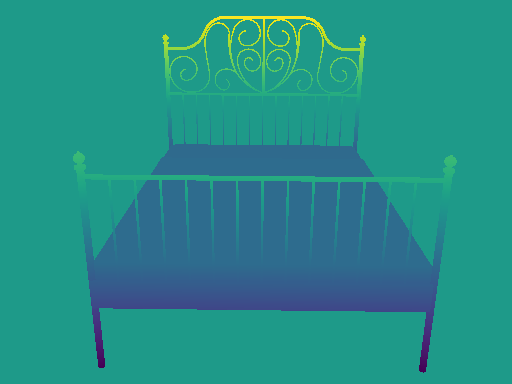}&   \colImgN{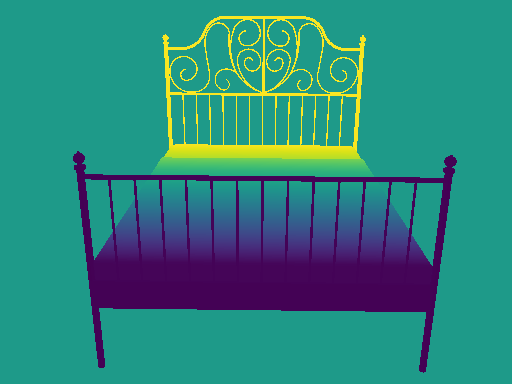}\\[-1.5pt]
		&\colImgN{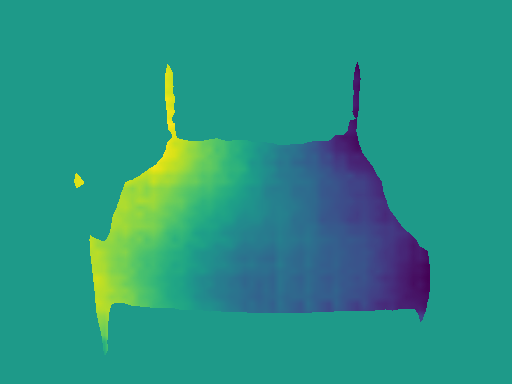}&   \colImgN{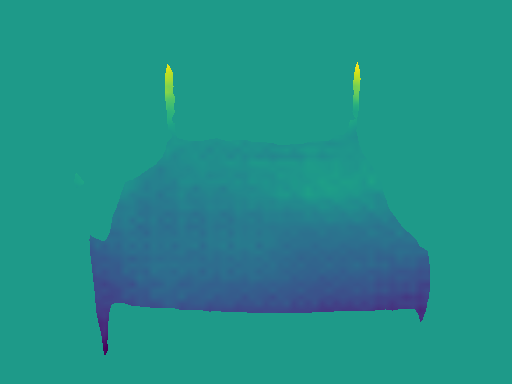}&   \colImgN{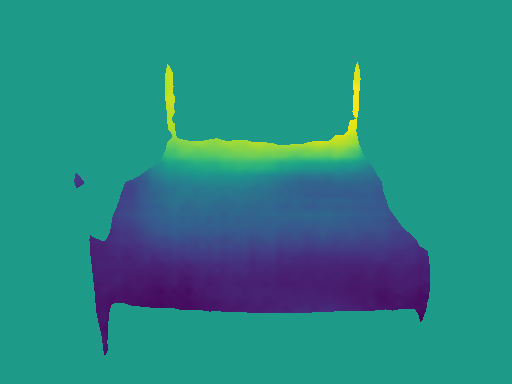}\\[-1.5pt]
		
		\colImgN{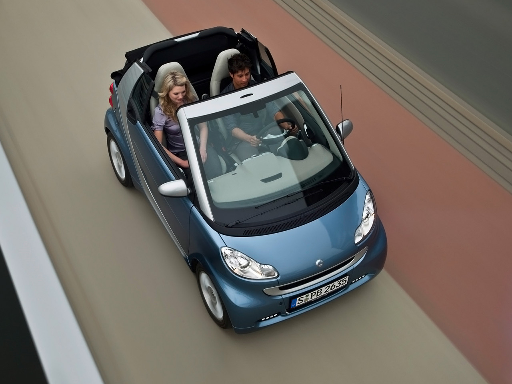}&   \colImgN{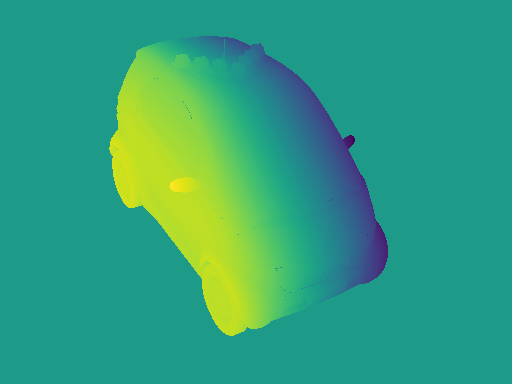}&   \colImgN{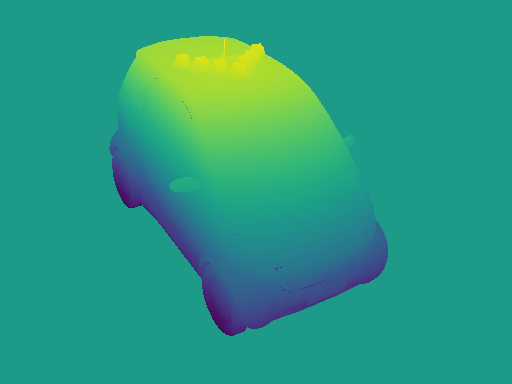}&   \colImgN{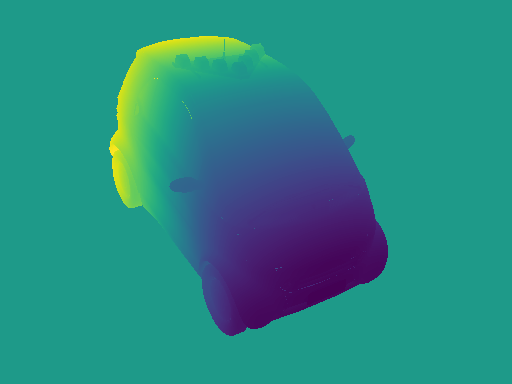}\\[-1.5pt]
		&\colImgN{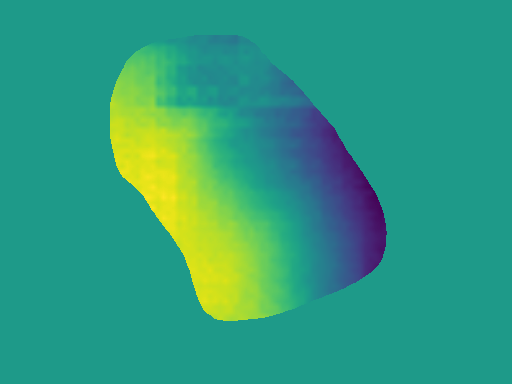}&   \colImgN{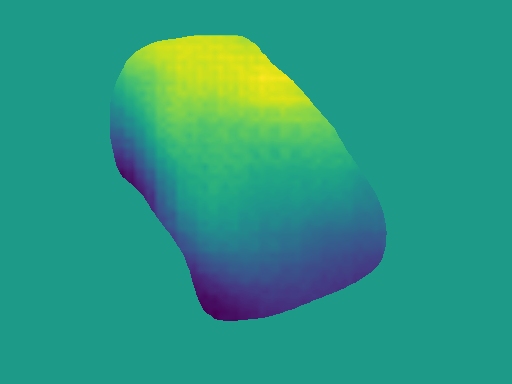}&   \colImgN{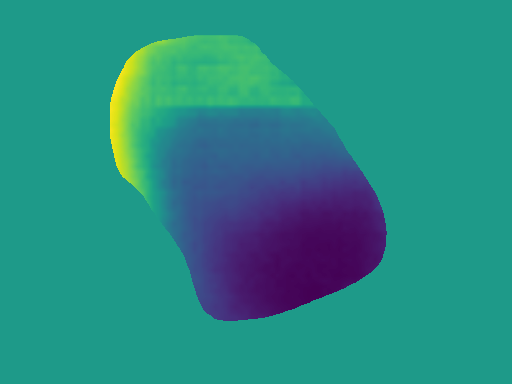}\\[-1.5pt]
		
		\colImgN{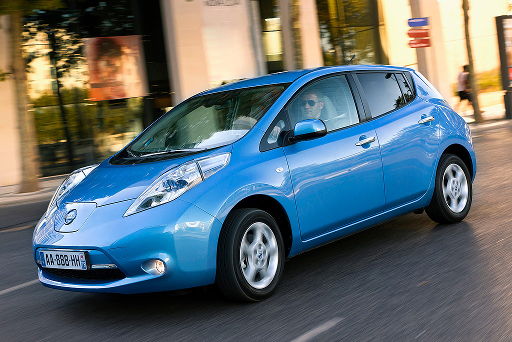}&   \colImgN{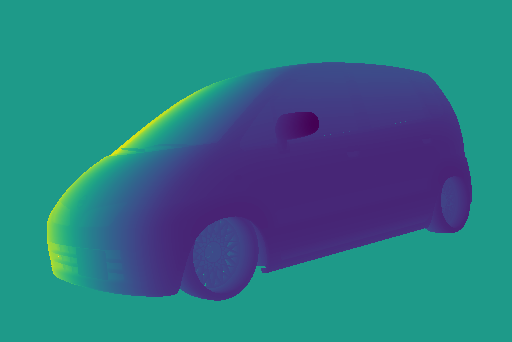}&   \colImgN{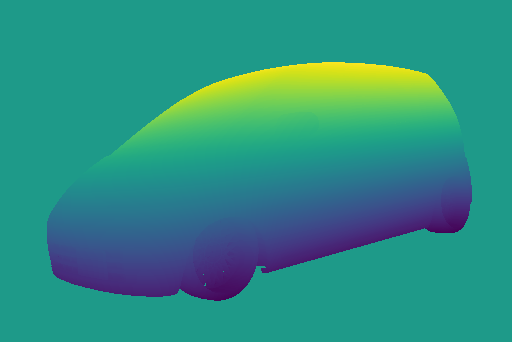}&   \colImgN{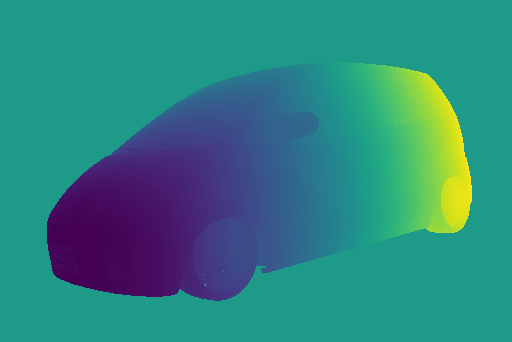}\\[-1.5pt]
		&\colImgN{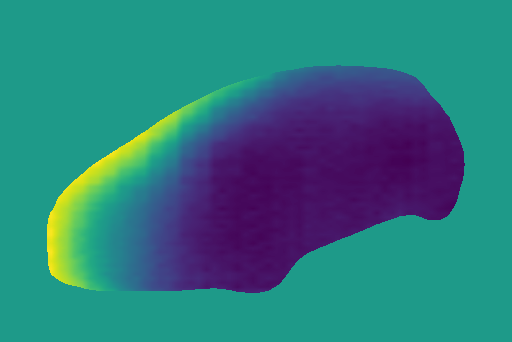}&   \colImgN{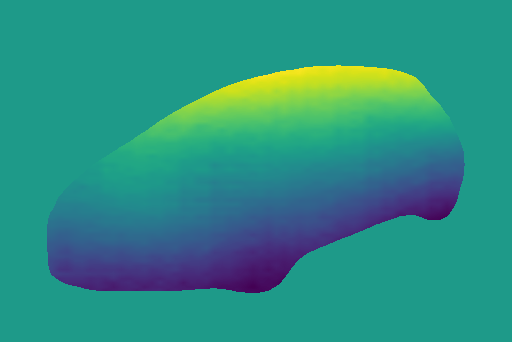}&   \colImgN{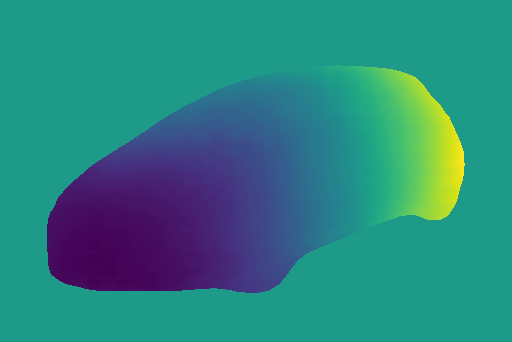}\\[-1.5pt]
		
		\colImgN{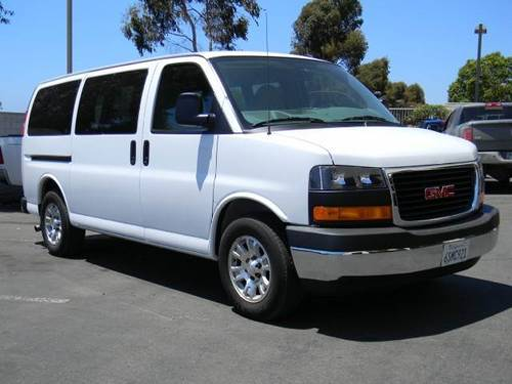}&   \colImgN{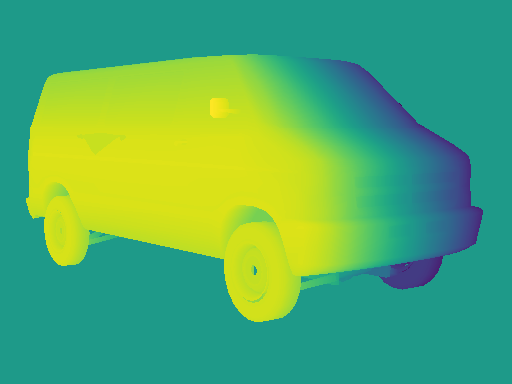}&   \colImgN{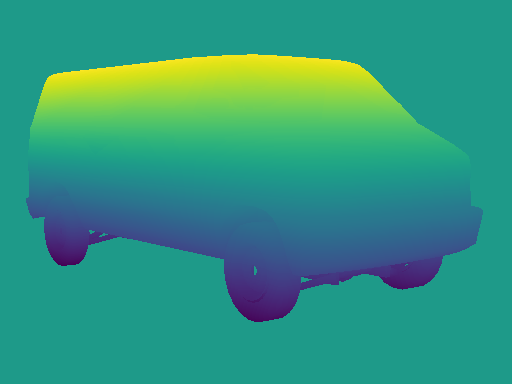}&   \colImgN{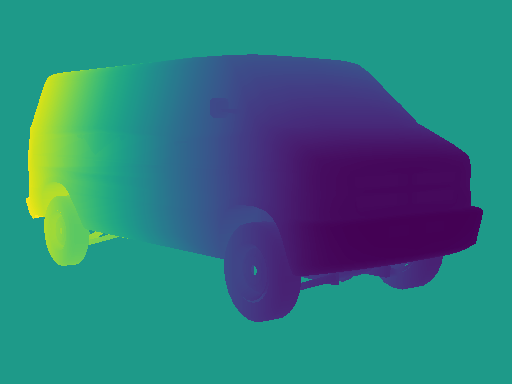}\\[-1.5pt]
		&\colImgN{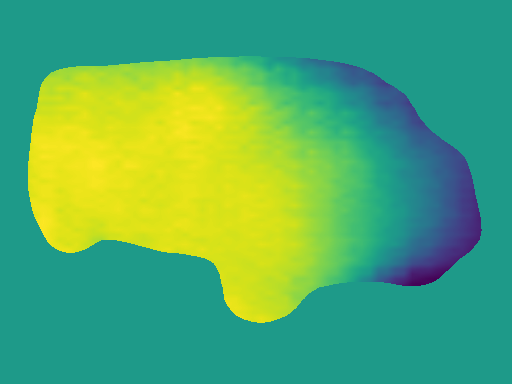}&   \colImgN{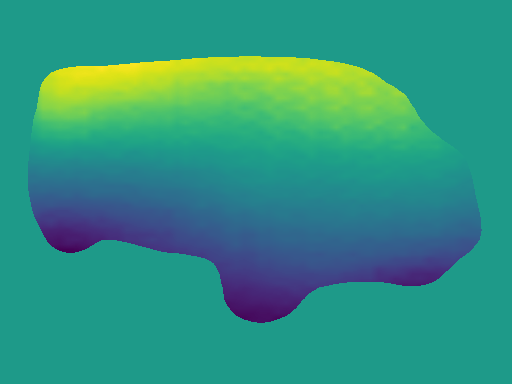}&   \colImgN{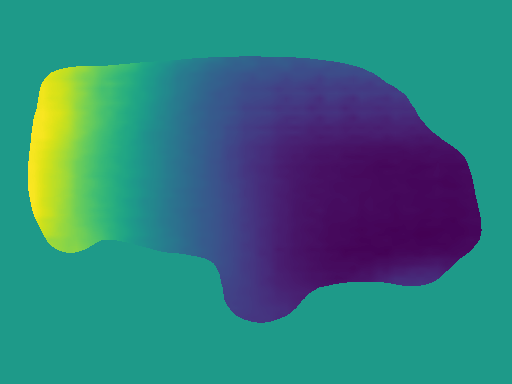}\\[-1.5pt]
		\footnotesize Image&\footnotesize LF (X)&\footnotesize LF (Y)&\footnotesize LF (Z)\\[-3pt]
	\end{tabular}
	\end{subfigure}
	\caption{Additional qualitative examples of our predicted location fields. For each example image, the top row shows the ground truth and the bottom row shows our prediction. The overall 3D shape is recovered well, but fine-grained details like the side mirrors of cars or thin structures like the frame ornaments of tables and beds are missed.}
	\label{fig:pred_lfs2}
\end{figure*}

Fig.~\ref{fig:pred_lfs2} presents further qualitative examples of our predicted location fields. We upscale and pad the predicted location fields to match the input image resolution. The overall 3D shape is recovered well in the location fields, but fine-grained details like the side mirrors of cars or thin structures like the frame ornaments of tables and beds are missed.

\begin{figure*}
	\setlength{\tabcolsep}{0pt}
	\newcommand{\colImgN}[1]{{\includegraphics[width=0.08333\linewidth]{#1}}}
	\newcommand{\colImgI}[1]{{\includegraphics[height=0.08333\linewidth]{#1}}}
	\centering
	\begin{tabular}{cc|cccccccccc}
		\colImgI{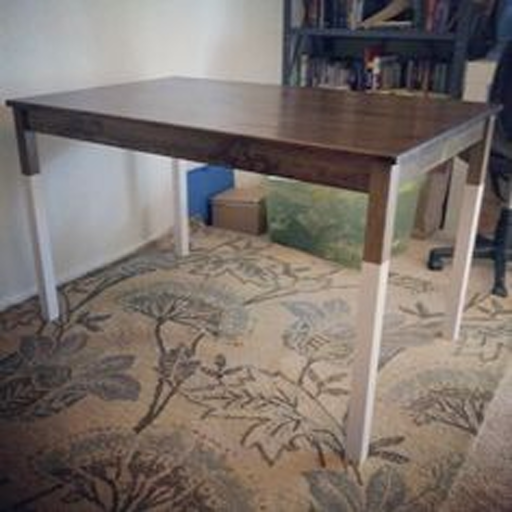}&   \colImgN{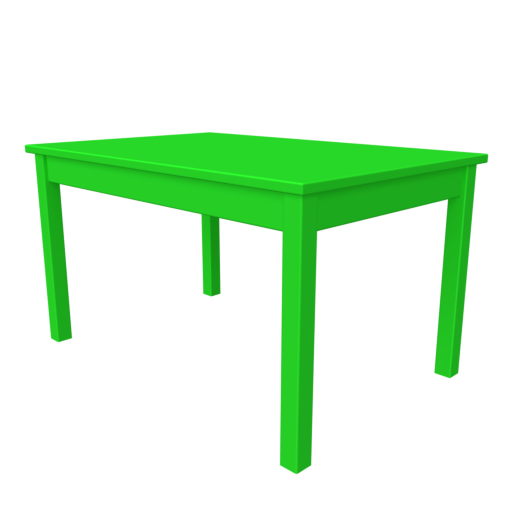}&
		\colImgN{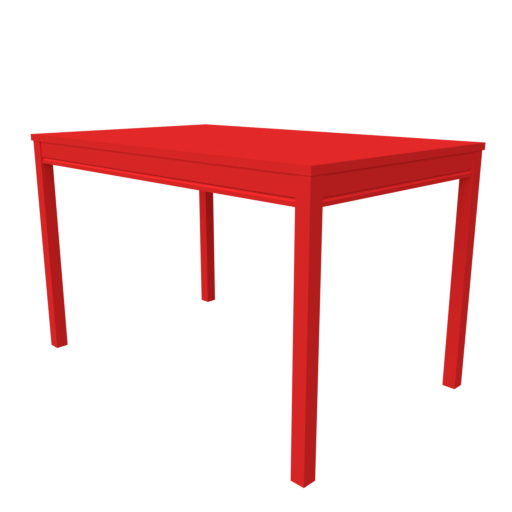}&
		\colImgN{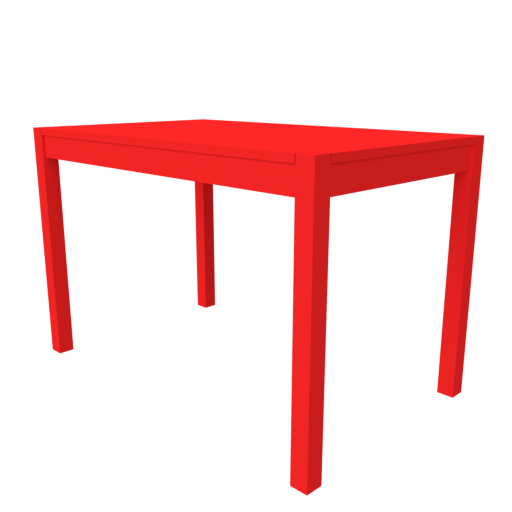}&
		\colImgN{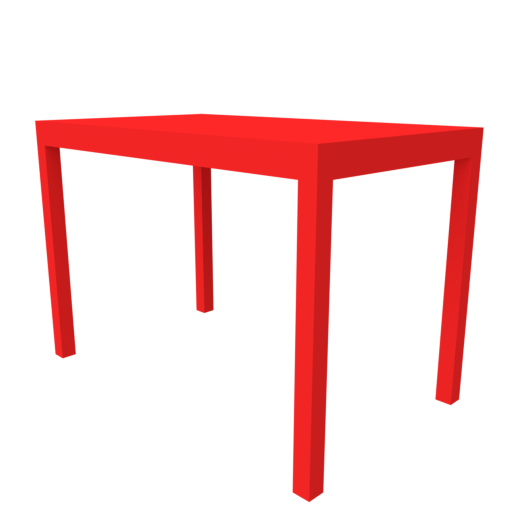}&
		\colImgN{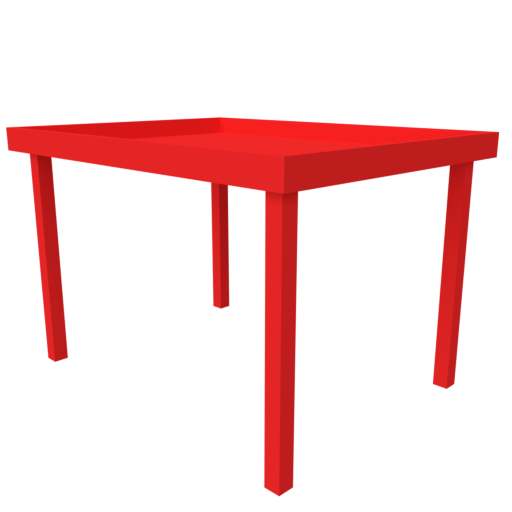}&
		\colImgN{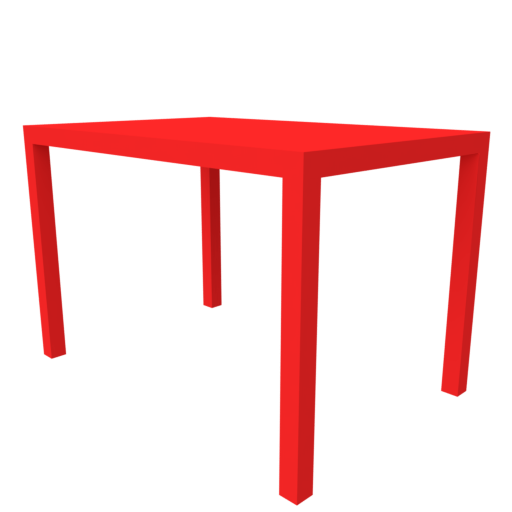}&
		\colImgN{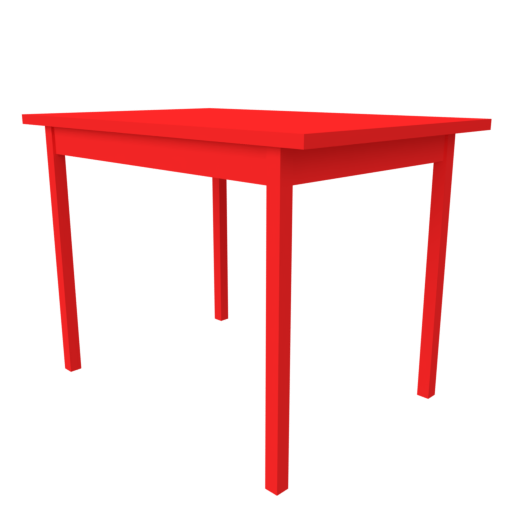}&
		\colImgN{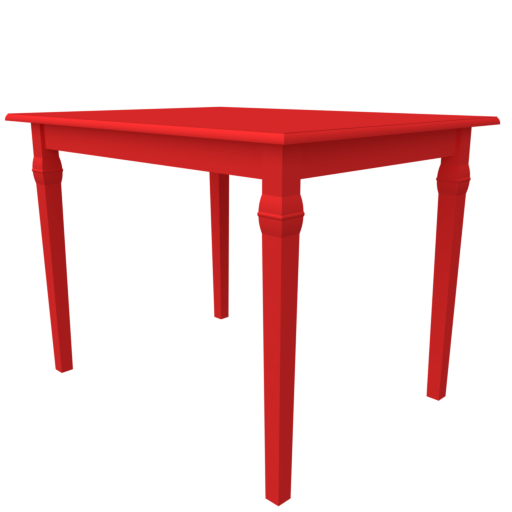}&
		\colImgN{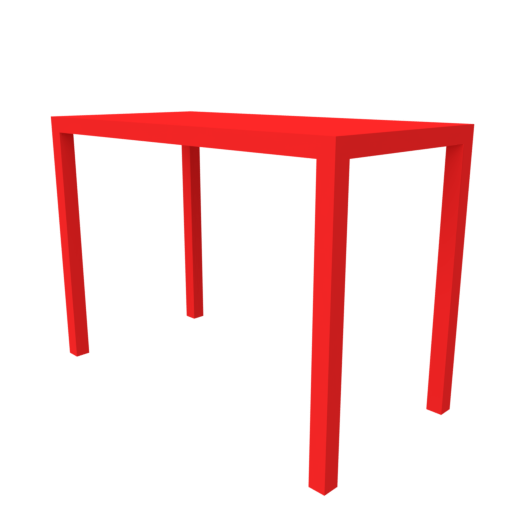}&
		\colImgN{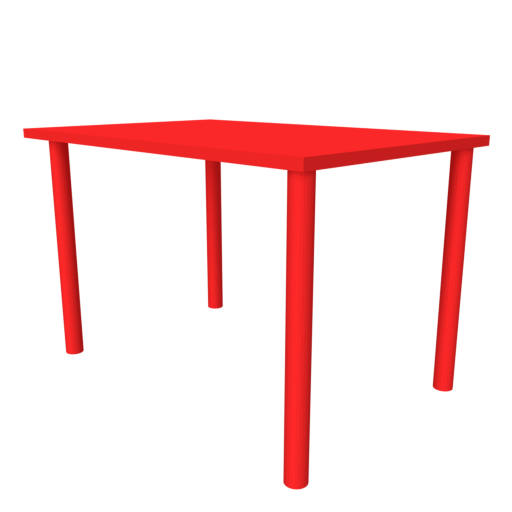}&
		\colImgN{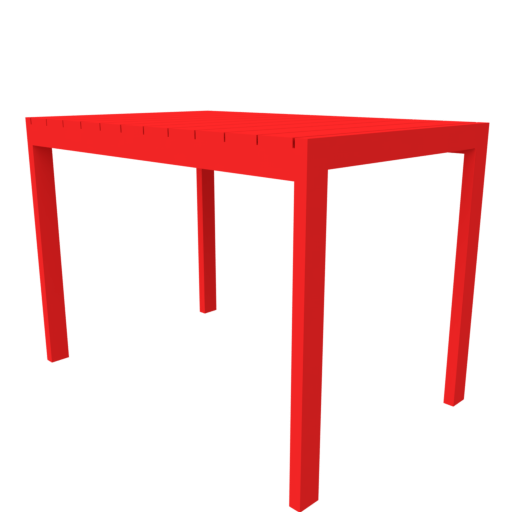}\\[-5pt]
		\colImgN{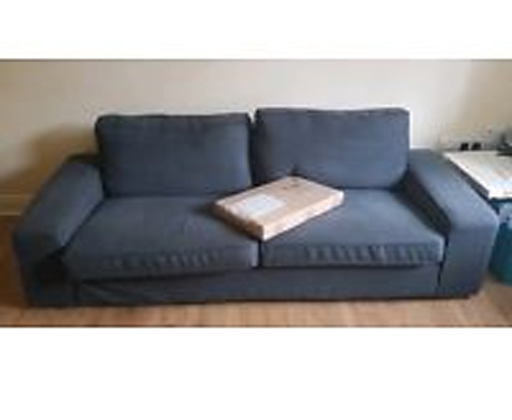}&   \colImgN{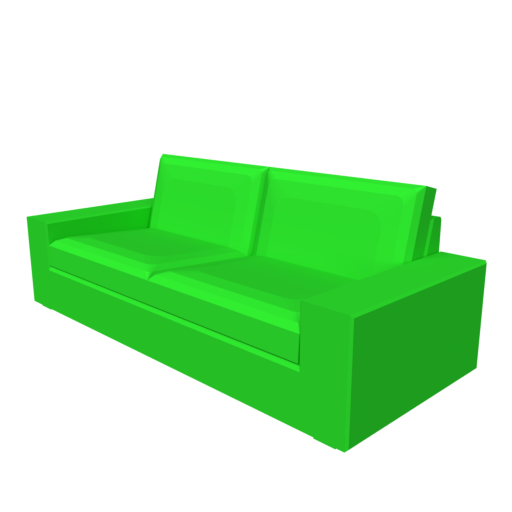}&
		\colImgN{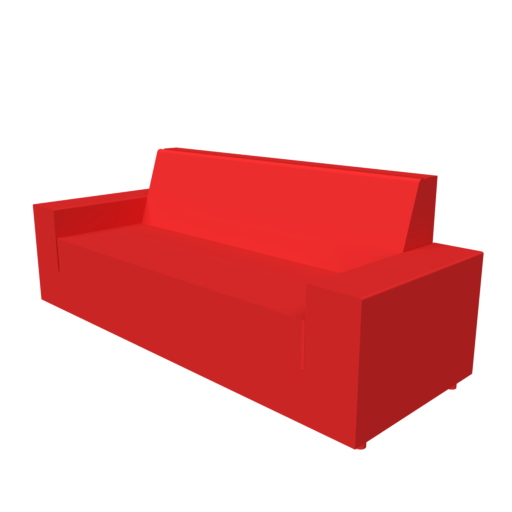}&
		\colImgN{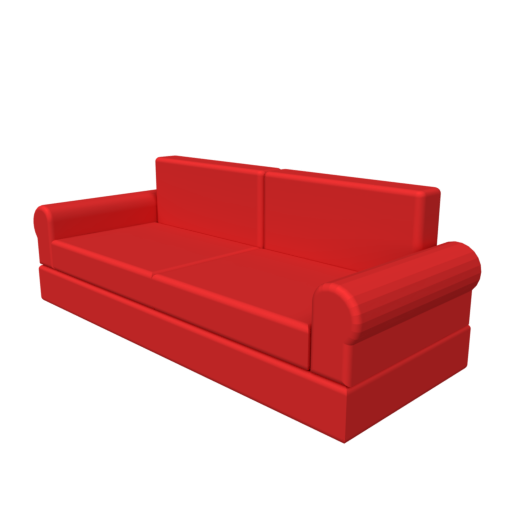}&
		\colImgN{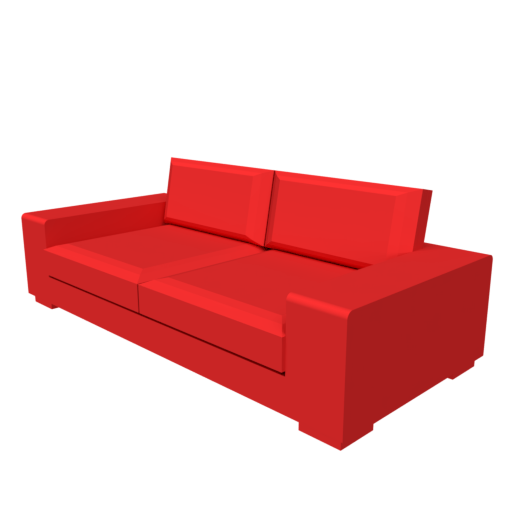}&
		\colImgN{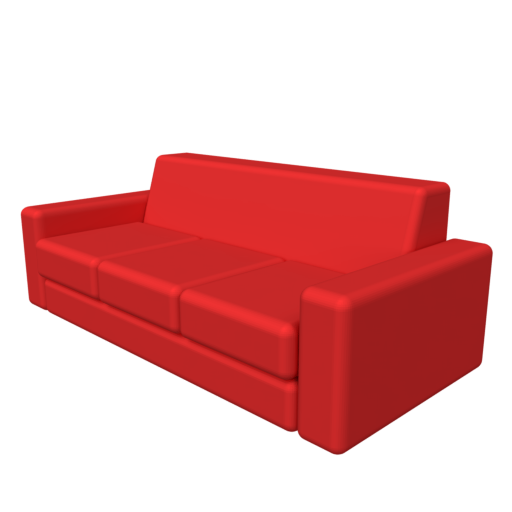}&
		\colImgN{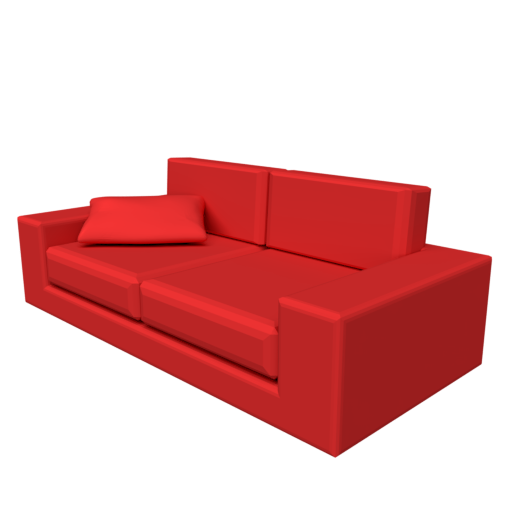}&
		\colImgN{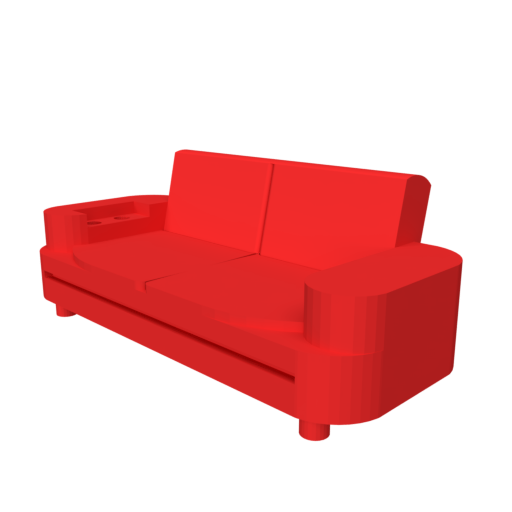}&
		\colImgN{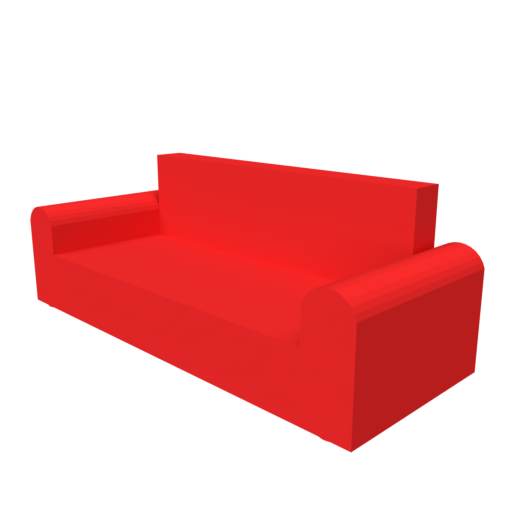}&
		\colImgN{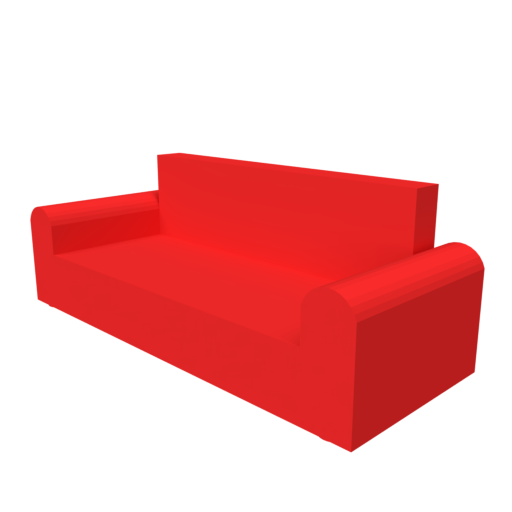}&
		\colImgN{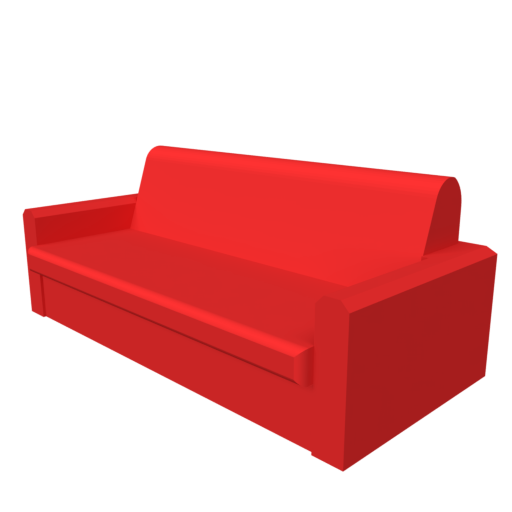}&
		\colImgN{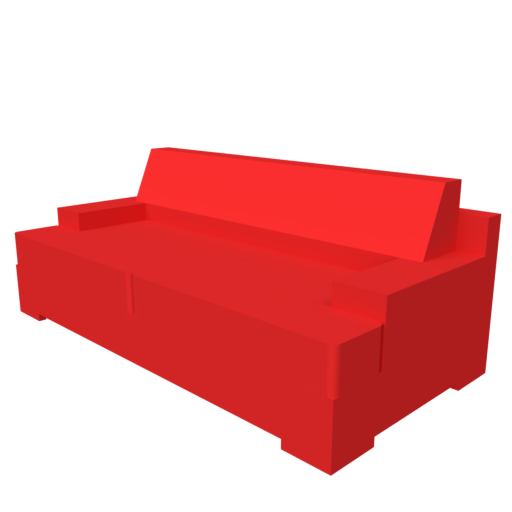}\\[0pt]
		\colImgI{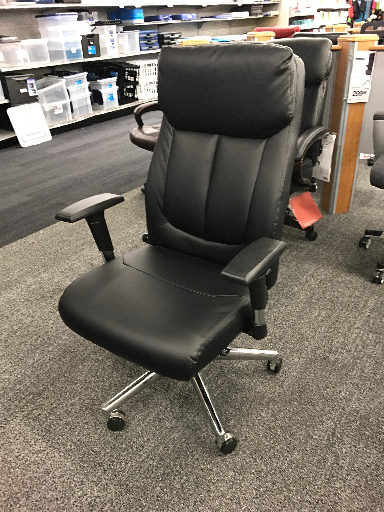}&   \colImgN{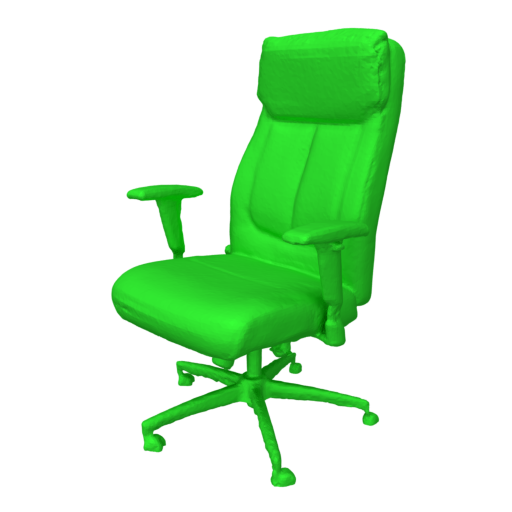}&
		\colImgN{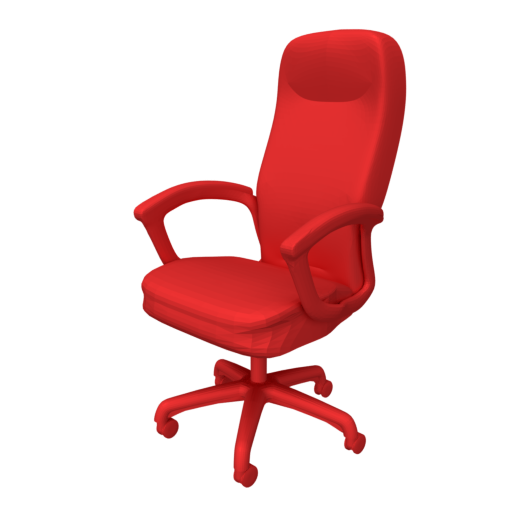}&
		\colImgN{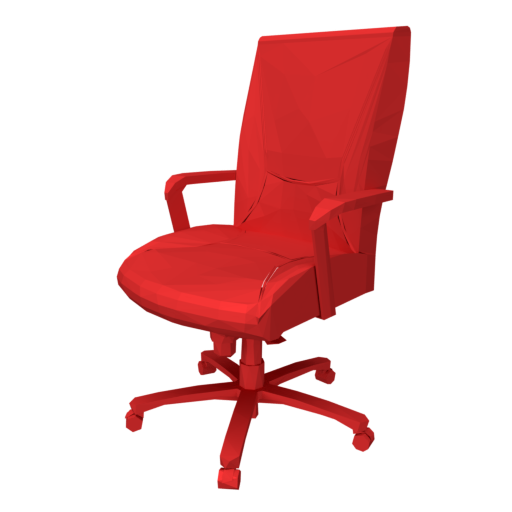}&
		\colImgN{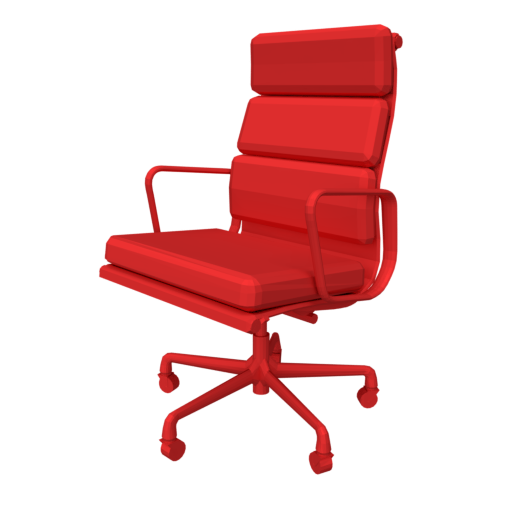}&
		\colImgN{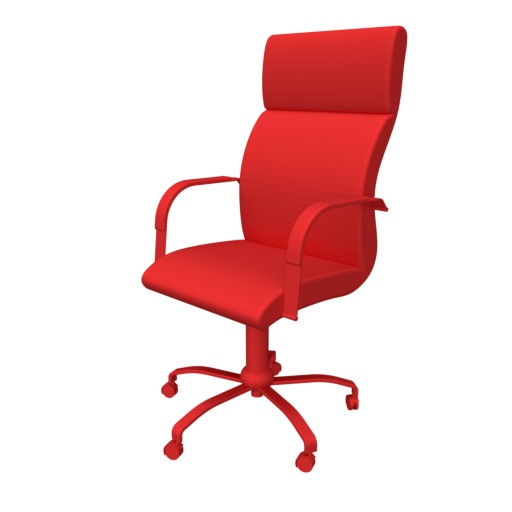}&
		\colImgN{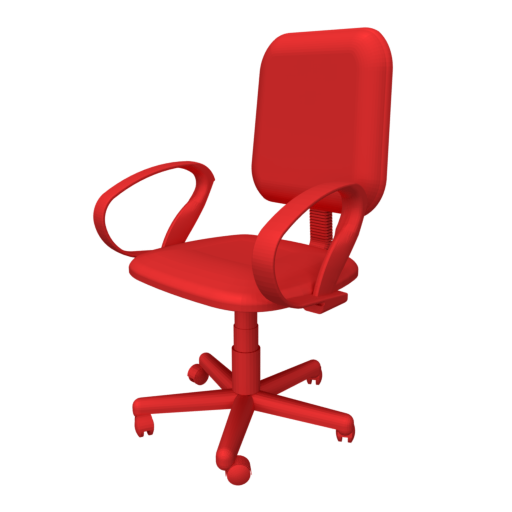}&
		\colImgN{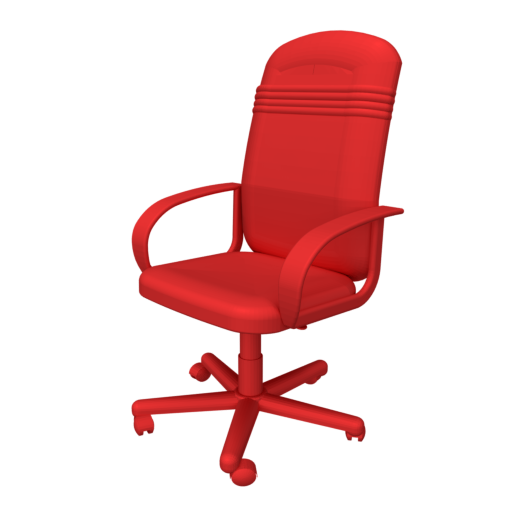}&
		\colImgN{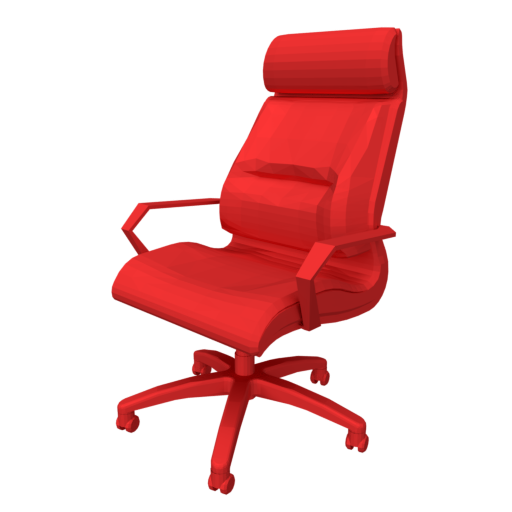}&
		\colImgN{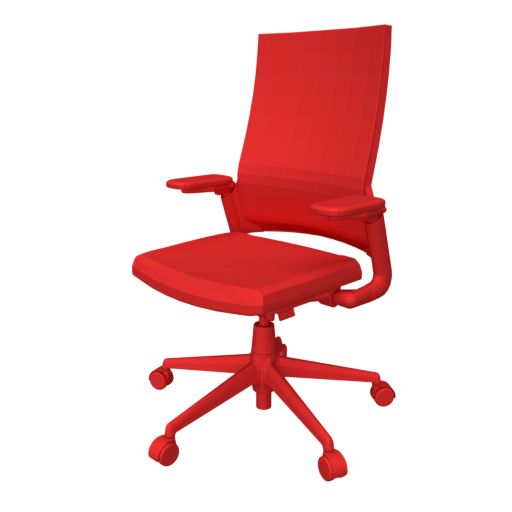}&
		\colImgN{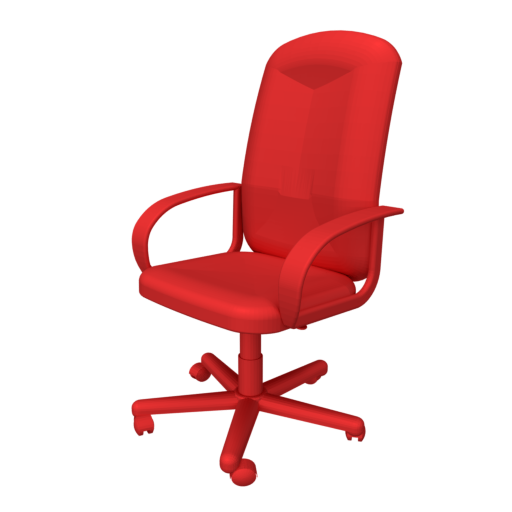}&
		\colImgN{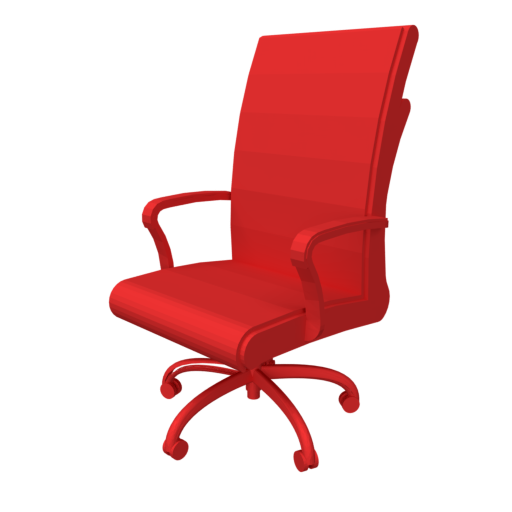}\\[-5pt]
		\colImgN{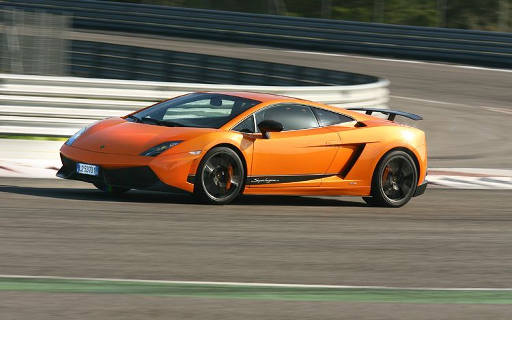}&
		\colImgN{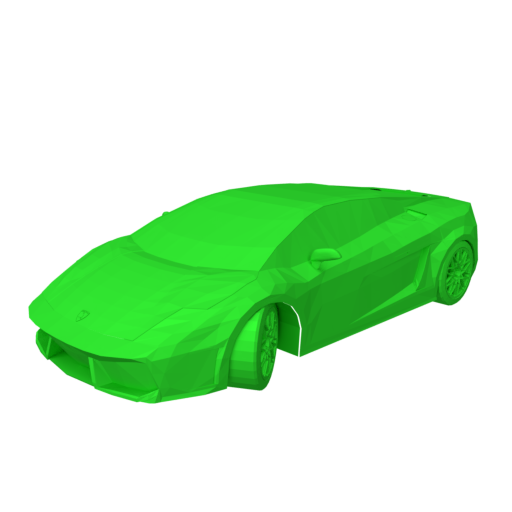}&
		\colImgN{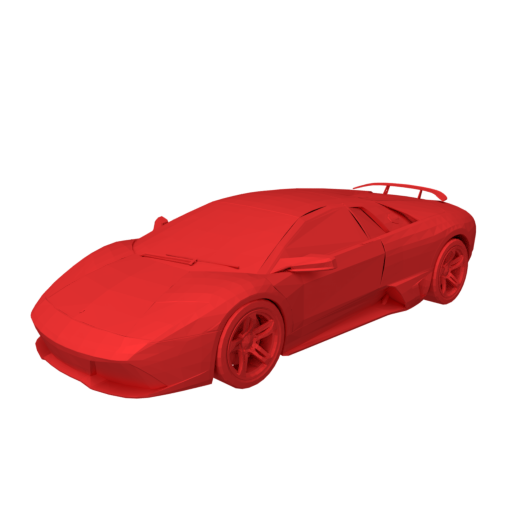}&
		\colImgN{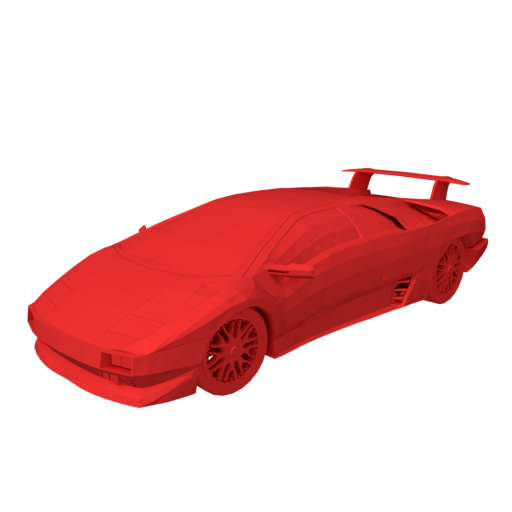}&
		\colImgN{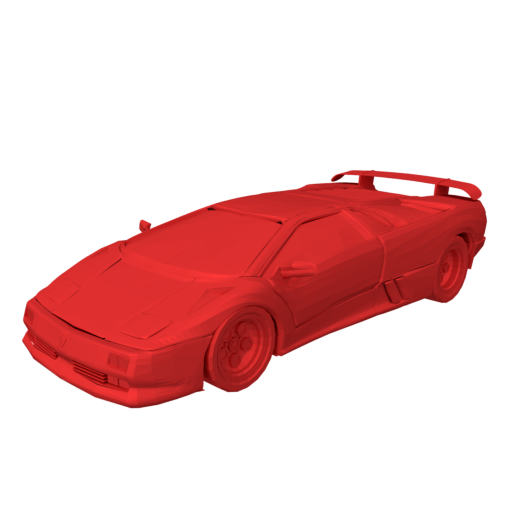}&
		\colImgN{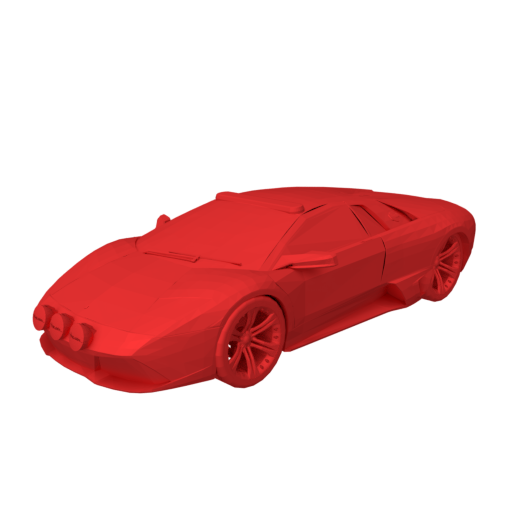}&
		\colImgN{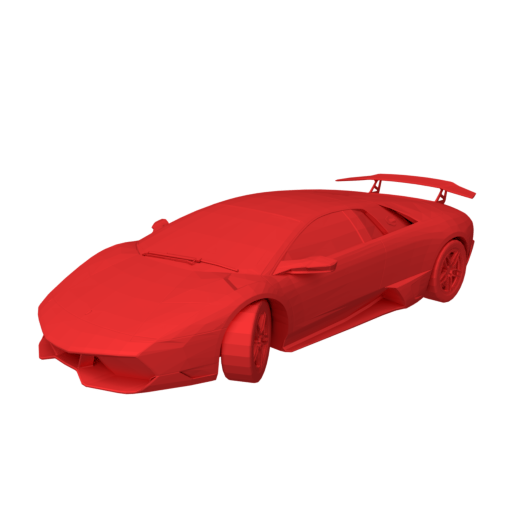}&
		\colImgN{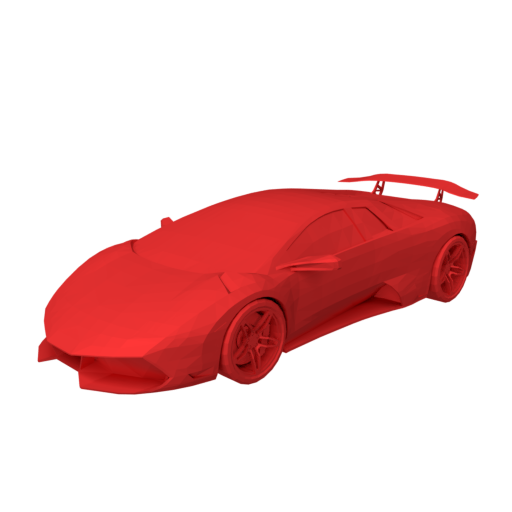}&
		\colImgN{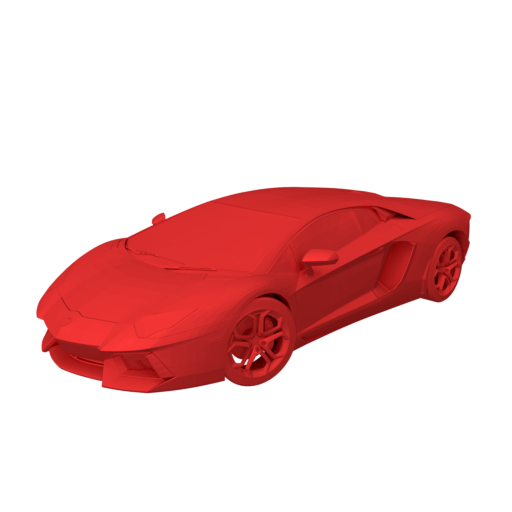}&
		\colImgN{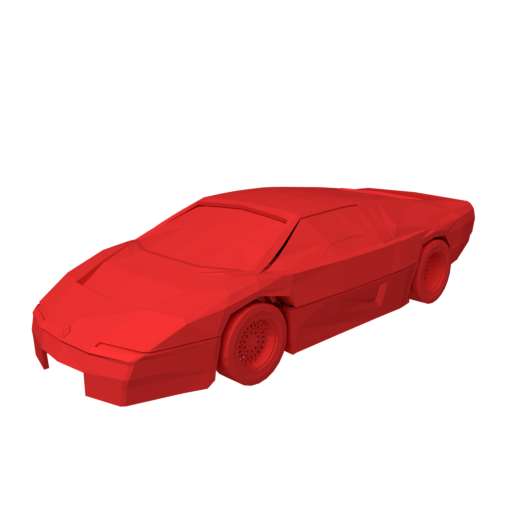}&
		\colImgN{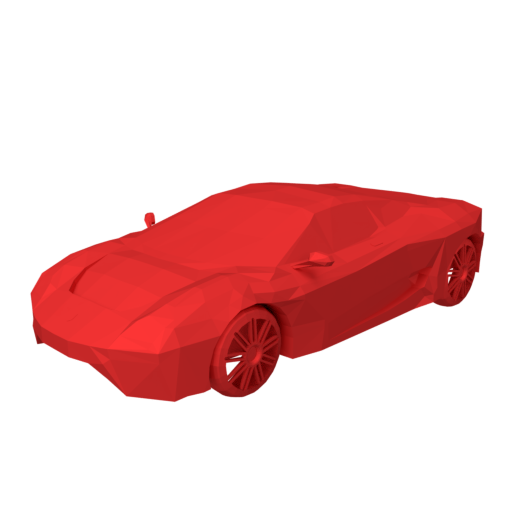}&
		\colImgN{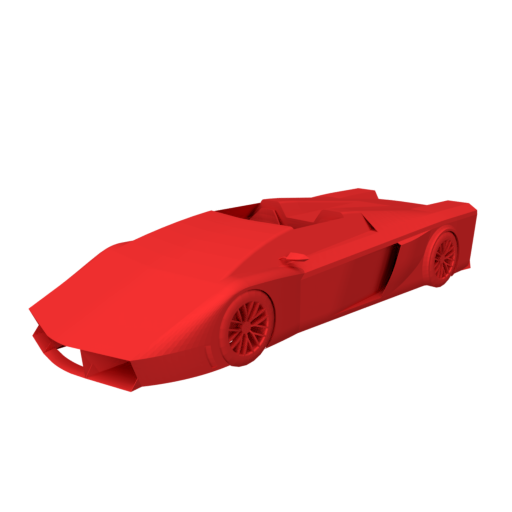}\\[-1.5pt]
		\footnotesize Image&\footnotesize GT&\footnotesize 1&\footnotesize 2&\footnotesize 3&\footnotesize 4&\footnotesize 5&\footnotesize 6&\footnotesize 7&\footnotesize 8&\footnotesize 9&\footnotesize 10\\[-3pt]
	\end{tabular}
	\caption{Additional qualitative results for 3D model retrieval from ShapeNet. From left to right, we show the input image, the ground truth 3D model and the top ten ranked 3D models. The overall 3D shape of the retrieved models is consistent and accurate.}
	\label{fig:top10C}
\end{figure*}

Fig.~\ref{fig:top10C} shows additional qualitative results for 3D model retrieval from ShapeNet. Considering the top ten ranked 3D models, we observe that the retrieved models have a consistent and accurate overall 3D shape and geometry. 

\begin{figure*}
	\setlength{\tabcolsep}{1pt}
	\setlength{\fboxsep}{-2pt}
	\setlength{\fboxrule}{2pt}
	\newcommand{\colImgN}[1]{{\includegraphics[width=0.24\linewidth]{#1}}}
	\centering
	\begin{subfigure}[t]{\columnwidth}
	\begin{tabular}{cccc}
		\colImgN{}&   \colImgN{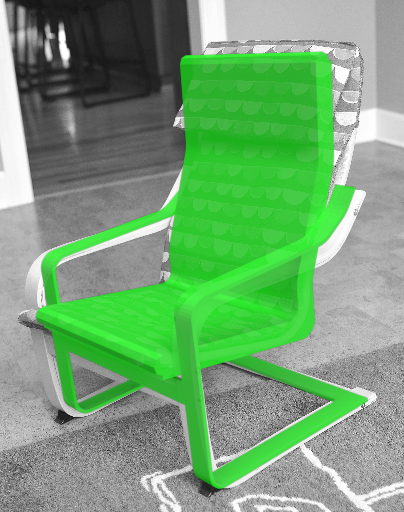}&
		\colImgN{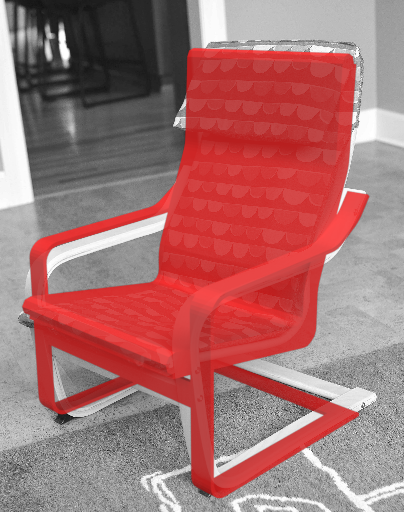}&  \colImgN{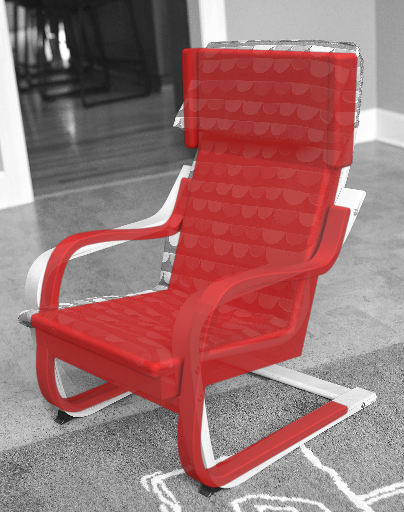}\\[-1.5pt]
		
		\colImgN{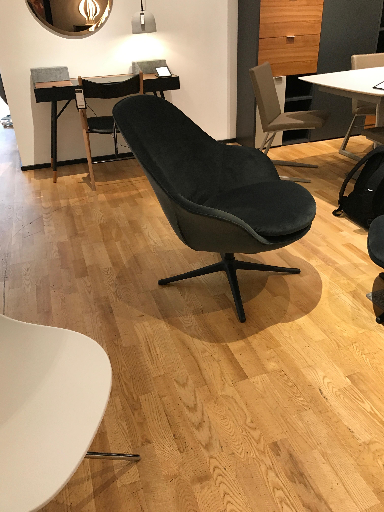}&   \colImgN{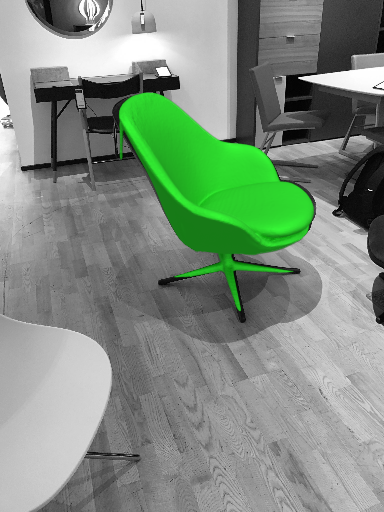}&
		\colImgN{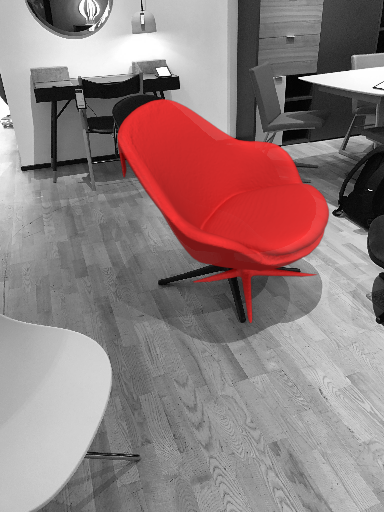}&  \colImgN{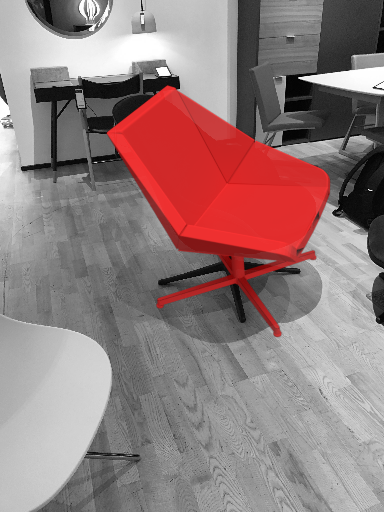}\\[-1.5pt]
		
		
		\colImgN{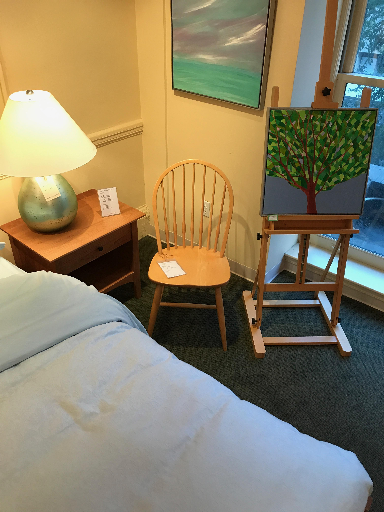}&   \colImgN{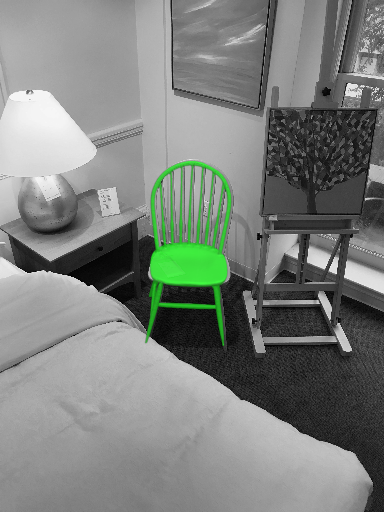}&
		\colImgN{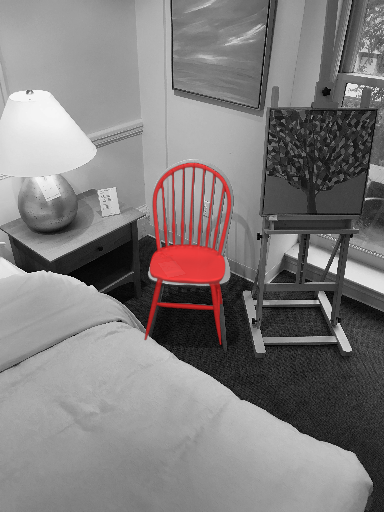}&  \colImgN{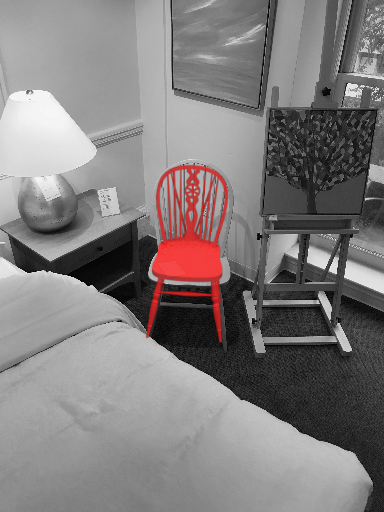}\\[-1.5pt]
		
		\colImgN{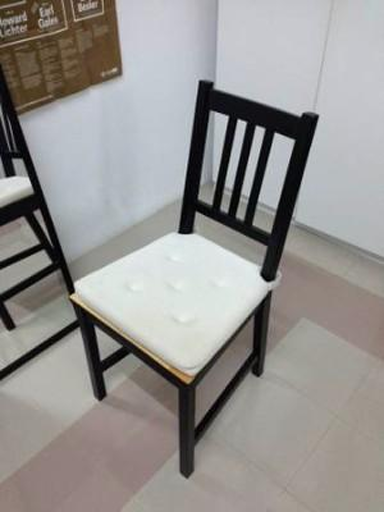}&   \colImgN{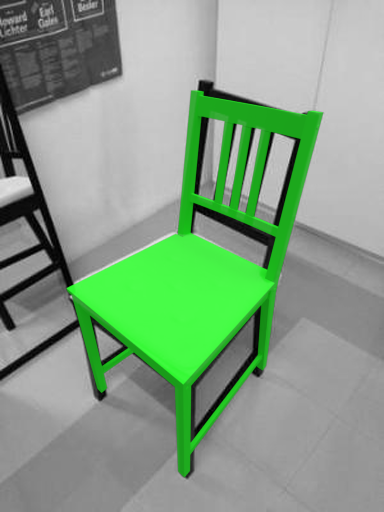}&
		\colImgN{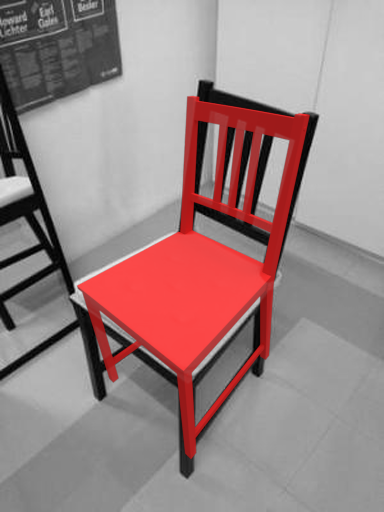}&  \colImgN{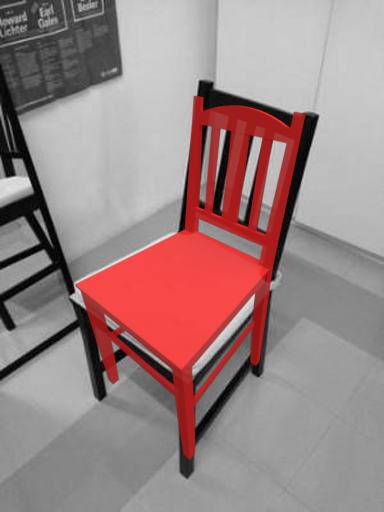}\\[-1.5pt]
		
		\colImgN{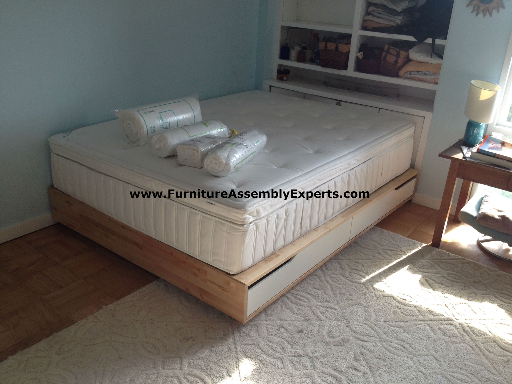}&   \colImgN{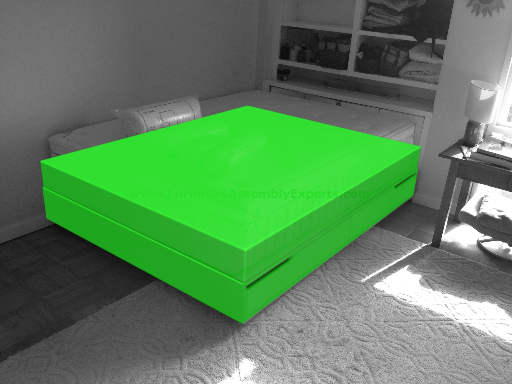}&
		\colImgN{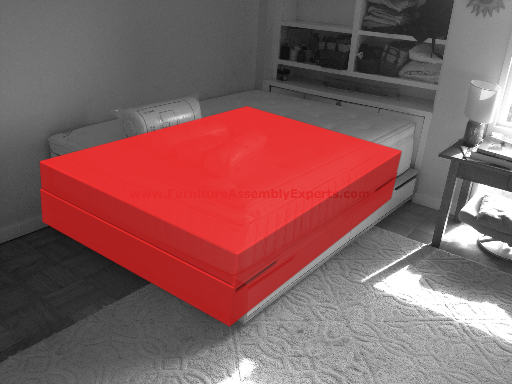}&  \colImgN{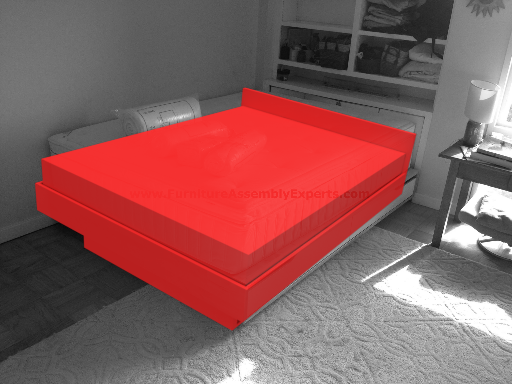}\\[-1.5pt]
		
		\colImgN{}&   \colImgN{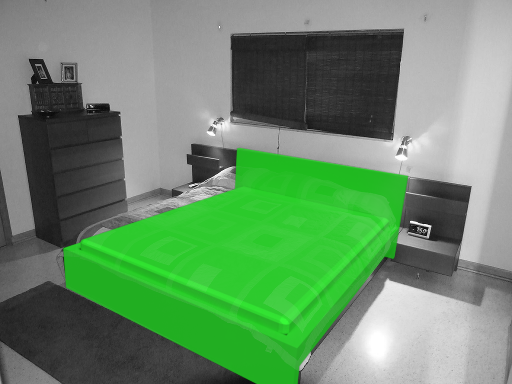}&
		\colImgN{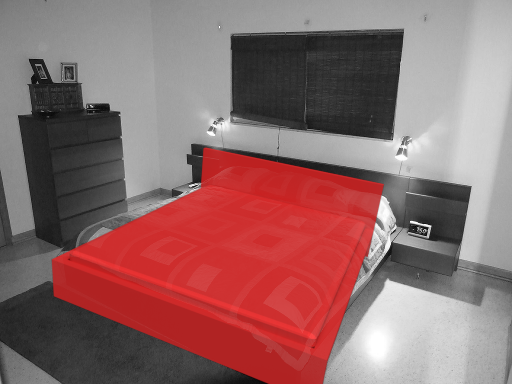}&  \colImgN{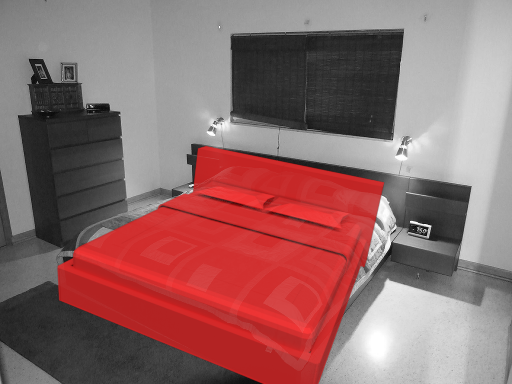}\\[-1.5pt]
		
		\colImgN{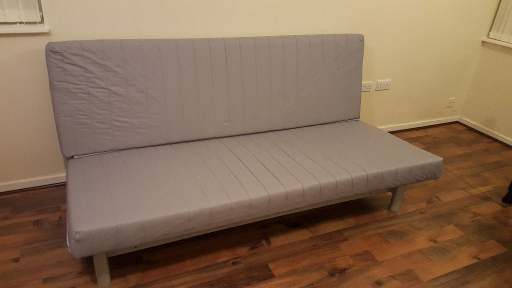}&   \colImgN{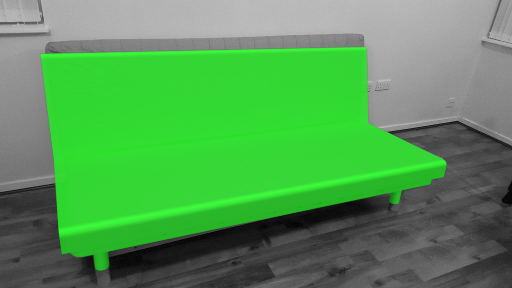}&
		\colImgN{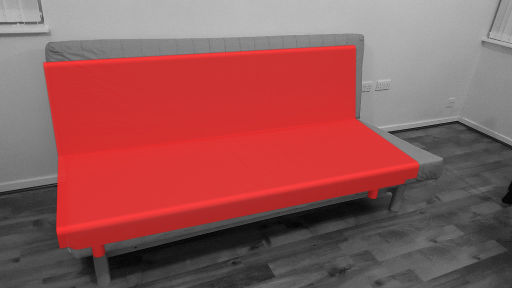}&  \colImgN{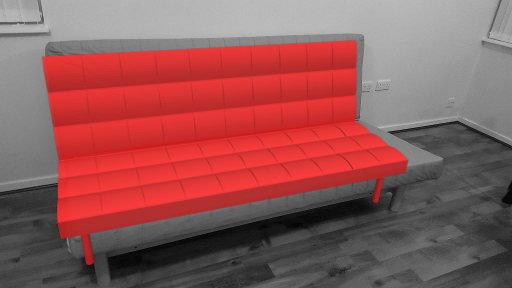}\\[-1.5pt]
		
		\footnotesize Image&\footnotesize GT&\footnotesize from seen&\footnotesize from unseen\\[-3pt]
	\end{tabular}
	\end{subfigure}\hfill\begin{subfigure}[t]{\columnwidth}
	\begin{tabular}{cccc}
		
		\colImgN{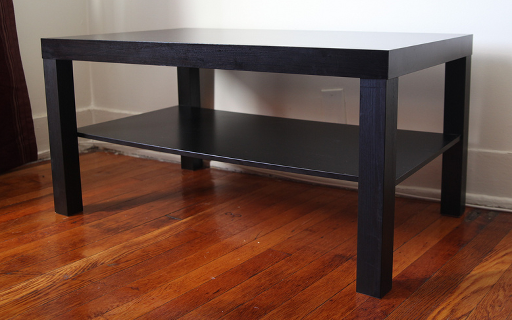}&   \colImgN{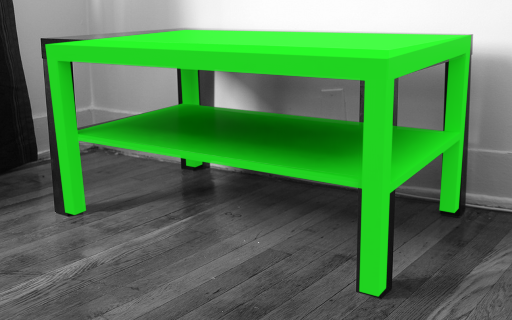}&
		\colImgN{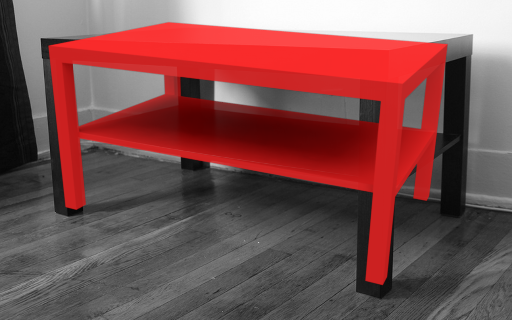}&  \colImgN{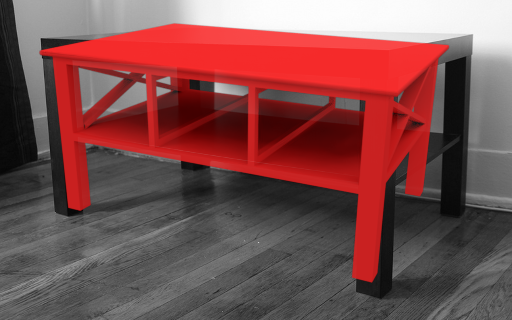}\\[-1.5pt]
				
		\colImgN{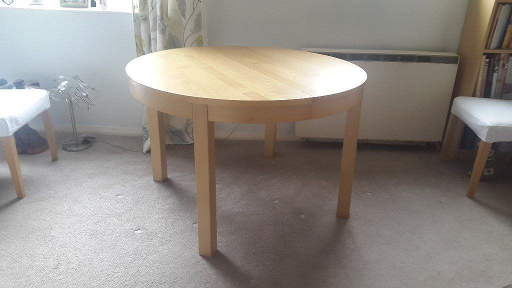}&   \colImgN{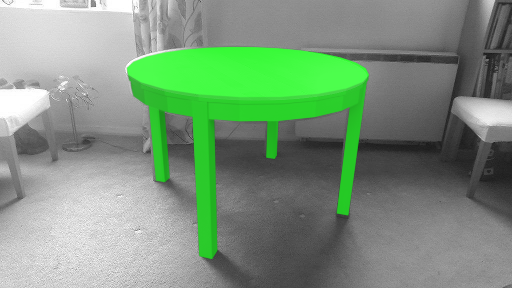}&
		\colImgN{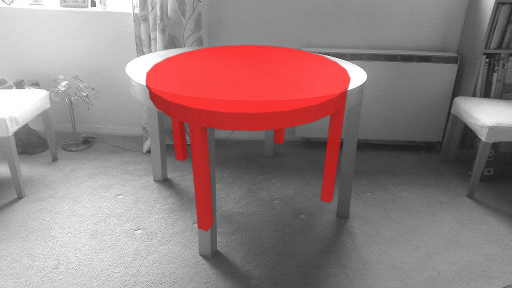}&  \colImgN{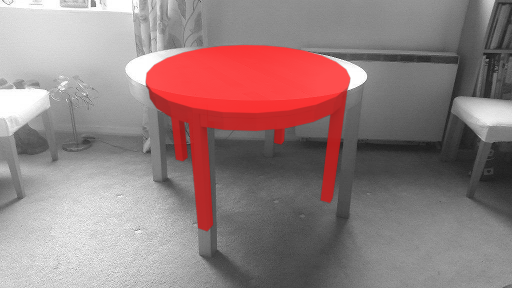}\\[-1.5pt]
		
		\colImgN{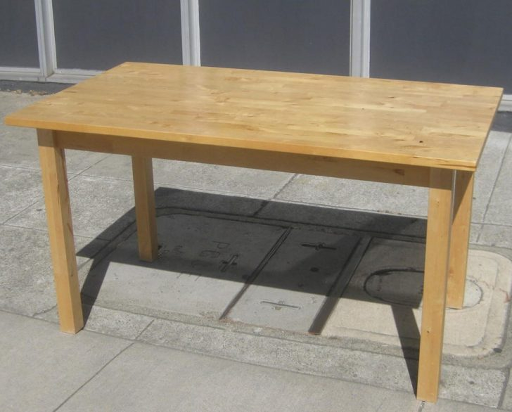}&   \colImgN{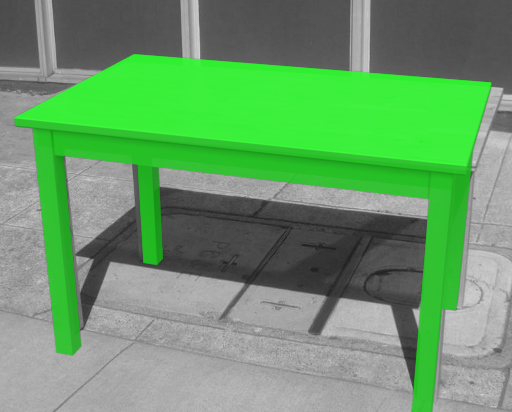}&
		\colImgN{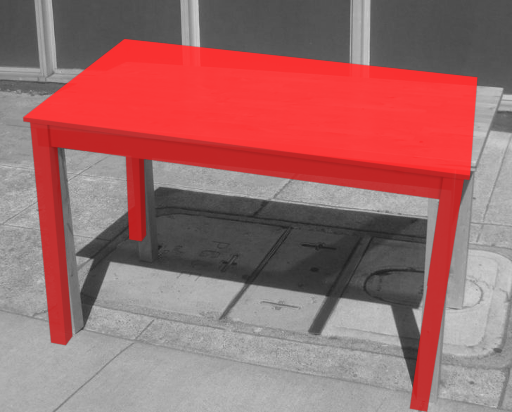}&  \colImgN{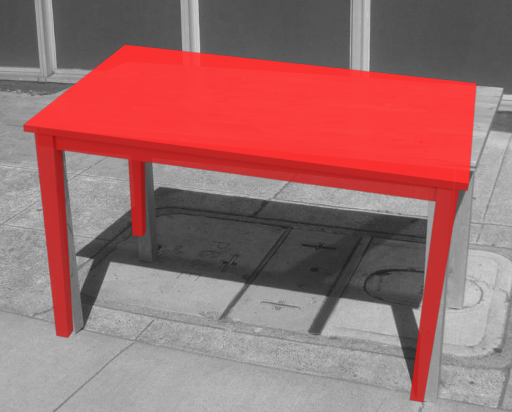}\\[-1.5pt]
		
		\colImgN{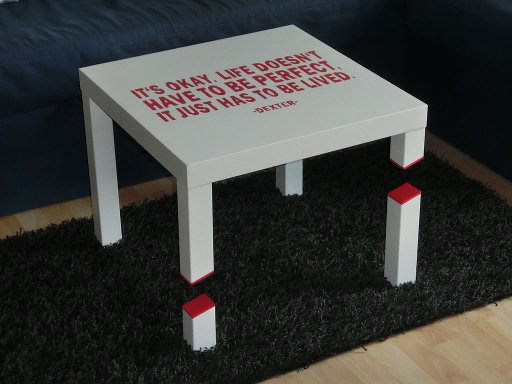}&   \colImgN{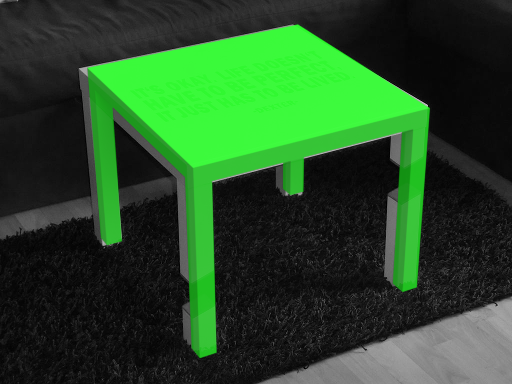}&
		\colImgN{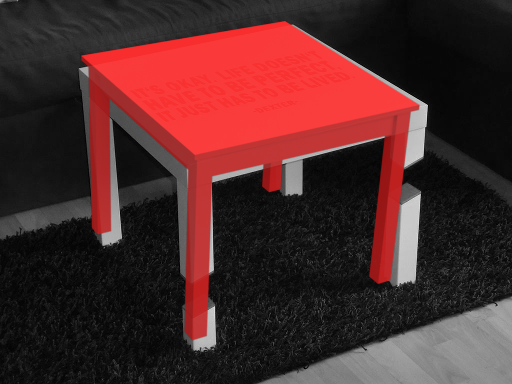}&  \colImgN{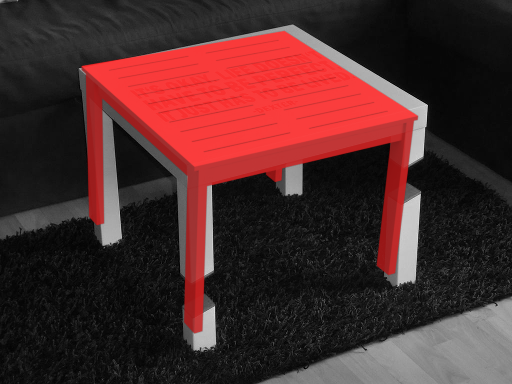}\\[-1.5pt]
		
		\colImgN{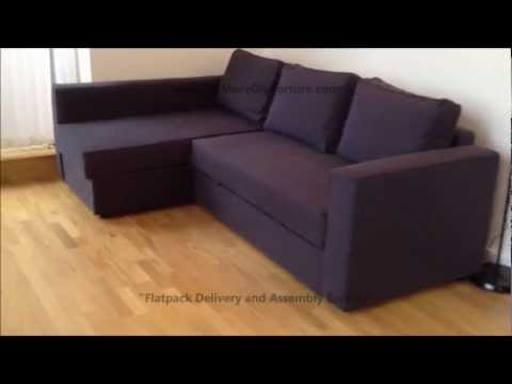}&   \colImgN{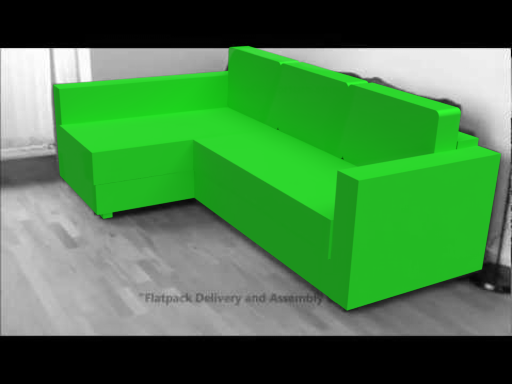}&
		\colImgN{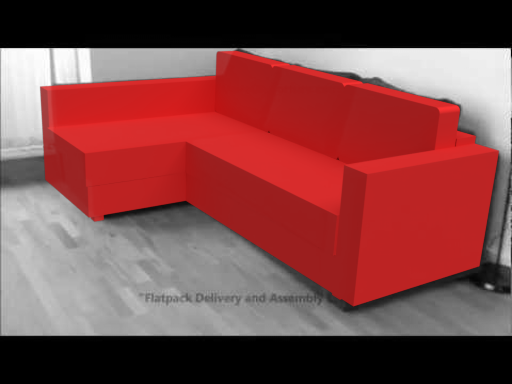}&  \colImgN{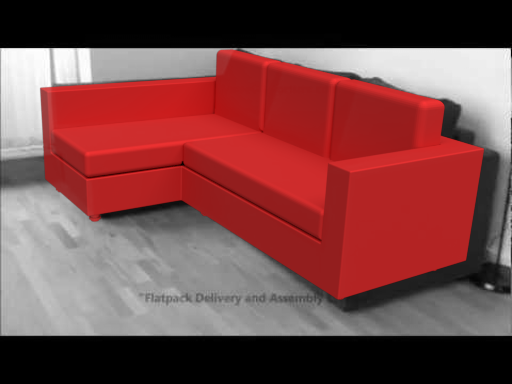}\\[-1.5pt]
		
		\colImgN{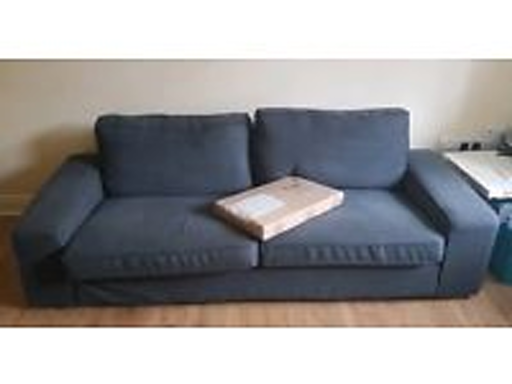}&   \colImgN{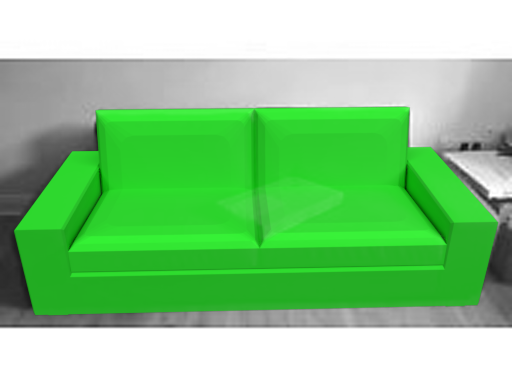}&
		\colImgN{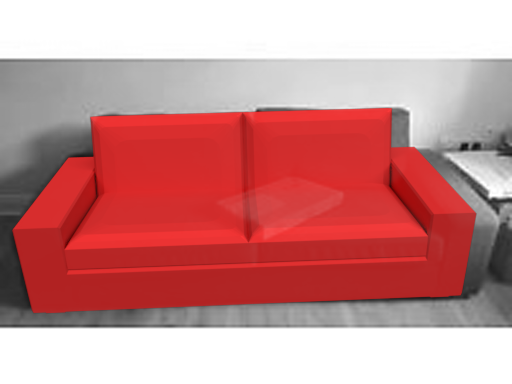}&  \colImgN{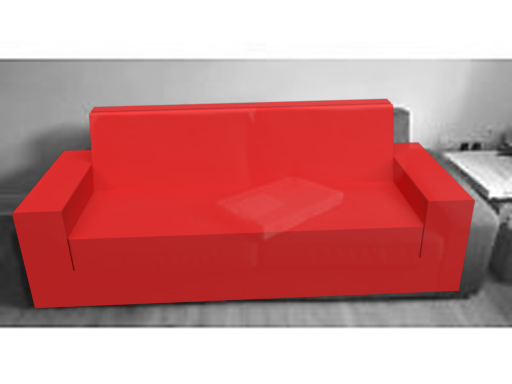}\\[-1.5pt]
		
		\colImgN{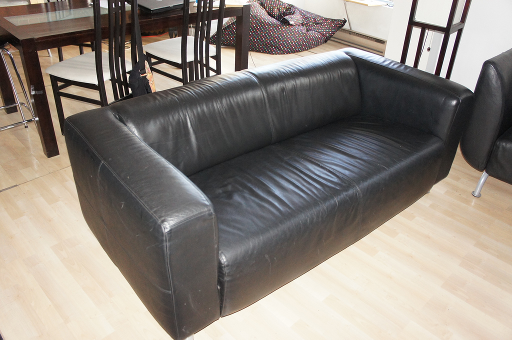}&   \colImgN{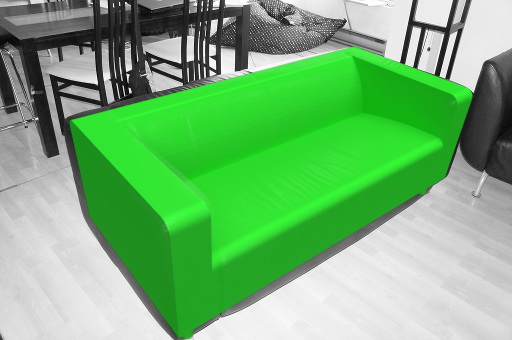}&
		\colImgN{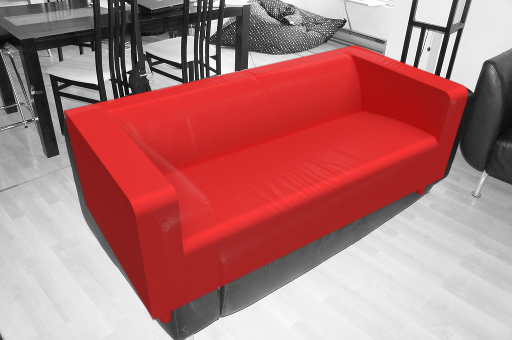}&  \colImgN{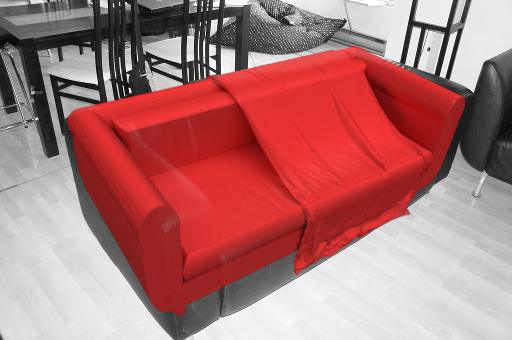}\\[-1.5pt]
		
		\colImgN{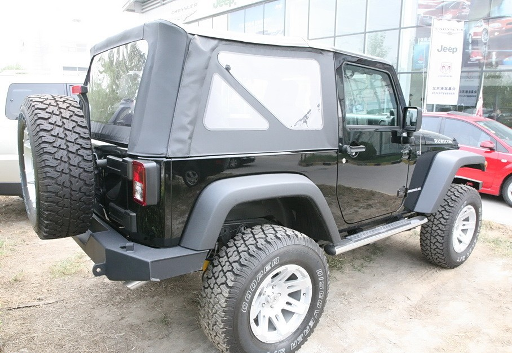}&   \colImgN{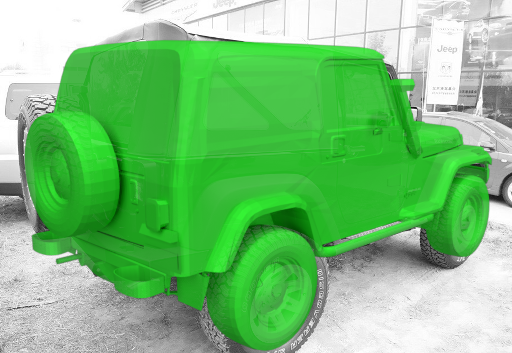}&
		\colImgN{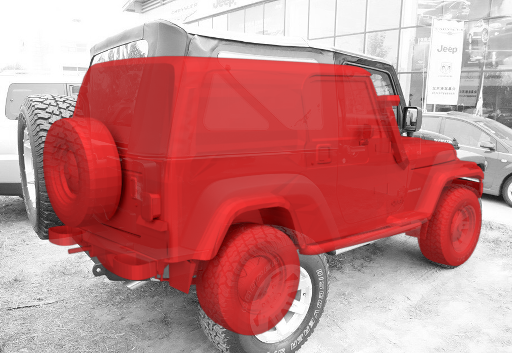}&  \colImgN{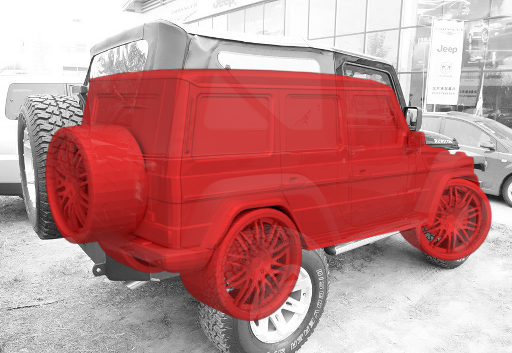}\\[-1.5pt]
		
		\colImgN{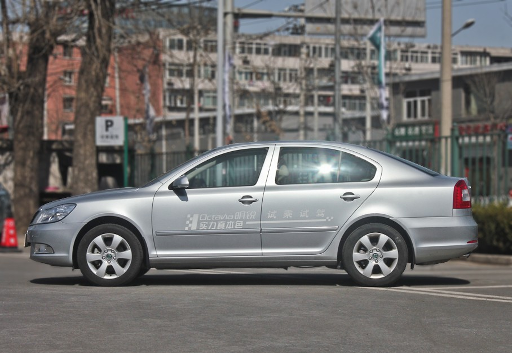}&   \colImgN{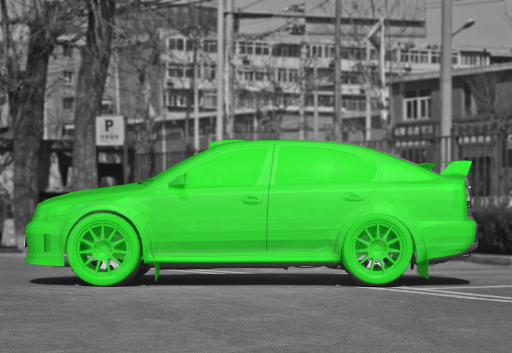}&
		\colImgN{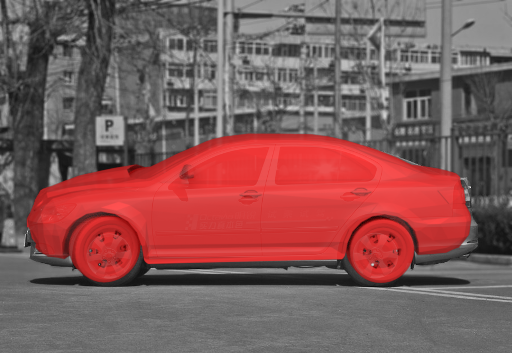}&  \colImgN{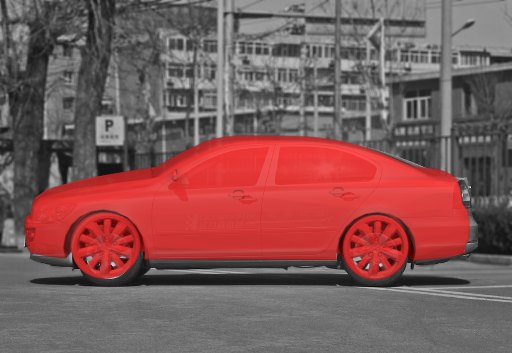}\\[-1.5pt]
		
		\colImgN{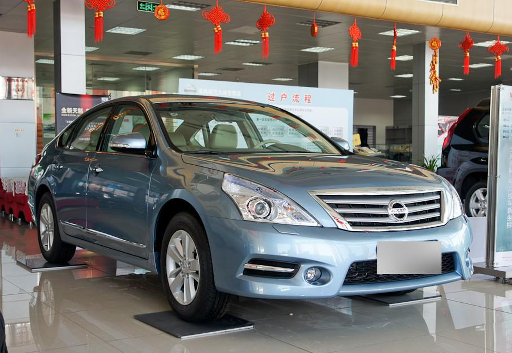}&   \colImgN{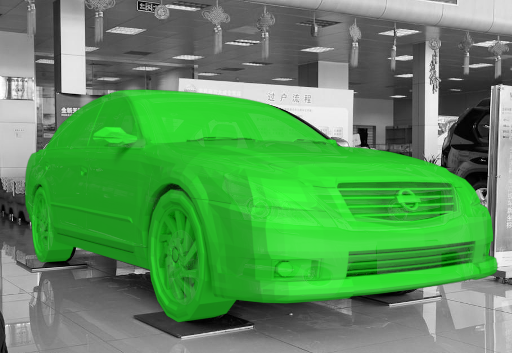}&
		\colImgN{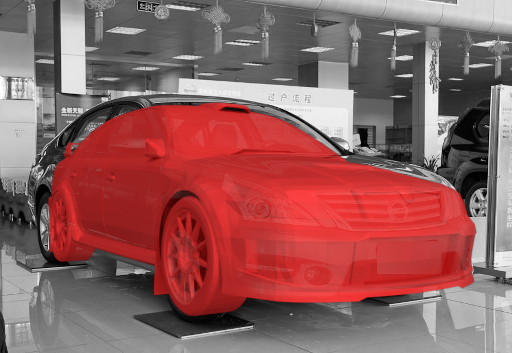}&  \colImgN{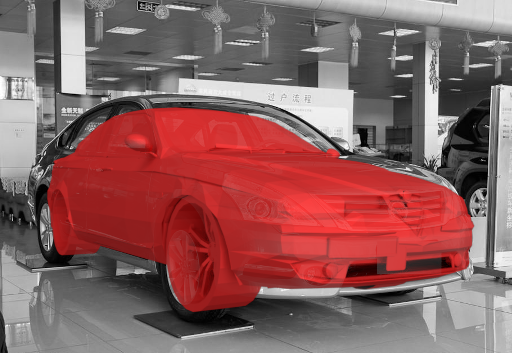}\\[-1.5pt]
		
		\colImgN{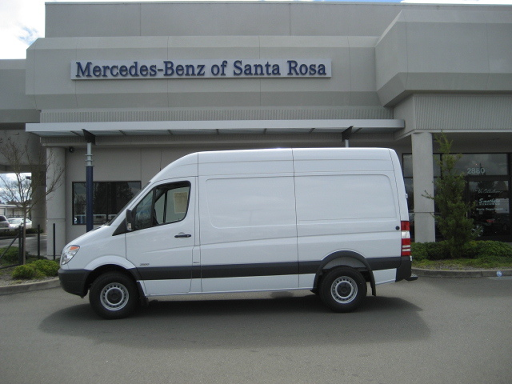}&   \colImgN{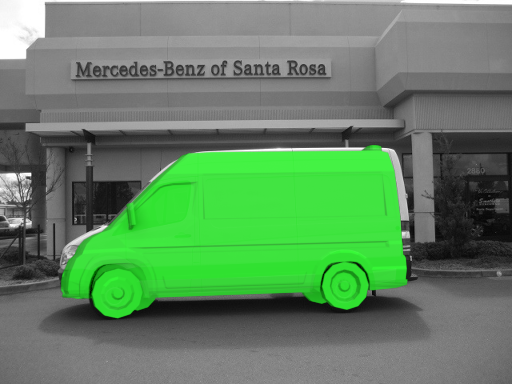}&
		\colImgN{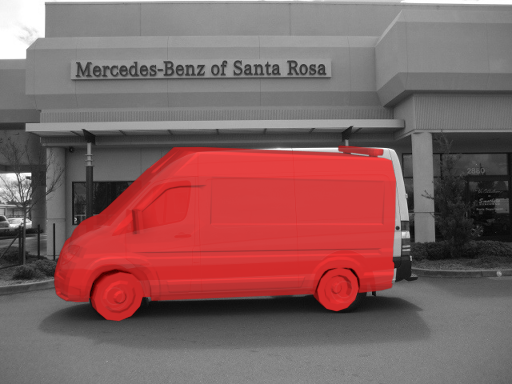}&  \colImgN{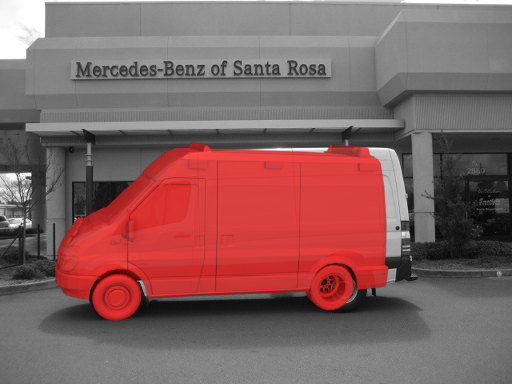}\\[-1.5pt]
		
		\footnotesize Image&\footnotesize GT&\footnotesize from seen&\footnotesize from unseen\\[-3pt]
	\end{tabular}
	\end{subfigure}
	\caption{Additional qualitative results for 3D pose estimation and 3D model retrieval from both seen and unseen databases. We project the retrieved 3D model onto the image using a 3D pose computed from the predicted location field by solving a \PNP~problem. For the ground truth 3D model, we use the ground truth 3D pose. In fact, location fields provide all relevant information to jointly address both tasks.}
	\label{fig:ret_pose1}
\end{figure*}

\begin{figure*}
	\setlength{\tabcolsep}{1pt}
	\setlength{\fboxsep}{-2pt}
	\setlength{\fboxrule}{2pt}
	\newcommand{\colImgN}[1]{{\includegraphics[width=0.24\linewidth]{#1}}}
	\centering
	\begin{subfigure}[t]{\columnwidth}
	\begin{tabular}{cccc}
		
		\colImgN{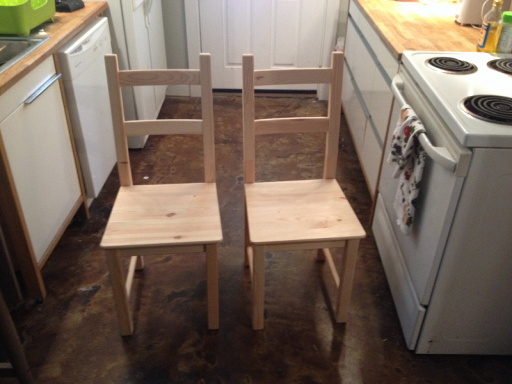}&   \colImgN{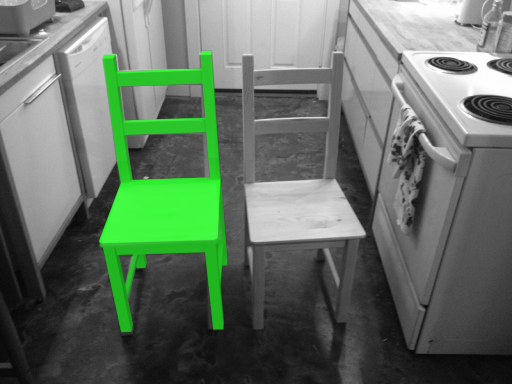}&
		\colImgN{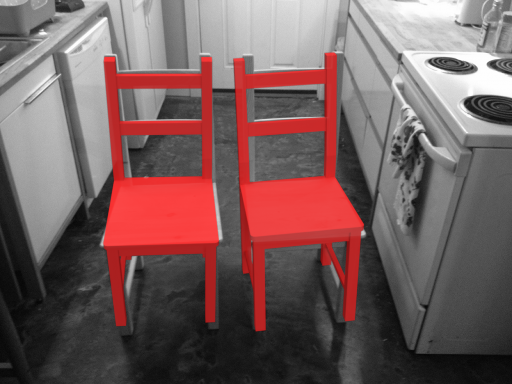}&  \colImgN{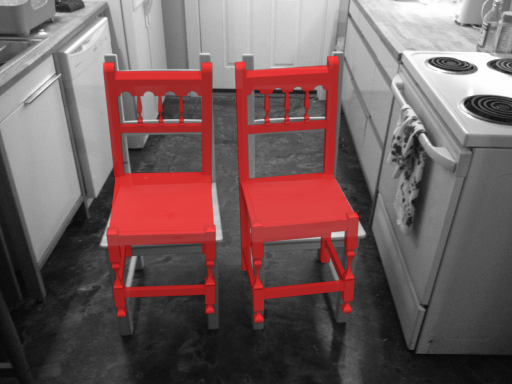}\\[-1.5pt]
		
		\colImgN{}&   \colImgN{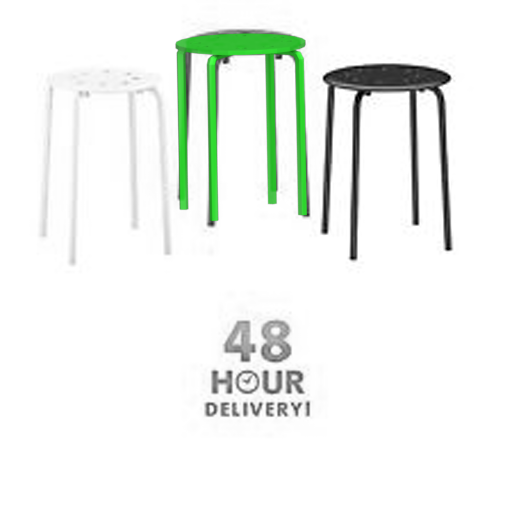}&
		\colImgN{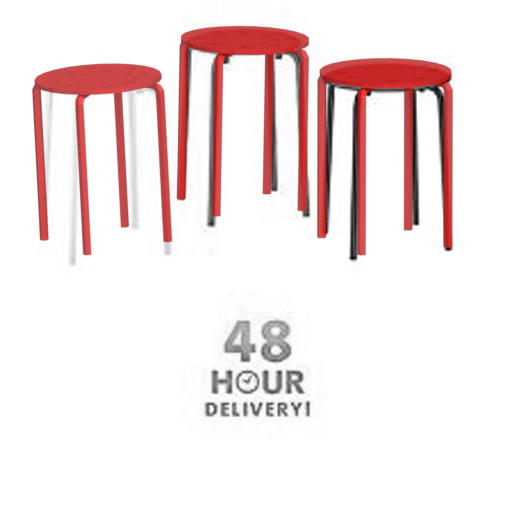}&  \colImgN{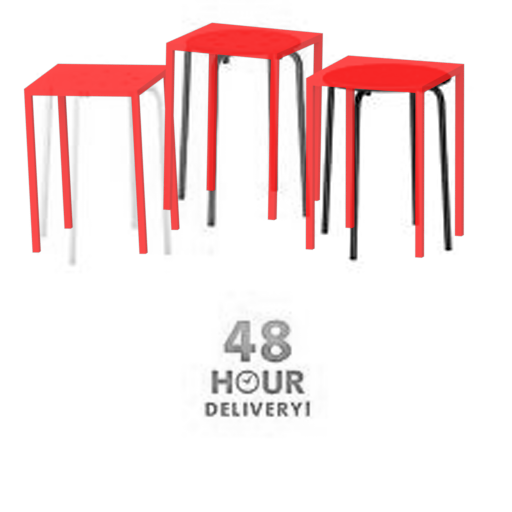}\\[-1.5pt]
		
		\colImgN{}&   \colImgN{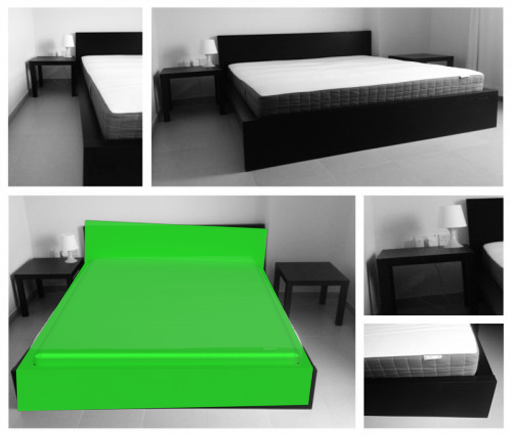}&
		\colImgN{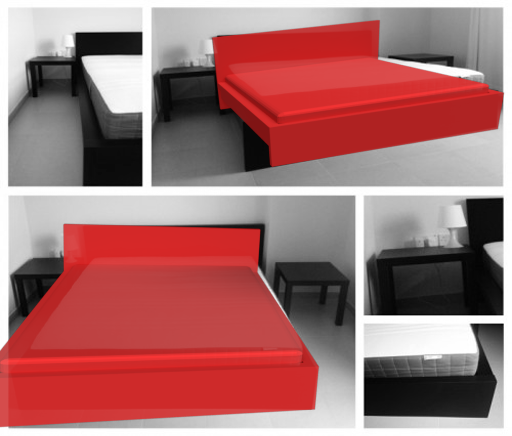}&  \colImgN{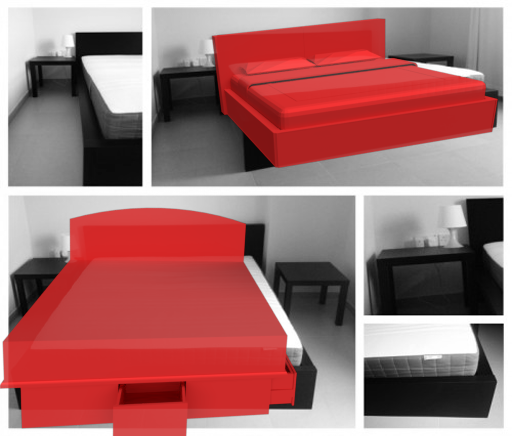}\\[-1.5pt]
		
		\colImgN{}&   \colImgN{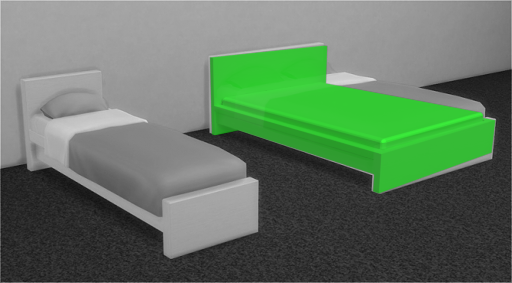}&
		\colImgN{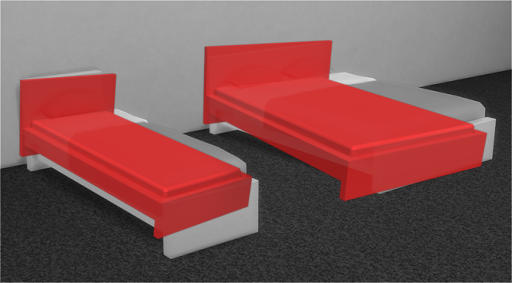}&  \colImgN{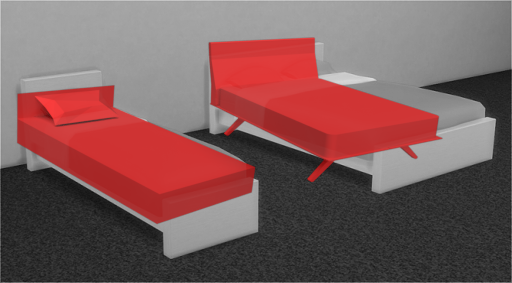}\\[-1.5pt]
		
		\colImgN{}&   \colImgN{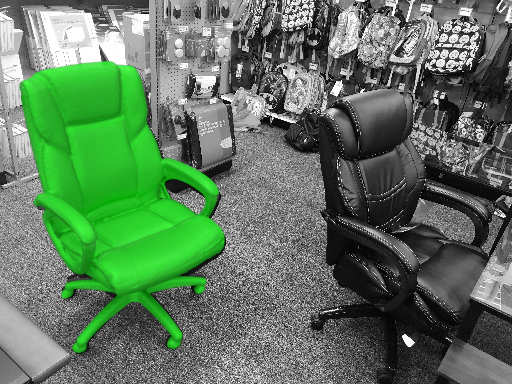}&
		\colImgN{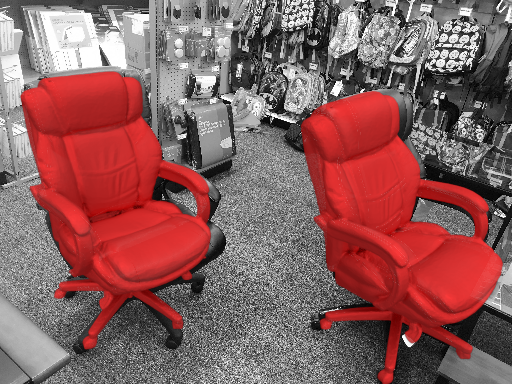}&  \colImgN{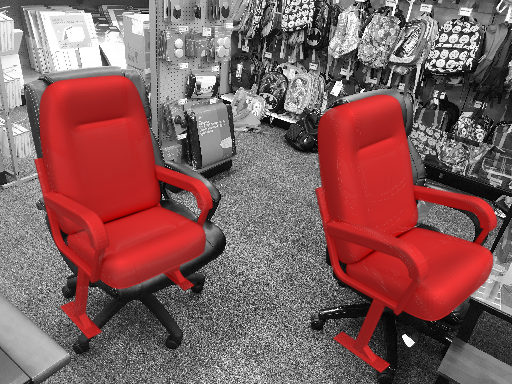}\\[-1.5pt]
		
		\colImgN{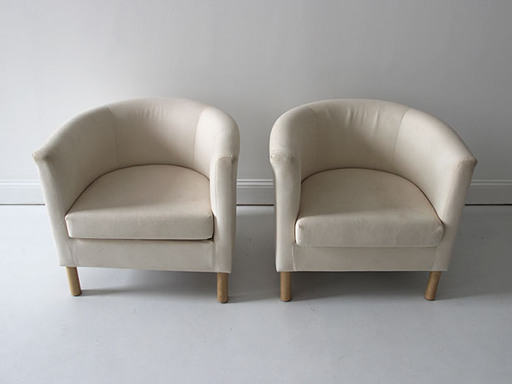}&   \colImgN{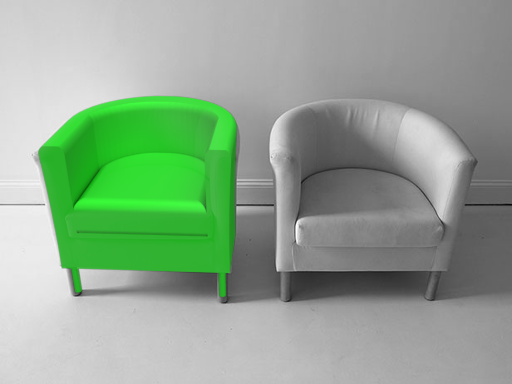}&
		\colImgN{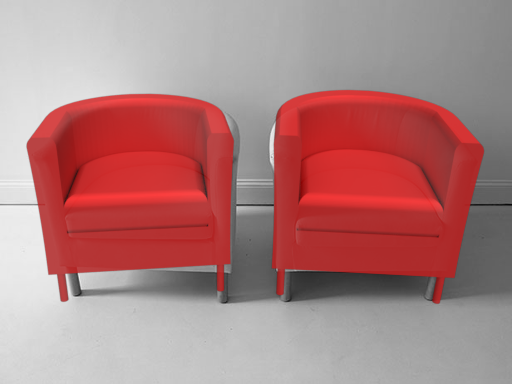}&  \colImgN{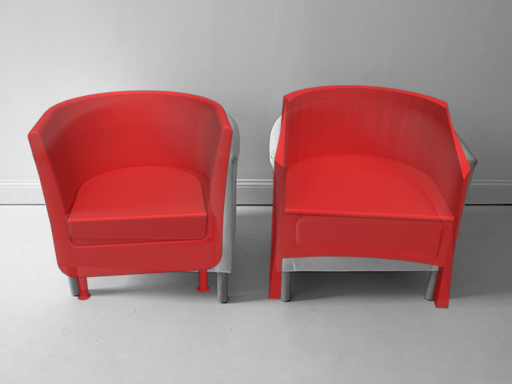}\\[-1.5pt]
		
		\colImgN{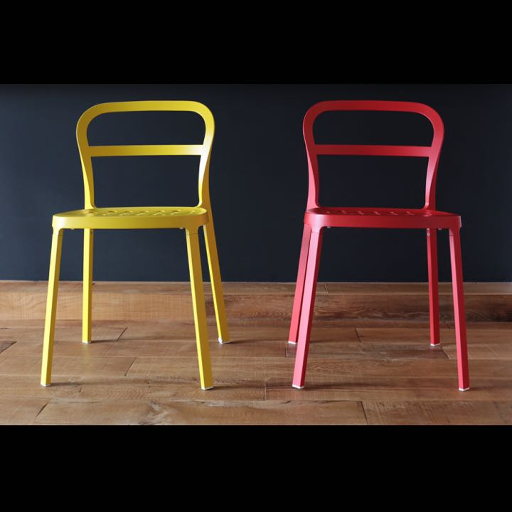}&   \colImgN{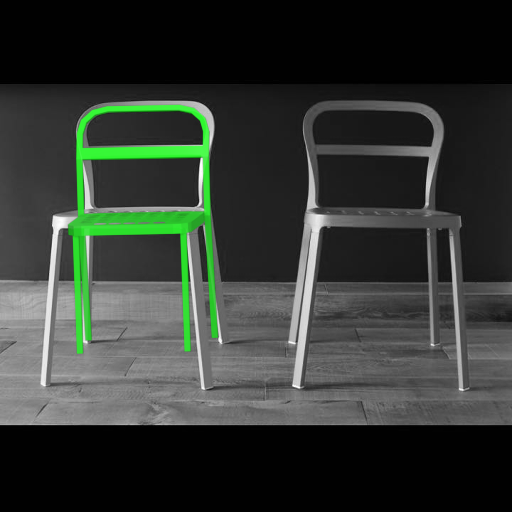}&
		\colImgN{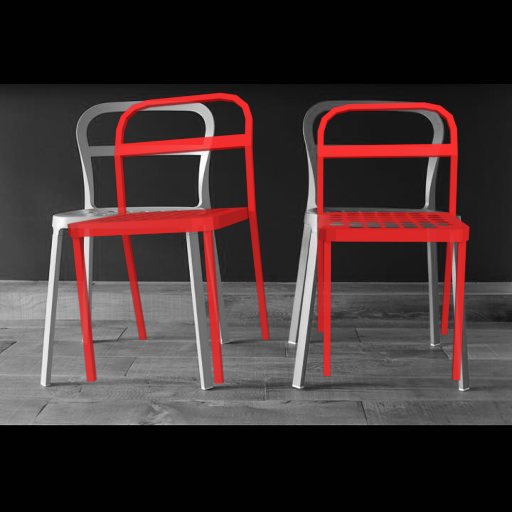}&  \colImgN{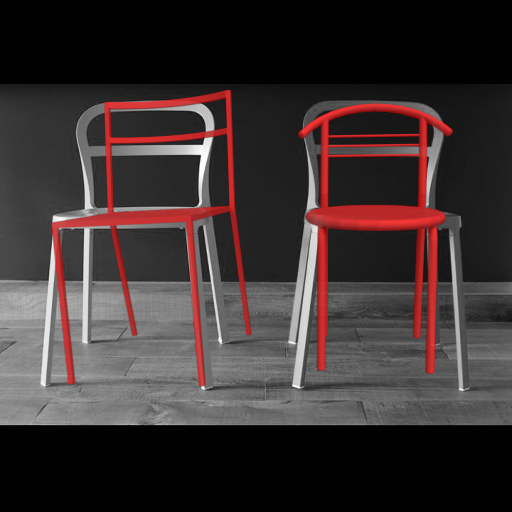}\\[-1.5pt]
		
		\colImgN{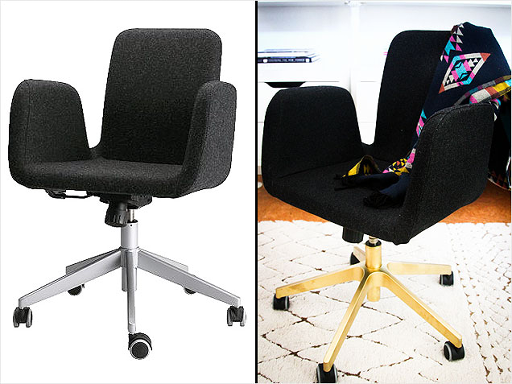}&   \colImgN{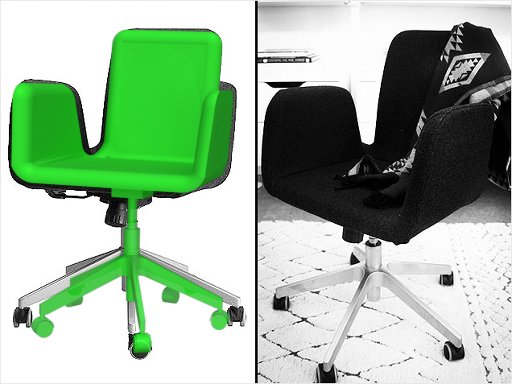}&
		\colImgN{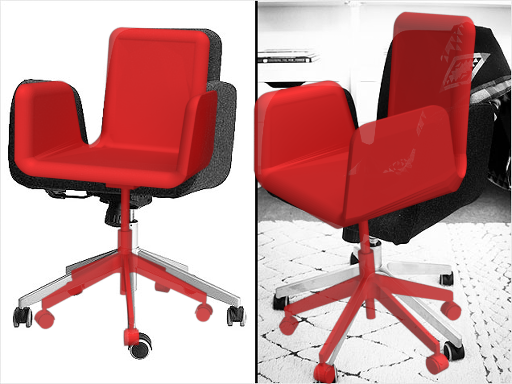}&  \colImgN{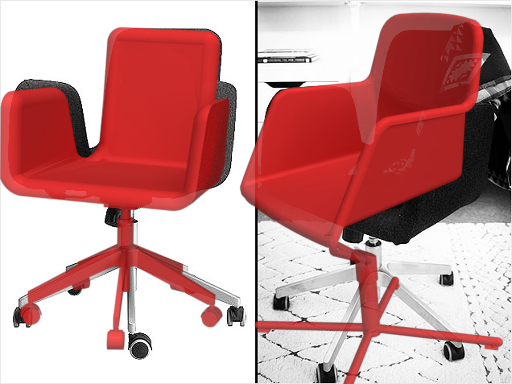}\\[-1.5pt]
		
		\colImgN{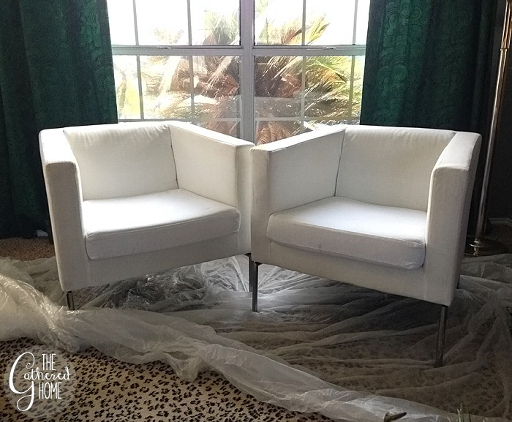}&   \colImgN{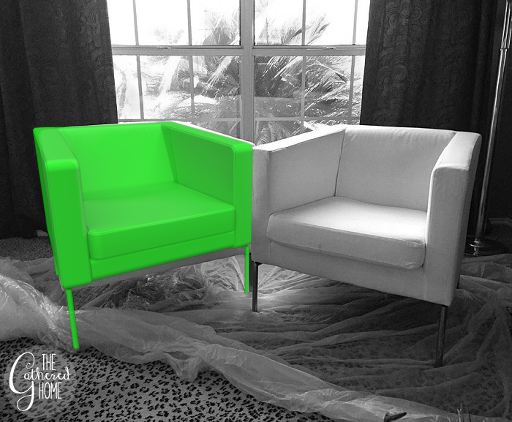}&
		\colImgN{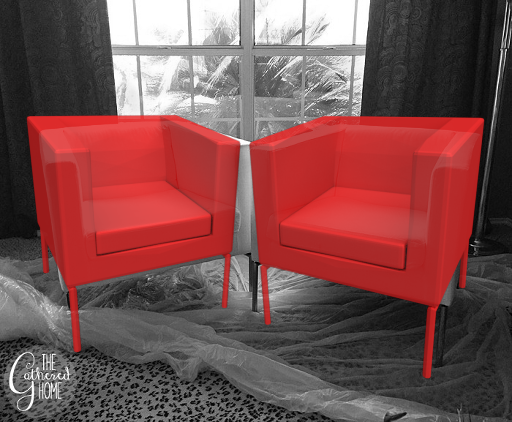}&  \colImgN{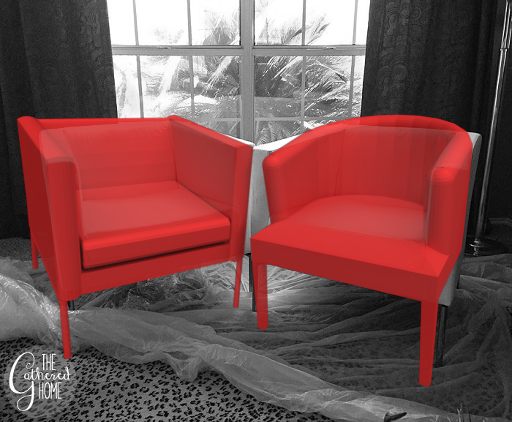}\\[-1.5pt]
		
		\footnotesize Image&\footnotesize GT&\footnotesize from seen&\footnotesize from unseen\\[-3pt]
	\end{tabular}
	\end{subfigure}\hfill\begin{subfigure}[t]{\columnwidth}
	\begin{tabular}{cccc}
		
		\colImgN{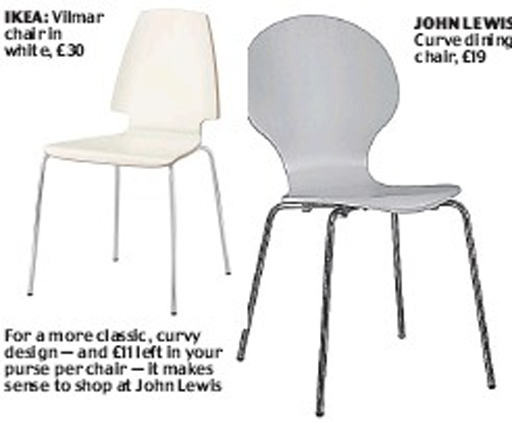}&   \colImgN{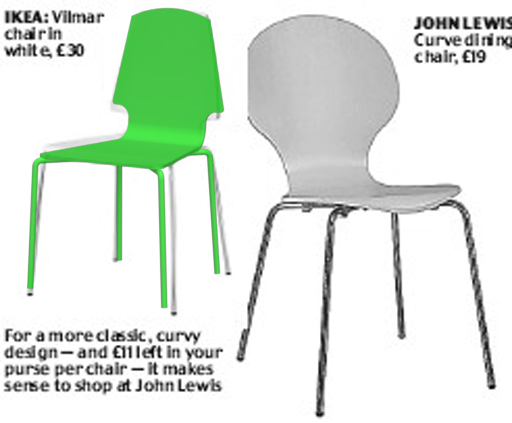}&
		\colImgN{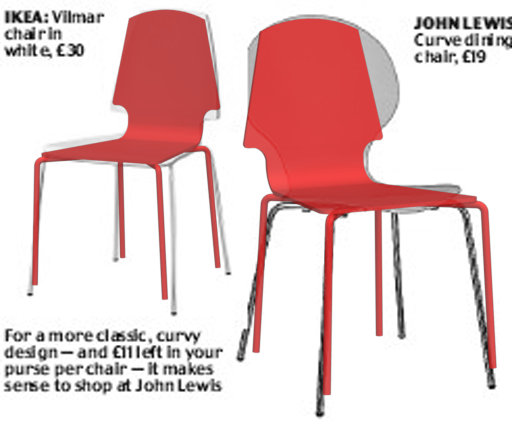}&  \colImgN{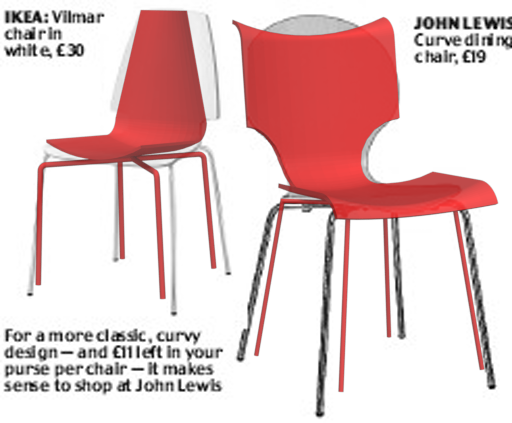}\\[-1.5pt]
		
		\colImgN{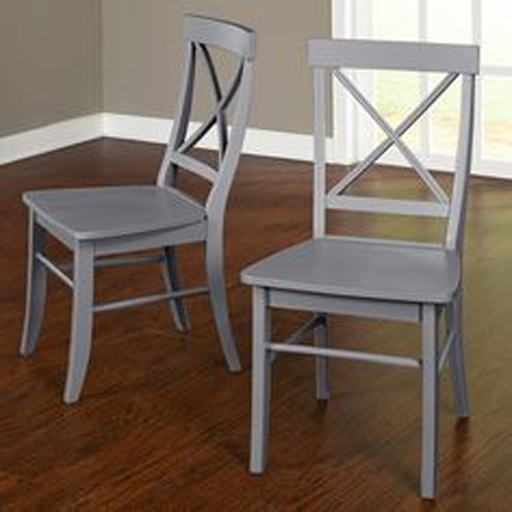}&   \colImgN{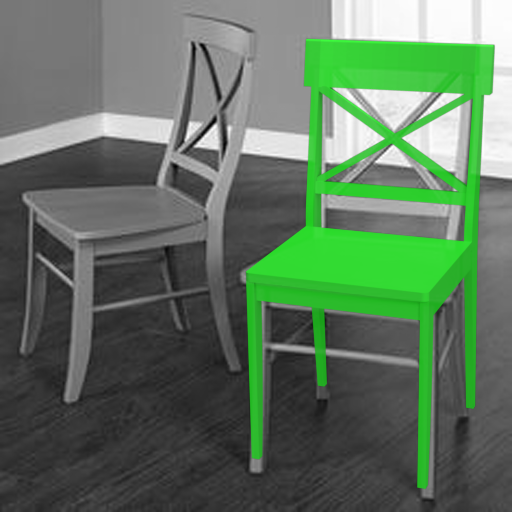}&
		\colImgN{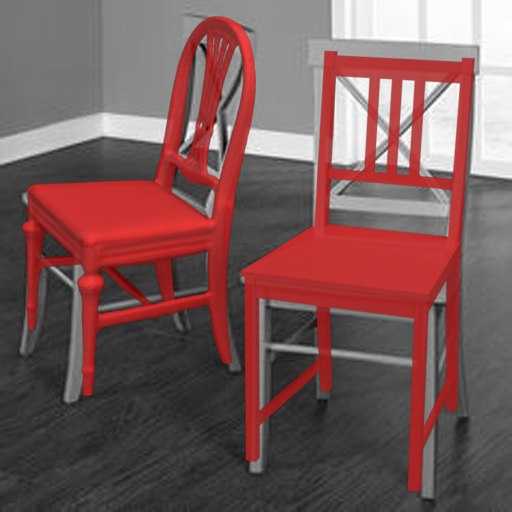}&  \colImgN{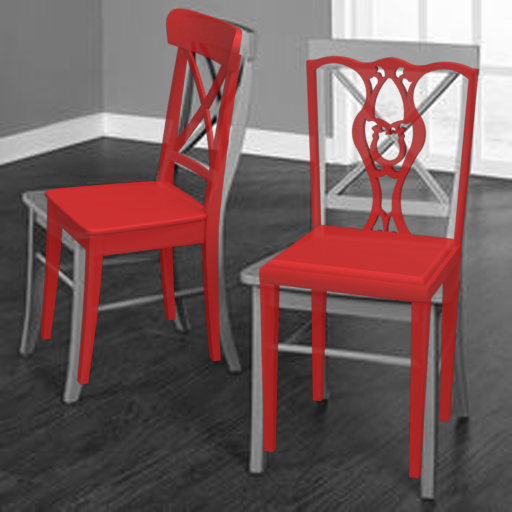}\\[-1.5pt]
		
		\colImgN{}&   \colImgN{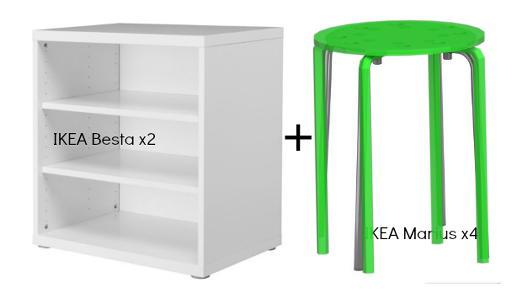}&
		\colImgN{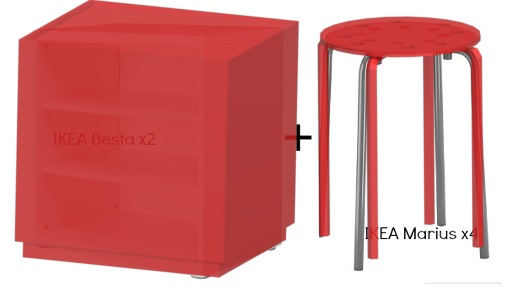}&  \colImgN{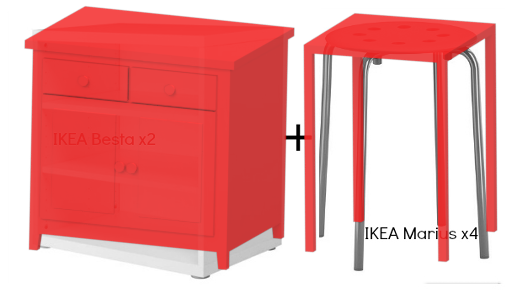}\\[-1.5pt]
		
		\colImgN{Images/multi/multi_I_20.png}&   \colImgN{Images/multi/multi_gt_20.png}&
		\colImgN{Images/multi/multi_pred_20.png}&  \colImgN{Images/multi/multi_shapenet_20.png}\\[-1.5pt]
		
		\colImgN{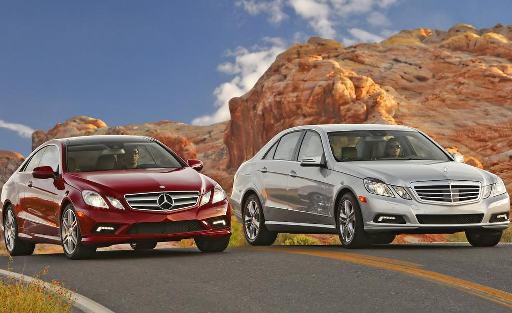}&   \colImgN{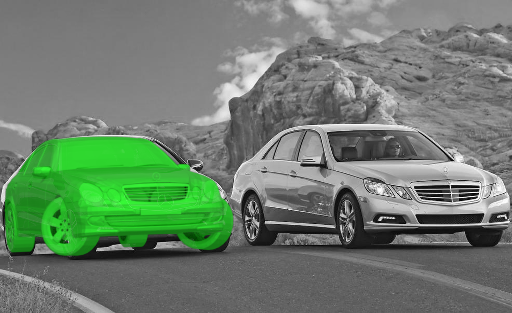}&
		\colImgN{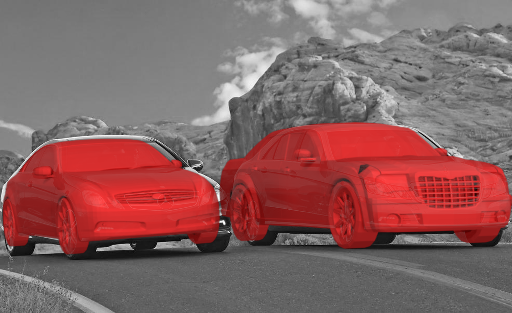}&  \colImgN{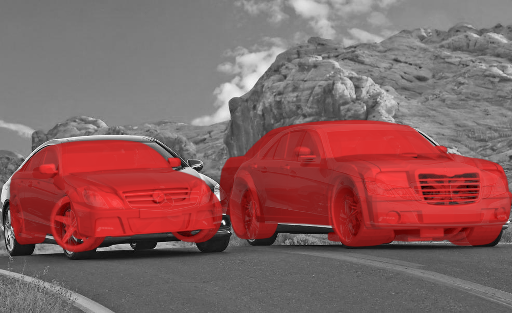}\\[-1.5pt]
		
		\colImgN{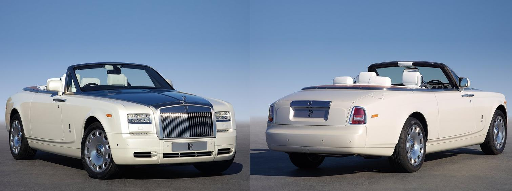}&   \colImgN{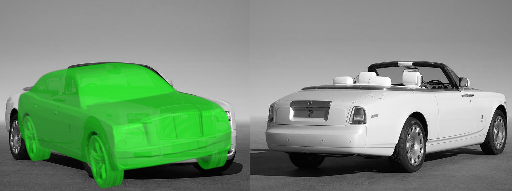}&
		\colImgN{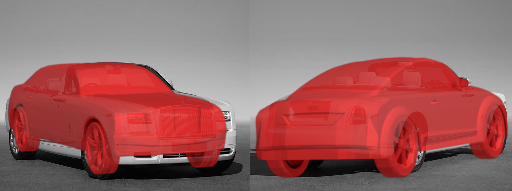}&  \colImgN{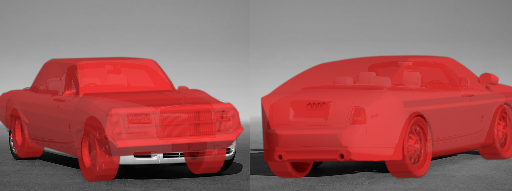}\\[-1.5pt]
		
		\colImgN{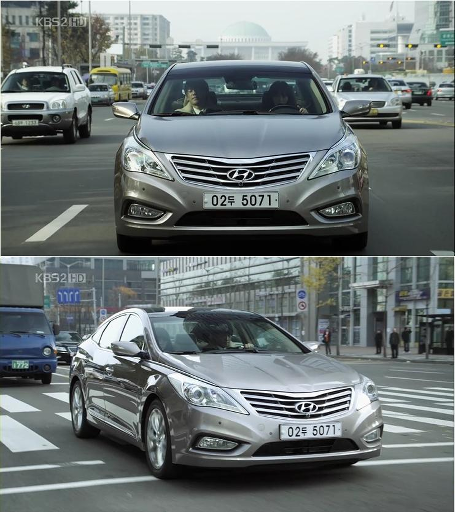}&   \colImgN{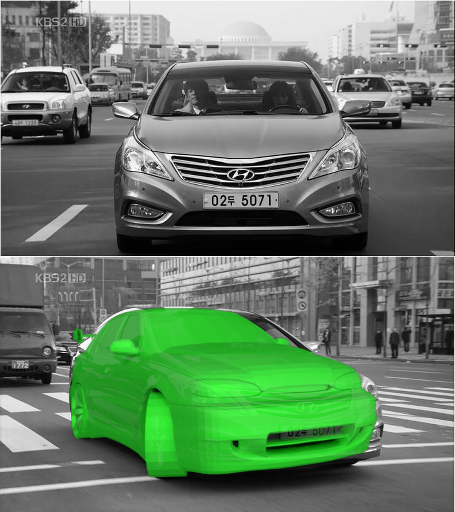}&
		\colImgN{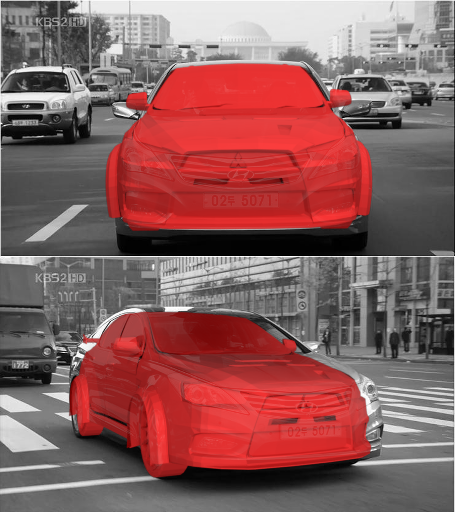}&  \colImgN{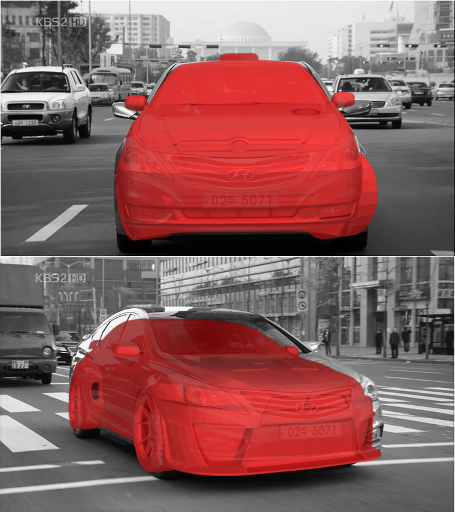}\\[-1.5pt]
		
		\colImgN{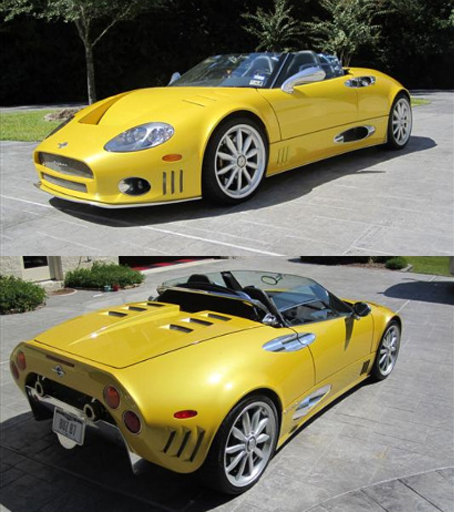}&   \colImgN{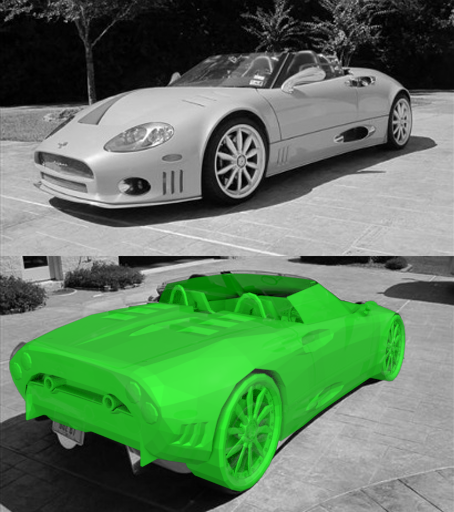}&
		\colImgN{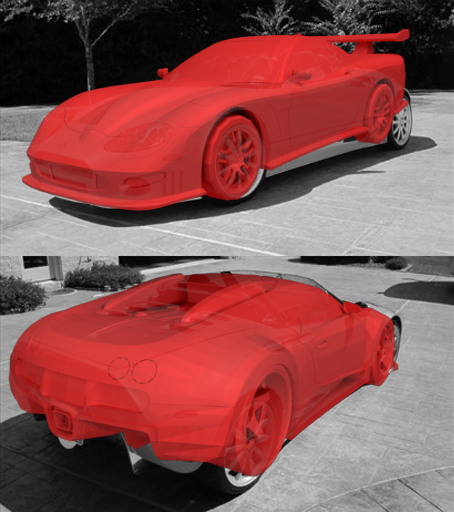}&  \colImgN{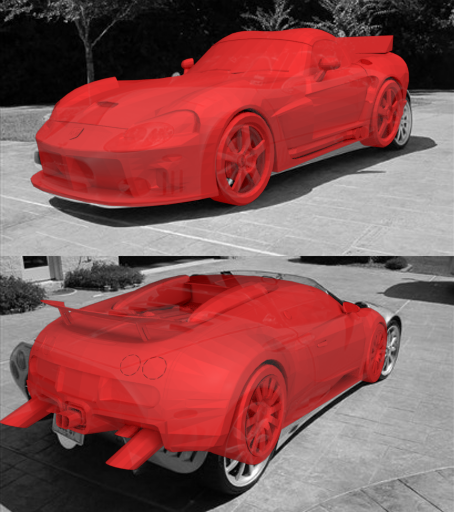}\\[-1.5pt]
		
		\colImgN{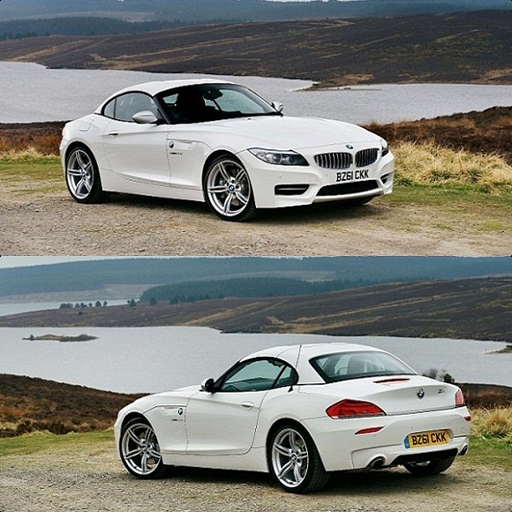}&   \colImgN{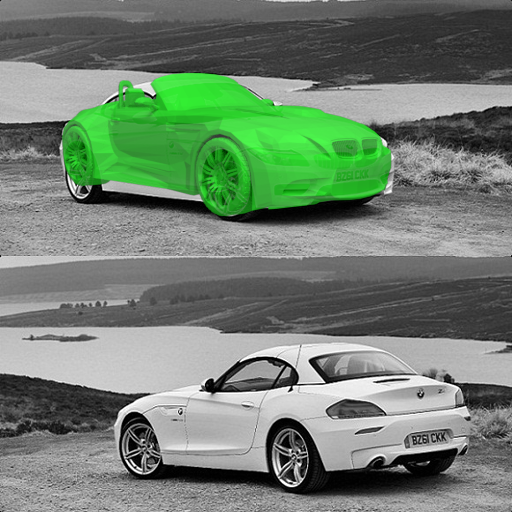}&
		\colImgN{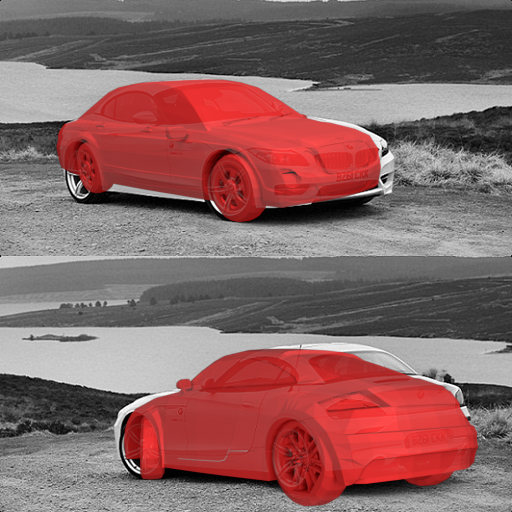}&  \colImgN{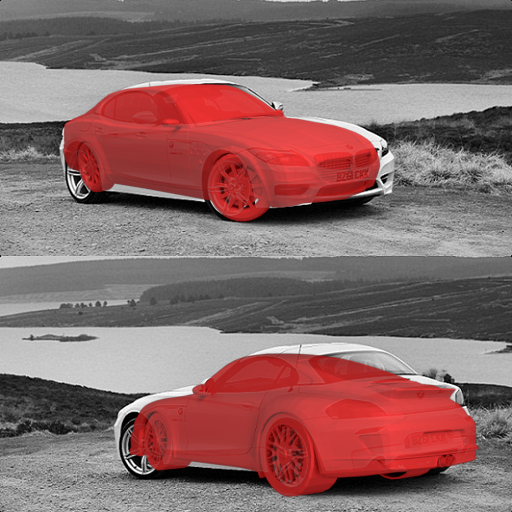}\\[-1.5pt]
		
		\footnotesize Image&\footnotesize GT&\footnotesize from seen&\footnotesize from unseen\\[-3pt]
	\end{tabular}
	\end{subfigure}
	\caption{Additional qualitative results for 3D pose estimation and 3D model retrieval from both seen and unseen databases. We project the retrieved 3D model onto the image using a 3D pose computed from the predicted location field by solving a \PNP~problem. For the ground truth 3D model, we use the ground truth 3D pose. In fact, location fields provide all relevant information to jointly address both tasks.}
	\label{fig:ret_pose2}
\end{figure*}

Figs.~\ref{fig:ret_pose1} and~\ref{fig:ret_pose2} present more qualitative results for 3D model retrieval from both seen and unseen databases. In addition, we show that our predicted location fields provide all relevant information to also compute the 3D pose of objects. For this purpose, we sample 2D-3D correspondences from the location field and solve a \PNP~ problem during inference. The projections onto the image show that both our retrieved 3D models and our computed 3D poses are highly accurate. Our approach naturally handles multiple objects in a single image, however, all evaluated datasets only provide annotations for a single instance per image, as shown in Fig.~\ref{fig:ret_pose2}.

\end{document}